%% file: main_arxiv.tex
\documentclass[english]{article}

\usepackage{microtype}
\usepackage{graphicx}
\usepackage{subfigure}
\usepackage{makecell}
\usepackage{booktabs} % for professional tables
\usepackage{multirow}
\usepackage{microtype}
\usepackage{graphicx}
\usepackage{subfigure}
\usepackage{booktabs} % for professional tables
\usepackage{amsfonts}
\usepackage{tcolorbox}
\usepackage{multirow}
\usepackage{amsmath}
\usepackage{amssymb}
\usepackage{hyperref}

\usepackage[preprint]{mystyle}

\usepackage{qiangstyle}

\title{How to Fill the Optimum Set?  \\ Population Gradient Descent with Harmless Diversity} 
\author{%
  Chengyue Gong \thanks{Equal Contribution} \\
  UT Austin \\
  \texttt{cygong@cs.utexas.edu} \\
    \and
  Lemeng Wu \textsuperscript{*} \\
  UT Austin\\
  \texttt{lmwu@cs.utexas.edu} \\
  \and
   Qiang Liu \\
  UT Austin \\
  \texttt{lqiang@cs.utexas.edu} \\
}
\date{} 
\begin{document}

\twocolumn[
  
    \maketitle

]
{
  \renewcommand{\thefootnote}%
    {\fnsymbol{footnote}}
  \footnotetext[1]{Equal contribution}
}

%\twocolumn[
% \icmltitle{Diversity-aware Gradient descent: a harmless approach}
%\icmltitle{How to Fill the Optimal Set?  \\ Population Gradient Descent with Harmless Diversity}
% \icmltitle{Finding Diverse Points within Optimum Set: \\ Population Gradient Descent with Harmless Diversity}






%\printAffiliationsAndNotice{}  % leave blank if no need to mention equal contribution\printAffiliationsAndNotice{\icmlEqualContribution} % otherwise use the standard text.

\begin{abstract}
Although traditional optimization methods focus on finding a single optimal solution, 
most objective functions in modern machine learning problems, especially those in deep learning, often have multiple or infinite number of optima. 
Therefore, it is useful to consider the problem of 
finding a set of diverse points in the optimum set of an objective function.  
In this work, 
we frame this problem as a bi-level optimization problem of maximizing a diversity score inside the optimum set of the main loss function, and 
solve it with a simple population gradient descent 
framework that iteratively updates the points to maximize the diversity score in a fashion that does not hurt the optimization of the main loss. 
We demonstrate that our method can efficiently generate diverse solutions on a variety of applications, including text-to-image generation, text-to-mesh generation, molecular conformation generation and ensemble neural network training.
\end{abstract}

\input{tex/introduction}

\input{tex/main_method}

\input{tex/related_works}

\input{tex/experiment}

\input{tex/conclusion}

\bibliography{z_ref}
\bibliographystyle{icml2021}

\appendix

\onecolumn
\newpage \clearpage

\input{tex/appendix}

\end{document}

%% file: tex/introduction.tex
\section{Introduction}

Most traditional 
%theories and algorithms in gradient based optimization 
optimization methods in machine learning aim to find a single optimal solution 
for a given objective function. 
%that minimize an objective function. 
%The underlying assumption is often that 
%tend to operate under the assumption that the target is to find a single solution in the optimal set. % \citep[e.g.][]{boyd2004convex, ben2001lectures, bertsekas2003convex}.
However, in many practical applications, 
the objective functions
tend to have multiple or even infinite number of (local or global) optimum points, for which it is of great interest to find  
%(local or global) optimal solutions,%and people are interested in
 a set of diverse points that are representative of the whole optimum set.   
%parallelly or sequentially, 
This is tremendously useful in a variety of machine learning tasks, including, for example, 
ensemble learning \citep{lakshminarayanan2016simple, pang2019improving}, 
robotics \citep{Cully2015RobotsTC, osa2020multimodal},
generative models \citep{lee2018diverse, shi2021learning}, 
% global optimization \citep{laguna2005experimental},
latent space exploration of generation models \citep{liu2021fusedream, fontaine2021differentiable}
robotics and reinforcement learning  \citep{vannoy2008real, conti2017improving, parker2020effective}.

%This is needed because the objective functions in practical applications, especially these in deep learning, tend to have multiple (local or global) optimal solutions, %and finding diverse multiple solutions can provide robust results when combined, offer users more useful candidates and enlarge the solution set for sequential decision making. 
%there are often multiple solutions that perform well Two special and well-known cases are: 
%\textbf{Multi-modal Objectives} 
%The objective has many isolated local optima that are separated by energy barriers. Multi-start gradient descent \citep[e.g.][]{wu2017bayesian, toscano2018effort} is practically used to fill multiple modals. However, it is hard to guarantee that multiple restart can find different in different modals.

%\textbf{Over-parameterized Objectives}
Finding diverse solutions is particularly relevant 
in modern deep learning applications, 
in which it is common to use very large, overparameterized neural networks whose number of parameters is larger than the size of training data \citep[e.g.][]{radford2021clip, fedus2021switch, brown2020gpt3}. 
In these cases, 
the set of models that perfectly fit the training data (and hence optimal w.r.t. the training loss) consist of 
low dimensional manifolds of an infinite number of points. 
%we can often perfectly fit the clean training data with zero training error, yielding loss functions whose  minimum set  is 
%low dimensional manifolds that consists of infinite number of points \citep[e.g.,][]{du2019gradient}.
It is hence useful to 
explore and profile the whole solution manifold by finding diverse representative points. % in the solution manifold.  
%rather than a set of isolated points. 

% Denote that the optimum set of the loss function in these cases are non-convex, gradient-based methods usually fail to fill the optimum set for both above cases.

%Perhaps the simplest approach
A straightforward approach to obtaining multiple optimal solutions is to run multiple trials of optimization with random initialization 
%use multi-start gradient descent 
\citep[e.g.][]{wu2017bayesian, toscano2018effort}. 
However, this does not explicitly enforce the diversity preference. 
%it is hard to guarantee that these different trials can find different solutions. %Gradient-based methods, on the other hand, often
Another approach is to jointly optimize a set of solutions with a diversity promoting regularization term  \citep[e.g.,][]{pang2019improving, xie2015diversifying, croce2020reliable, xie2016diversity}. % or reinitialization \citep{lakshminarayanan2016simple}.
However, the regularization term can hurt the optimization of the main objective function without a careful tuning of the regularization coefficient. 
%These approaches cannot effectively get diverse solutions without hurting the target loss in many situations. 
%Previously,
Evolutionary algorithms \citep[e.g.][]{Cully2015RobotsTC, flageat2020fast, mouret2015illuminating} and genetic algorithms \citep[e.g.][]{lehman2011evolving, gomes2013evolution, lehman2011abandoning} 
are also useful for finding diverse solutions. % have been developed to find diverse solutions. 
However, these black-box algorithms do not leverage gradient information and tend to 
require a large number of query points for large-scale optimization problems.  % and is neither effective nor efficient for neural network training compared to gradient descent based optimization algorithms.
%Gradient-based methods, on the other hand, often encourage diversity with a regularization term \citep{pang2019improving} or reinitialization \citep{lakshminarayanan2016simple}.
%These approaches cannot effectively get diverse solutions without hurting the target loss in many situations. 

In this work, we consider this problem with a bi-level optimization perspective: we want to maximize a diversity score of a set of points within the minimum set of a given objective function (i.e., \emph{diversity within the optimum set}). 
We  solve the problem with a simple gradient descent like approach that iteratively updates a set of points to maximize the diversity score while minimizing the main loss in a guaranteed fashion. The key feature of our method is that it ensures to optimize the main loss as a typical optimization method while adding diversity score as a secondary loss that is minimized to the degree that 
does not hurt the main loss. 
%based on minimiz
%In particular, we propose 

%it finds solutions of 

%1) the points are ensured to converge 
%we encourage 
%a  population gradient descent framework for obtaining a set diverse points in the optimum set of a given objective function $f$. 
%The idea is to modify the vanilla gradient descent algorithm  to introduce  a .
%harmless diversity with a proposed population gradient descent framework. 
%We formulate the optimization as a constrained optimization problem, and then derive the update rule to guarantee the monotonic decrease the target function value. We propose two different version algorithms, $\Fsum$ and $\Fmax$ within our framework by changing the weights for different particles. 
%After understanding the behaviour of our methods on toy examples, 
We propose two variants of our method 
that control the minimization of the main loss in different ways (by descending the sum and max of the population loss, respectively). 
For the choice of the diversity score, 
we advocate using a Newtonian energy, 
which provides more uniformly distributed points than typical variance-based metrics.
We test our methods in a variety of  practical problems, including text-to-image, text-to-mesh, molecular conformation generation, and ensemble neural network training. 
%In these cases, 
Our methods yield an efficient trade-off between diversity and quality, both 
 quantitatively and qualitatively. % have a better
%Compared to black-box algorithms, ours are more efficient and wider applicability.
% In both cases, how to find a set of diversified points that are representative of the optimum set? 

%% file: tex/main_method.tex
\section{Harmless Diversity Promotion} 
\label{sec:main}
%Let $f \colon \RR^d \to \RR$ be  a loss function of interest. Our want to 
\paragraph{Problem Formulation} 
Let $f(x)$ be a differentiable loss function $f(x)$ on domain $\set X\defeq \RR^d$.  Let $\argmin f$ be the set of minima of $f$, which we assume is non-empty. 
 %Assume $f$ is differentiable and lower bounded. 
Our goal is to find a set of $m$ points (a.k.a. particles) $\vv x  \defeq \{x_i\}_{i=1}^m$ in the minimum set $\argmin f$ that minimizes 
a preference function  $\Phi(\vx ) = \Phi(x_1,\ldots, x_m)$. 
Formally, this 
yields a bi-level optimization problem: 
\bbb \label{equ:argmin} 
\min_{\vv x \in \X^m }\Phi(\vv x)
%= \Phi(x_1,\ldots, x_m),
~~~~s.t.~~~~
%x_1,\ldots, x_m \in 
\vv x \subseteq \argmin f. 
\eee 
%This formulation allows us to 
So we want to 
minimize $\Phi$ as much as possible, but \emph{without} scarifying the main loss $f$. Because practical loss functions, such as these in deep learning, 
often have multiple or infinite numbers of minimum, 
optimizing $\Phi$ inside the optimum set allows us to gain diversity ``for free", compared with applying standard optimization methods on $f$. 

%\paragraph{How to Measure Diversity}
$\Phi$ can be a general differentiable function that can encode arbitrary preference that we have on the particles.  
%But we focus on encouraging diversity using 
In this work, for encouraging diversity, we consider 
%$\Phi$ to be  
the Riesz $s$-energy \citep[e.g.,][]{gotz2003riesz, kuijlaars2007separation}, % \red{[xxx]}, %which encourages the diversity amount the points: 
\begin{align*}
\Phi_s(\vv x)    = 
\begin{cases} 
\displaystyle 
\frac{1}{s}\sum_{i\neq j} {\norm{x_i - x_j}^{-s}}, & \text{if $s \neq 0$},  \\ 
%^\vspace{1\baselineskip}\\
\displaystyle
\sum_{i\neq j} \log\left (\norm{x_i- x_j}^{-1}\right), & \text{if $s = 0$},  
\end{cases} 
\end{align*}
 where $s \in 
\RR$ is a coefficient. 
Different choices of $s$ yield different energy-minimizing configurations of points. 
A common choice is $s=-2$, 
with which $\Phi_s$ reduces to the negative variance. % or Gini index. 
%reducing to 
%Compared to $\ell_2$ ($s=-2$) distance, intuitively, Riesz $s$-energy has larger penalty when two particles are close.
On the other hand, when $s = d-2$ where $d$ is the dimension of the input $x$, 
it reduces to the 
%the %energy is known as
Newtonian energy in physics. 
The case when $s=0$ is known as the logarithm energy. 
In this work,  %to better encourage diversity, 
we advocate using 
a non-negative $s\geq0$, which places a strong penalty on the small distances between points, and hence yields more uniformly distributed points as shown in the 
experiments and the toy example blew.   %without paying that much a%ttention to the distances that are relatively large.
%In parti
% As suggested by Newtonian and Couloumbian law, it is typical to choose $s = d-2$,
%The points that minimizes the Riesz $s$-energy on a manifold are called Fekete points. 
\begin{exa}
Consider two sets of points in $\RR$: 
\bb 
\vv x= \{0, ~0, ~2\}, && \vv x'= \{0,~ 1,~ 2\}. 
\ee 
Although  $\vv x'$ is  clearly more uniformly distributed, 
one can show that $\vv x$ has larger variance and hence is preferred by $\Phi_s$ with $s = -2$. On the other hand, $\vv x'$ is preferred over $\vv x$ by $\Phi_s$ with any $s \geq 0$. In fact, it is easy to see that $\Phi_s(\vv x) =+\infty > \Phi_s(\vv x')$ for $\forall s\geq 0$.
\end{exa}

In practice, when $\vv x$ is a structured objective such as image or text, it is useful to  map the input into a feature space before applying Riesz $s$-energy, i.e., 
%$\Phi$ can be defined as applying 
we  define
$\Phi(\vv x) = \Phi_s(\psi(x_1),\ldots, \psi(x_m))$, where $\psi$ is a neural network feature extractor trained separately that 
maps  each $x_i$ to a feature vector. % $z_i = \psi(x_i)$. % that is suitable for the problem.

\textbf{Main Idea } The %harmless diversity
bi-level optimization 
problem in \eqref{equ:argmin} is equivalent to a constrained optimization problem: 
\bbb \label{equ:fsta}
\min_{\vv x\in \set X^m} \Phi(\vv x) ~~~~~s.t. ~~~~  f(x_i) \leq f^*,~~~~\forall i\in[m],
%x_i \in \argmin f,~~~  \forall i \in[n].
\eee 
where $f^* \defeq \min_{x} f(x)$ and the $m$ constraints $f(x_i) \leq f^*$ ensure that all $\{x_i\}_{i=1}^m$ are optima of $f$.  
 
To yield a simple and efficient algorithm, we propose to combine the $m$ constraints  in \eqref{equ:fsta} 
into a single constraint: 
%Equivalently, we have 
\bbb \label{equ:phiF}
\min_{\vv x \in \set X^m} \Phi(\vv x) ~~~~~s.t. ~~~~  %\sum_{i=1}^m f(x_i) 
F(\vv x) 
\leq F^*, % \defeq \min_{\vv z} F(\vv z), 
%x_i \in \argmin f. 
\eee 
where 
$F^*\defeq \min_{\vv z} F(\vv z)$  and 
$F(\vv x)$ is a utility function defined such that \eqref{equ:fsta} and \eqref{equ:phiF} are equivalent: 
$$
\left \{ F(\vv x) \leq F^* \right \}
\iff 
\left \{ f(x_i) \leq f^*,~~\forall i\in[m] \right \}. 
$$
%Here, $F$ is a weighted function whose solution setis %equivalent to the solution set of minimizing $f$ for %each particle in $\vv x$
%^\paragraph{How to Choose $F$}
%
In this work, we consider two natural choices of $F(\vx)$: 
\bb
\Fsum(\vv x) = \sum_{i=1}^m f(x_i), &&
\Fmax(\vv x) = \max_{i\in[m]} f(x_i),
\ee 
both of which clearly ensures the equivalence of 
\eqref{equ:fsta} and \eqref{equ:phiF}. % are equivalent: 
 %these 
%$introduction of 
%the choice of $F$ determin
%or more generally any
%See \red{xxx} for more choices of $F$. 
%and $F^* = \min_{ }$
%where $f^* = \min_x f(x)$. 

We proceed to develope the two  algorithms  based on $\Fsum$ in 
Section~\ref{sec:fsum} and $\Fmax$ in Section~\ref{sec:fmax}, respectively. 
%Our method is designed to 
The idea of both methods 
is to iteratively update $\vx $ following a gradient-based  direction which ensures that 
%in a way such that 
%in a way that ensures that 
%We propose to solve \eqref{equ:phiF} 
%with iterative update 
%$$\vv x^{t+1} \gets \vv x_t - \epsilon_t \vv v_t,$$where $\vv v = \{v\ti\}_{i=1}^m\subset \RR^{d}$ is an update direction chosen toensure that 

1) $F(\vv x)$ is monotonically decreased stably across the iteration, ensuring all $\vx$ to converge to (local) optimal of $f$;  

2) $\Phi(\vx)$ is minimized as the secondary loss to the degree that it does not conflict with the descent of $F(\vx)$.  
%when it is not conflicting the decrease of $F$, we should %minimize $\Phi.$
%\emph{simultaneously} optimize $
%F$ and $\Phi$, with 
% 

Besides %simplifying the algorithm, 
the benefit of obtaining a single constraint, 
the introduction of $F$ allows 
different particles to exchange loss to decrease $\Phi$ more efficiently: 
it is possible for some particles $x_i$ to increase their $f(x_i)$ to decrease $\Phi$, once the overall $F$ is ensured to decrease. 
%us to decrease $\Phi$ more efficiently 
%As we show, the introduction of $F$ simplifies the derivation of the algorithms since we do not 
As shown in the sequel, 
%$\Fsum$ and $\Fmax$ provide different flexibility in this respect. 
%for this kind of trade off. 
%provide two different ways for different particles to exchange loss during the optimization process. 
$\Fmax$ gives more flexibility for decreasing $\Phi$, and hence yields more diverse solutions than $\Fsum$, 
but with the trade-off of converging slower.

%\begin{tcolorbox}
% \scalebox{1}{
% \begin{minipage}{1.0\linewidth}
\begin{algorithm}[H] 
\caption{Diversity-aware Gradient Descent ($\Fsum$)}
\begin{algorithmic} 
\label{alg:mainsum} 
\STATE \textbf{Goal}: Find a set of $m$ diversified local optima of $f(x)$.  \\
\textbf{Parameters}: step size $\mu$, a repulsive coefficient $\eta$.  
\FOR{Iteration $t$} % At the $t$-th iteration: %Take
\vspace{-1.5\baselineskip}
\STATE\begingroup\makeatletter\def\f@size{8}\check@mathfonts
\def\maketag@@@#1{\hbox{\m@th\large\normalfont#1}}
\bb
\scriptsize x_{t+1,i} \longleftarrow x\ti - \mu \dd f(x\ti) - \eta \mu \sqrt{\frac{\sum_{i=1}^m \norms{\dd f(x\ti)}^2}{\sum_{i=1}^m \norms{g\ti}^2}} g\ti, 
\ee \endgroup
where $g\ti = \dd_{y\ti}\Phi(\vv y_t)$, 
and $\vv y_t$ is defined in \eqref{equ:yt}. 
\ENDFOR
%and $\alpha = \sart{\frac{\sum_{i=1}^m\norm{\dd f(x\ti)}^2}{\sum_{i=1}^m \norm{g_\ti}}$
\end{algorithmic}
\end{algorithm}
% \end{minipage}
% }  
%\end{tcolorbox} 

\subsection{$\Fsum$-Descent} \label{sec:fsum}
\iffalse 
\red{ 
We first focus on the case when $F = \Fsum$, derive a gradient-based iterative algorithm to optimize \eqref{equ:phiF} and do some analyses in the sequel. 
We start from a guarantee on monotonically decreasing 
$\Fsum$ value under proper condition, and then solve an inner optimization problem to get the update scheme at each iteration. 
} 
\fi 
%\paragraph{Derive Update Rule}

\iffalse
% we need to control the error of Taylor approximation. 
\begin{ass}
Assume $f$ is  $1/\mu$-smooth with $\mu\in(0,+\infty)$, that is, 
for any $x,x'\in \set X$, 
\bbb \label{equ:fsmooth}
f(x') \leq f(x) + \dd f(x) \tt  (x' - x) + \frac{1}{2\mu} \norms{x' - x}^2. 
\eee 
\end{ass}
\fi
We now derive a simple algorithm 
that decreases the sum of loss $\Fsum$ monotonically while minimizing $\Phi$ as the secondary loss.  % as much as possible. 

Assume we have $\vv x_t = \{x_{t,i}\}_{i=1}^m$ at the $t$-th iteration  
of the algorithm. 
To decrease $\Fsum$, 
the update direction $\vv x_{t+1} - \vv x_t$ should be close the gradient descent direction. 
Let $\vv y_t$ be the result of applying gradient descent on $\Fsum$ from $\vv x_t$: %, which corresponds to GD on individual $f$:
\bbb \label{equ:yt} 
y_{t,i} = x_{t,i}  - \mu \dd f( x_{t,i}),~~~\forall i \in[m],
%\vv y_{t} =\vv x_{t}  - \mu \dd \Fsum(\vv x_t),
\eee 
where $\mu>0$ is a step size. 
%Define $\vv y_t = \{y\ti\}_{i=1}^m$ 
%be the result we obtained by performing gradient descent on$x\ti$: 
%$$
%y\ti = x\ti - \mu \dd f(x\ti). 
%$$
Assume $f$ is $1/\mu$-smooth: %, which gives 
%We have
\bbb\label{equ:smooth}
f(x') \leq f(x) + \dd f(x) \tt  (x' - x) + \frac{1}{2\mu} \norms{x' - x}^2,
\eee 
for any $x, x'\in \X$. 
Applying \eqref{equ:smooth} to $x_{t,i}$ and sum over $i$ gives 
%and it follows immediately that 
%We have 
\bb 
\Fsum(\vv x) \leq 
\Fsum(\vv x_t) + 
\frac{1}{2\mu}\left (
\norms{\vv x - \vv y_t}^2
- \xi_t^2 \right), 
&& \forall \vx,
\ee 
where $\xi_t^2:=\norms{\vx_t - \vv y_t}^2 = \norms{\mu \dd F(\vx_t )}^2.$
\iffalse 
and $\vv y=\{y\ti\}_{t=1}^m$ is  
%$\norms{\vv x - \vv y_t}^2 = \sum_{i=1}^m \norms{x_i - y\ti}^2$ with %where %$y\ti$  
the result of gradient descent from $\vx_t = \{x\ti\}$: 
\bbb \label{equ:yt} 
y\ti = x\ti - \mu \dd f(x\ti), ~~~~\forall i \in[m]. 
%\xi^2= \sum_{i=1}^m \norms{\mu \dd f(x\ti)}^2, 
\eee  
\fi 
%and $\norm{\vv x - \vv y_t}^2 = \sum_{i=1}^m \norm{x_i - y\ti}^2$. 
Therefore, 
to ensure that $\Fsum$  decreases, 
it is sufficient to ensure that $\norms{\vx_{t+1} - \vv y_t}\leq \xi_t$. 
%we may want to choose $\vv x_t$ such that $\norms{\vx_t - \vv y_t}$ is smaller than $\xi$. 
%$$
%\sum_{i=1}^m \norm{x_i - y\ti}^2 
%$$

On the other hand, Taylor approximation of $\Phi$ on $\vv y_t$ gives 
$$
\Phi(\vx)= \Phi(\vv y_t) + %sum_{i=1}^n 
\dd \Phi(\vv y_t)\tt  (\vx - \vv y_t) + \bigO(\norms{\vx-\vv y_t}^2).
$$
Therefore, we propose to choose $\vx_{t+1}$ by solving
\bb%b \label{equ:xtFsum}
\vx_{t+1}=\argmin_{\vx \in \X }  
\left\{ \dd \Phi(\vv y_t) \tt \vv x  ~~~~s.t.~~~~ \norms{\vx - \vv y_t }^2\leq \eta \xi_t^2 \right\},
%\norm{}
\ee%e 
where $\eta\in(0,1]$. The constraint $\norms{\vx - \vv y_t }^2\leq \eta \xi_t^2$ ensures that $\Fsum$ is sufficiently decreased, and the objective $\dd \Phi(\vv y_t) \tt \vv x$ 
%says that we try to decrease the value of diversity promoting loss and 
allows us to promote  diversity as much as possible given the constraint (it approximately minimizes $\Phi(\vv x)$ when the step size $\mu$ is  small).  
% This minimizes the linearization of $\Phi$ subject to that $\Fsum$ is sufficiently decreased. 
Here $\eta$ trade-offs the decreasing speed of $\Fsum$ \emph{v.s.} $\Phi$. 

%Large $\eta$ will sacrifice the descent rate of optimizing $\Fsum$.
% 
Solving the optimization yields that 
\bbb 
\vx_{t+1} =
\vv y_t 
-  \eta \frac{\norms{\vv y_t-\vv x_t}}{\norms{\dd \Phi(\vv y_t)}}
{\dd\Phi(\vv y_t)}. 
\eee 
 See Algorithm~\ref{alg:mainsum} for the main procedure.  
 It is clear from the derivation 
 that the algorithm monotonically decreases $\Fsum$ with $\Fsum(\vv x_{t+1}) \leq \Fsum(\vv x_t) - (1-\eta) \xi_t^2/(2\mu)$, and all particles converge to a local optimum of $f$ when the algorithm terminates.  
 
In this algorithm, the updates of the different particles $x_i$ are coupled together due to the minimization of $\Fsum$ and $\Phi$.  
Although $\Fsum$ decreases monotonically, 
the individual $f(x_i)$ does not necessarily decrease. %our method c%\red{  %On the other hand, n%Note that we do not force and require that every $f(x_i)$ decrease. %It is sufficient to ensure $\sum_{i=1}^m f(x_i) \leq m f^*$ decrease, because
%This allows 
In fact, the particles can exchange the loss with each other to gain better diversity: we may find that  
some particles temporarily increase the loss $f$ of to better decrease $\Phi$, while ensuring the overall $\Fsum$ decreases.  
\begin{algorithm}[t] 
\caption{Diversity-aware Gradient Descent ($\Fmax$)}
\begin{algorithmic} 
\label{alg:mainmax} 
\STATE \textbf{Goal}: Find a set of diversified local optima of $f(x)$.  \\
\textbf{Parameters}:  step size $\mu$, a repulsive coefficient $\eta$. 
\vspace{-.0\baselineskip}
\!\!\!\!\!\!\FOR{Iteration $t$}
% At the $t$-th iteration: %Take 
\vspace{-2.\baselineskip}
\STATE  \bb 
& x_{t+1,i} \longleftarrow x\ti - \mu \dd f(x\ti) - 
\frac{\xi\ti}{\norms{g\ti}} g\ti, 
%,  \\
%&\bar y_t = \frac{1}{m} \sum_{i=1}^m y_i\\
%&\alpha^*_t =  \frac{\sqrt{\sum_{i=1}^m \norm{y_i - x_i}^2}}{\sqrt{\sum_{i=1}^m \norm{y_i - \bar y}^2}} \\
%&x_{i}^{t+1} \gets y\ti +  \eta \alpha_t^* (y\ti - \bar y_t)
\vspace{-1.\baselineskip}
\ee 
where $g\ti = \dd_{y\ti}\Phi(\vv y_t)$, and $\vv y_t$ is defined in \eqref{equ:yt},  
\begingroup\makeatletter\def\f@size{7}\check@mathfonts
\def\maketag@@@#1{\hbox{\m@th\large\normalfont#1}}
$$
\xi\ti = 
\sqrt{2\mu \left ( (1-\eta)\max_{i\in[m]}f^\triangleleft\ti   
+\eta  \max_{i\in[m]} f(x\ti)  
%- \max_{i\in [m]} f^\triangleleft(x\ti) ) 
- f^\triangleleft(x\ti) \right)}, 
$$\endgroup
and $f^\triangleleft\ti = f(x\ti) - \frac{\mu}{2}\norms{\dd f(x\ti)}^2$.  
\ENDFOR
%$\alpha = \sart{
%\frac{\sum_{i=1}^m\norm{\dd f(x\ti)}^2}{
%\sum_{i=1}^m \norm{g_\ti}}$
\end{algorithmic}
\end{algorithm}

\subsection{$\Fmax$-Descent} 
\label{sec:fmax} 
We now derive a version of our algorithm that leverages $\Fmax(\vx) = \max_{i\in[m]} f(x_i)$ as the descending criterion in \eqref{equ:phiF}. 
This variant of algorithm focuses on descending $f$ on the worst-case particle and %gives more flexibility to the non-dominate particles to maximize diversity. 
%explore the loss landscape
%In the following, we show that 
hence provides larger flexibility 
for the non-dominate particles to maximize the diversity. 
Note that because $\Fmax$ is non-smooth, we can not directly use the method for $\Fsum$. A special consideration is needed to exploit the special structure of the $\max$ function. 

Similar to $\Fsum$, 
we assume $f$ is $1/\mu$-smooth. 
By applying \eqref{equ:smooth} on $x_i$ and taking the max over $i$, we get for $\forall \vv x$
%It is easy to see that 
$$
\Fmax(\vx) \leq 
%\Fmax(\vv x_t) 
% +  \frac{1}{2\mu} 
\max_{i\in[m]} 
\left\{ 
f^\triangleleft\ti 
+ \frac{1}{2\mu} 
%\left (
\norms{x_i- y\ti }^2  
%- \xi_i^2 \right)
%- \norms{x\ti - y\ti}^2 
\right\}\defeq \hat F^t_\textrm{max}(\vx),
$$\textrm
%where $\xi_i^2=\norms{ x\ti-y\ti}^2$. We want 
where we define 
 $$f^\triangleleft\ti = f(x\ti) - \frac{\mu}{2}\norms{\dd f(x\ti)}^2. 
$$ 
So here $\hat F^t_\textrm{max}(\vx)$ is the upper bound of $\Fmax(\vx)$ implied by the $1/\mu$ smoothness of $f$. 

Without considering $\Phi$, 
the minimum of the upper bound $\hat F^t_\textrm{max}(\vx)$ is obviously attained 
by $\vx = \vv y_t$ following vanilla gradient descent \eqref{equ:yt} on each particle $x\ti$. 
%Taking $\Phi$ into consideration, 
In this case, 
the descent of $\Fmax$ is upper bounded by 
\bb
\Fmax(\vv y_t) - \Fmax(\vv x_t)
& \leq \hat F^t_\textrm{max} (\vv y_t) - \Fmax(\vv x_t) \\ 
& = \max_{i\in[m]} f^\triangle _{t,i} - \max_{i\in[m]} f(x\ti) 
 \defeq - \delta_t. 
\ee 
 %that is, 
 In our algorithm, 
 we want to ensure that $\Fmax(\vv x_t)$ is decreased 
 by at least an amount of $\eta \delta_t$, 
  where $\eta\in(0,1)$ is a factor that quantifies how much we are willing to sacrifice the decreasing  of $\Fmax$ for promoting diversity.   

Therefore, we choose $\vx_{t+1}$ by solving 
%\begingroup\makeatletter\def\f@size{8}\check@mathfonts
%\def\maketag@@@#1{\hbox{\m@th\large\normalfont#1}}%
 \bb 
 \min_{\vx } \bigg \{\dd \Phi(\vv y_t) \tt \vx ~~~~s.t.~~~~
 \hat F^t_\textrm{max}(\vx) \leq \Fmax(\vx_t) - \eta \delta_t \bigg \}. 
 \ee 
 %$$\endgroup
 %where $\eta\in(0,1)$ denotes how much we are willing to sacrifice the decreasing speed of $\Fmax$. 
 This is equivalent to 
 $$
 \min_{\vx } \dd \Phi(\vv y_t) \tt \vx ~~~~s.t.~~~~
 \norms{x_i - y\ti }^2 \leq \xi\ti ^2,~~~
 \forall i \in[m], 
 %\hat F^t_\textrm{max}(\vx) \leq %\Fmax(\vx_t) - \eta \delta\ti,
 $$
 where 
 $
 \xi\ti^2 =  
 2\mu (\Fmax(\vx_t) - \eta \delta_t - f^\triangleleft \ti).
 $
 Solving this gives 
% Solving this gives 
$$
x_{t+1,i} = 
y\ti - \frac{\xi\ti}{\norms{\phi\ti}}  \phi\ti.
$$

See Algorithm~\ref{alg:mainmax} for details.  %for the algorithm. 
%As is 
It is clear from the derivation that we monotonically decrease $\Fmax$ with 
$\Fmax(\vv x_{t+1}) - \Fmax(\vv x_t) \leq \eta \delta_t$, and the algorithm terminates when $\Fmax$ reaches a local minimum of $\Fmax$. 
%
%
%\iffalse 

\begin{wrapfigure}{r}{0.2\textwidth}
\centering
% \begin{figure}
\vspace{-10pt}
%\begin{tabular}{c}
% \raisebox{1.8em}{\rotatebox{90}{Top-1 validation accuracy}}
\hspace{-1em}\includegraphics[width=0.2\textwidth]{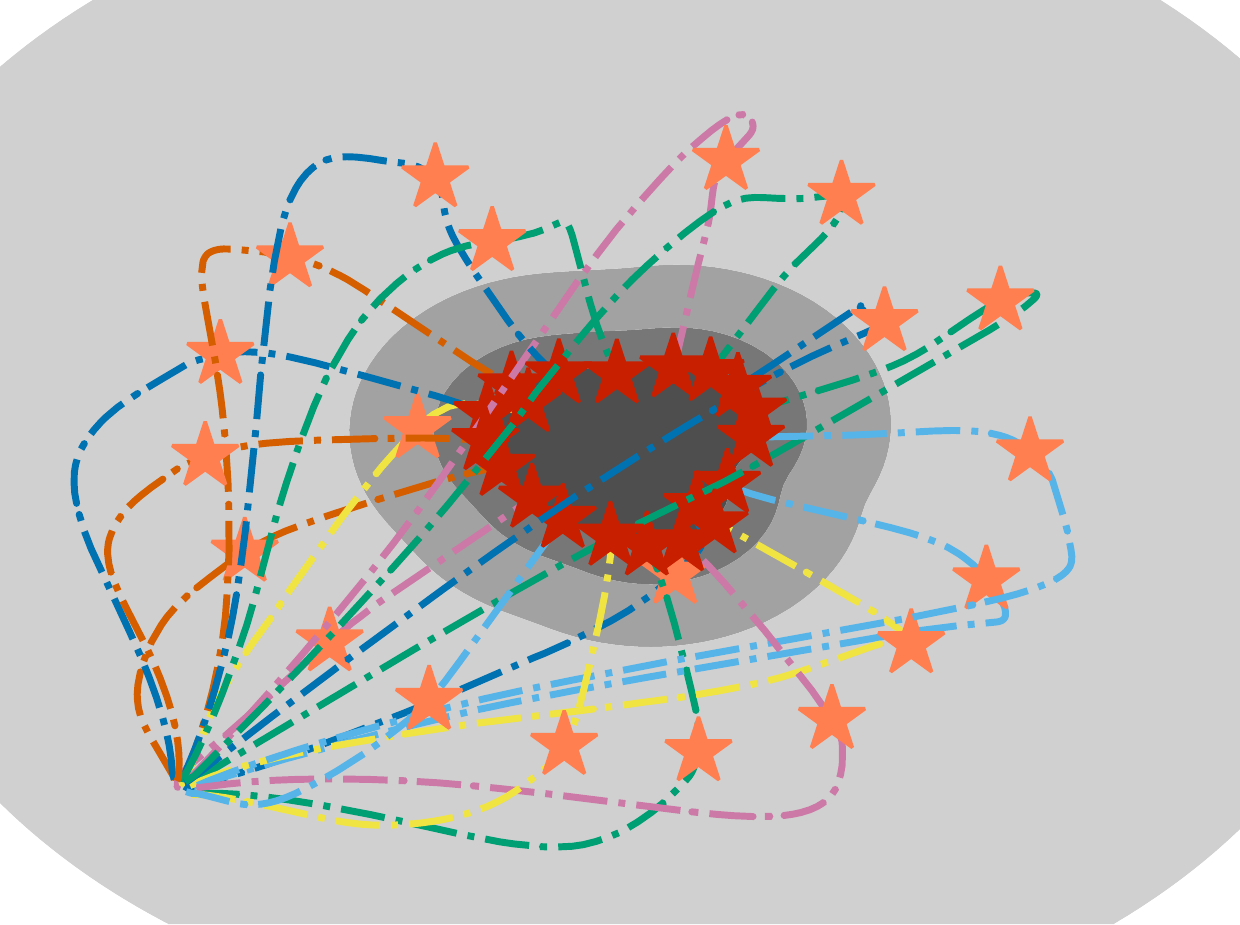} 
%\end{tabular}
\vspace{-10pt}
\end{wrapfigure}
\paragraph{Descending Along Contours}
An interesting feature of using $\Fmax$ 
is that the particles tend to lie on the contour lines of $f$ during the algorithm; see the right figure and  Fig.~\ref{fig:fmax} in Section \ref{sec:experiments}.
This is because the repulsive force from the diversity score tends to increase the loss $f$ of all the non-dominant particles, and as a result, makes their $f$ loss equal or close to the dominate particle as they descent on the landscape of $f$. 
%during the algorithm. 

%\paragraph{A Lock problem} 

%% file: tex/related_works.tex
\section{Related Works}

\iffalse
\citet{osa2020learning}: Learning the Solution Manifold in Optimization and
Its Application in Motion Planning; 
this work studies the problem of learning solution manifold of optimization, using an amortized approach. Considered applications in motion planning.  \url{https://arxiv.org/pdf/2007.12397.pdf} 
\fi

% \qq{I am not sure about this, but we need to cite gong in some way. what do you think?}

%\paragraph{Linear Combination and Sampling Methods} 
\textbf{Linear Combination Method } 
A naive way to trade-off two objectives to minimize their linear combination. For encouraging diversity, 
we consider  
%a straightforward approach is %we consider 
\bbb \label{equ:linear} 
\min_{\vx}(1-\alpha)\Fsum(\vx) + \alpha \Phi(\vx), 
\eee  
%$(1-\alpha)F + \alpha \Phi$, 
where $\alpha\in[0,1]$ is a fixed coefficient. 
The main drawback of this method is that we need to select $\alpha$ case-by-case, since the optimal choice of $\alpha$ depends on the relative scale of $F$ and $\Phi$, which may not be on the same scale; this is especially the case for Riesz $s$-energy with $s>0$ which goes to infinite when different points collapse together.  
In addition, if $\alpha >0$, the linear combination method necessarily scarifies loss $f$ for diversity. 
Note that \eqref{equ:linear} reduces to the naive multi-start approach if $\alpha = 0$ and $\{x_i\}$ starts from different random initialization. 
% and needs to sacrifice loss for diversity. 
%yielding a hard  trade-off between loss and diversity. 
%
%
%(unless $\alpha $ is approached to zero). 
% 
In comparison, our method does not require selecting $\alpha$ manually, 
and does not scarify loss $f$ for diversity by design.  
A key point that we want to make is that 
since the set of optimal solutions almost always consist of  multiple infinite number of points  in non-convex, deep learning, %, and many times 
%consists of an infinite number of points as the case when using overparameterized neural networks, 
it is feasible and desirable to find diverse points inside the optimum set, while gaining diversity for free. 

\textbf{Sampling-based methods}
provide another approach to finding diverse results. From the Gibbs variational principle, 
sampling can be viewed as solving \eqref{equ:linear} with $\Phi$ replaced by the entropy functional and $\alpha$ viewed as the temperature parameter. A notable example is Stein variational gradient descent \citep{liu2016stein}, which yields an interacting gradient-based update with repulsive force. 
Similar to the linear combination method, 
these methods require manually selecting a positive temperature $\alpha$ and yield a ``hard'' trade-off between loss and diversity. 

%\paragraph{Black-box optimization methods for Multimodal Objectives} 
\textbf{Population black-box optimization algorithms} 
have also been used to find diverse solutions. 
%provides 
%usually include diversity among multiple solutions during optimization.These 
Examples include 
genetic algorithms \citep[e.g.][]{lehman2011evolving, gomes2013evolution, lehman2011abandoning}, evolutionary algorithms \citep[e.g.][]{hansen2003reducing, Cully2015RobotsTC, flageat2020fast}, 
and Cross-entropy method (CEM) \citep{de2005tutorial}.
% and Bayesian optimization \citep[e.g.][]{kathuria2016batched, gonzalez2016batch, kathuria2016batched}.
%
% \red{They have been widely applied to robotics  \citep{Cully2015RobotsTC, osa2020multimodal}, reinforcement learning \citep{conti2017improving, parker2020effective}, and exploring the latent space of generation models \citep{liu2021fusedream, fontaine2021differentiable}, etc.} \qq{merge this sentence with the application sentence intro?}
%
A notable example is the 
\emph{MAP-Elites}  \citep{mouret2015illuminating}, 
%which discretizes the feature space into grids, 
which finds solutions in different grid cells 
of a feature space with different selection rules   \citep{sfikas2021monte, gravina2018quality}. 
%One of the key challenge problem for
The main bottleneck of these algorithms is the high computation cost. 
% due to the use of derivative-free optimization and the need of discretization of the feature space. To accelerate the optimization, 
Hence, a differentiable version MAP-Elites \citep{fontaine2021differentiable} was recently proposed to 
 speed up the computation. 
%leverages first-order gradient at each iteration in MAP-Elites to efficiently explore the joint range of quality and diversity. 

\textbf{Dynamic Barrier Gradient Descent} 
The $\Fsum$ method is similar to the dynamic barrier algorithm of \citet{gong2021automatic}, 
which provides a general algorithm for solving bilevel optimization of form $\min_x f(x)$ s.t. $x \in \argmin g$. A key difference is that we use the quadratic constraint $\norm{\vx- \vv y_t}^2 \leq\eta \xi_t^2$ to constraint the update direction,
while \citet{gong2021automatic} uses the inner product constraint of form $(\vv x-\vv x_t) \tt (\vv y_t-\vv x_t) \geq \eta\xi_t^2$. 
Using the quadratic constraint provides a stronger control to descent $f$, and ensures that the algorithm converges when it is on the optimum set.
The $\Fmax$ method, on the other hand, is very different from existing approaches by leveraging the special structure of the $\max$ function. 
%This is especially helpful when $\Phi_s$ singular when } 

\iffalse
\paragraph{Applications} % of Diversity-aware Optimization} % is Robotics, Reinforcement Learning, Image Generation, etc.}
\qq{I feel we give too much space on cully and Osa here. Maybe just a citation in the list, and review more broadly (have a balanced list of different areas: robotics, reinforcement learning, image generation, etc.) }
Diversity-aware optimizations have been widely applied in robotics. 
%We provide 
Here we only give a few examples. 
%For example, 
\citet{Cully2015RobotsTC} uses MAP-Elites \citep{mouret2015illuminating} to create  behavior-performance map and uses Bayesian optimization \citep{Cully2015RobotsTC} to enable online adaptation for the robots. 
As the result, it reveals successful adaptations for a legged robot injured in different ways, including damaged, broken, missing legs, etc. 
\citet{osa2020multimodal} aims to determine multiple trajectories  that correspond to the different modes of the cost function for motion planning. 
% It studies the problem of learning solution manifold of optimization using an amortized approach. 
During optimization, it divides the manifold of the trajectory plan and determines solutions in each region.
These black-box optimization methods have also been widely used in reinforcement learning \citep{conti2017improving, parker2020effective} and exploring the latent space of generation models \citep{liu2021fusedream, fontaine2021differentiable}.
\fi

%% file: tex/experiment.tex
\section{Experiments}
\label{sec:experiments}

We first examine and understand our method in some toy examples, 
and then apply $\Fsum$ and $\Fmax$ to more difficult deep learning applications: text-to-image~\citep{liu2021fusedream, ramesh2021dalle}, text-to-mesh~\citep{michel2021text2mesh}, molecular conformation generation~\citep{shi2021learning} and neural network ensemble.
In all these cases, 
we verify and confirm that our method can serve as a plug-in module and 
% for many different problems.
can obtain 1) visually more diverse examples, and
2) a better trade-off between main loss (e.g. cross-entropy loss, quality score) and diversity without tuning co-efficient.
We set $s=0$ for Reisz $s$-energy distance if there is no special instructions, set $\eta=0.5$ for $\Fmax$ and report $-\Phi$ score to measure diversity.
We report the average score over 3 trials for each experiment.

\subsection{Toy Examples}

We verify our proposed methods on toy test functions, study the impact of the Riesz $s$-energy, 
the trade-off of the target function and diversity term,
and the trade-off of using $\Fsum$ vs. $\Fmax$. 
We adopt gradient descent with a constant learning rate $5\times10^{-4}$ and 1,000 iterations.

\begin{figure}[tbh]
\centering
\scalebox{0.96}{
\begin{tabular}{ccc}
% \hspace{-25pt}
\includegraphics[width=.14\textwidth]{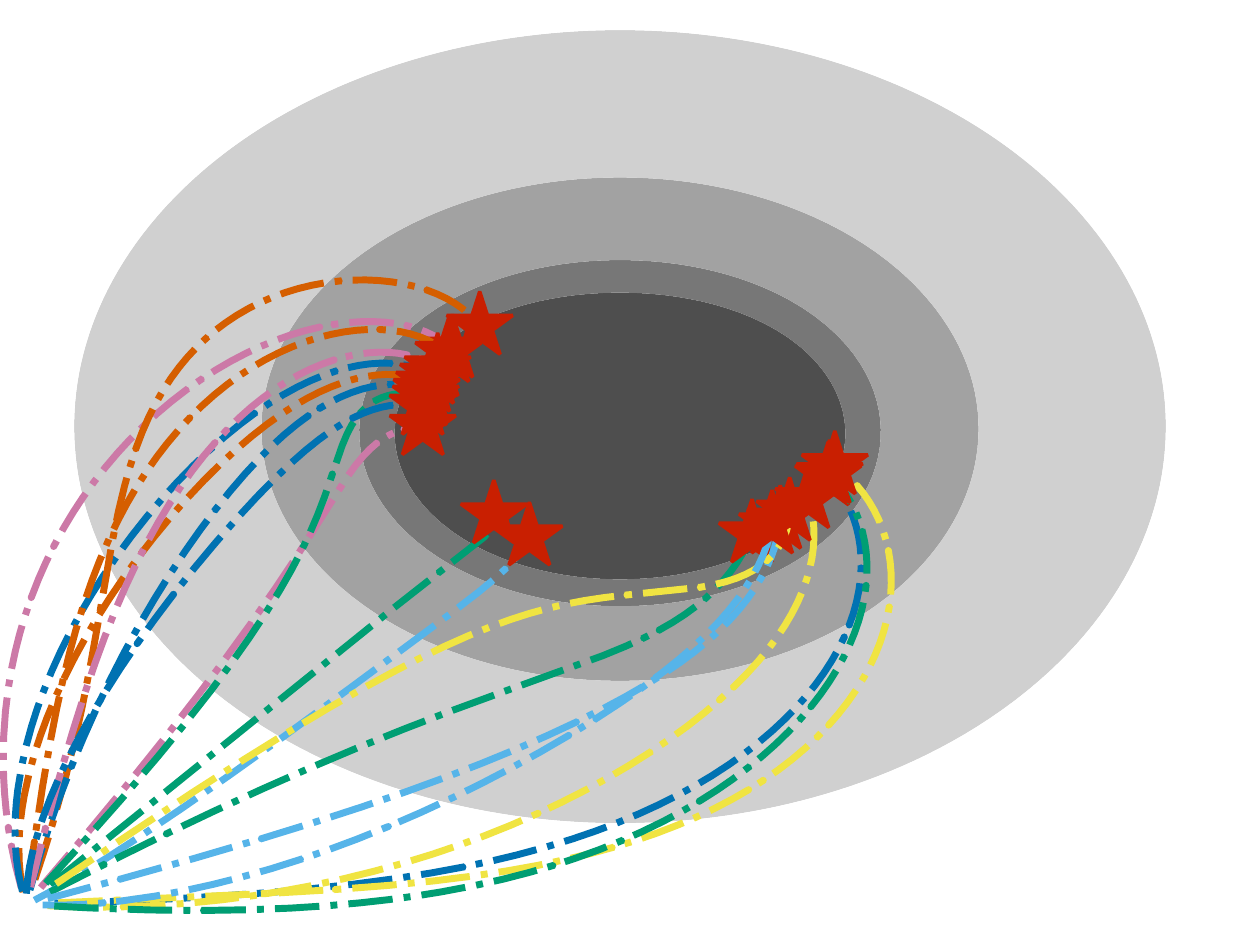} & 
\hspace{-45pt}
\includegraphics[width=.14\textwidth]{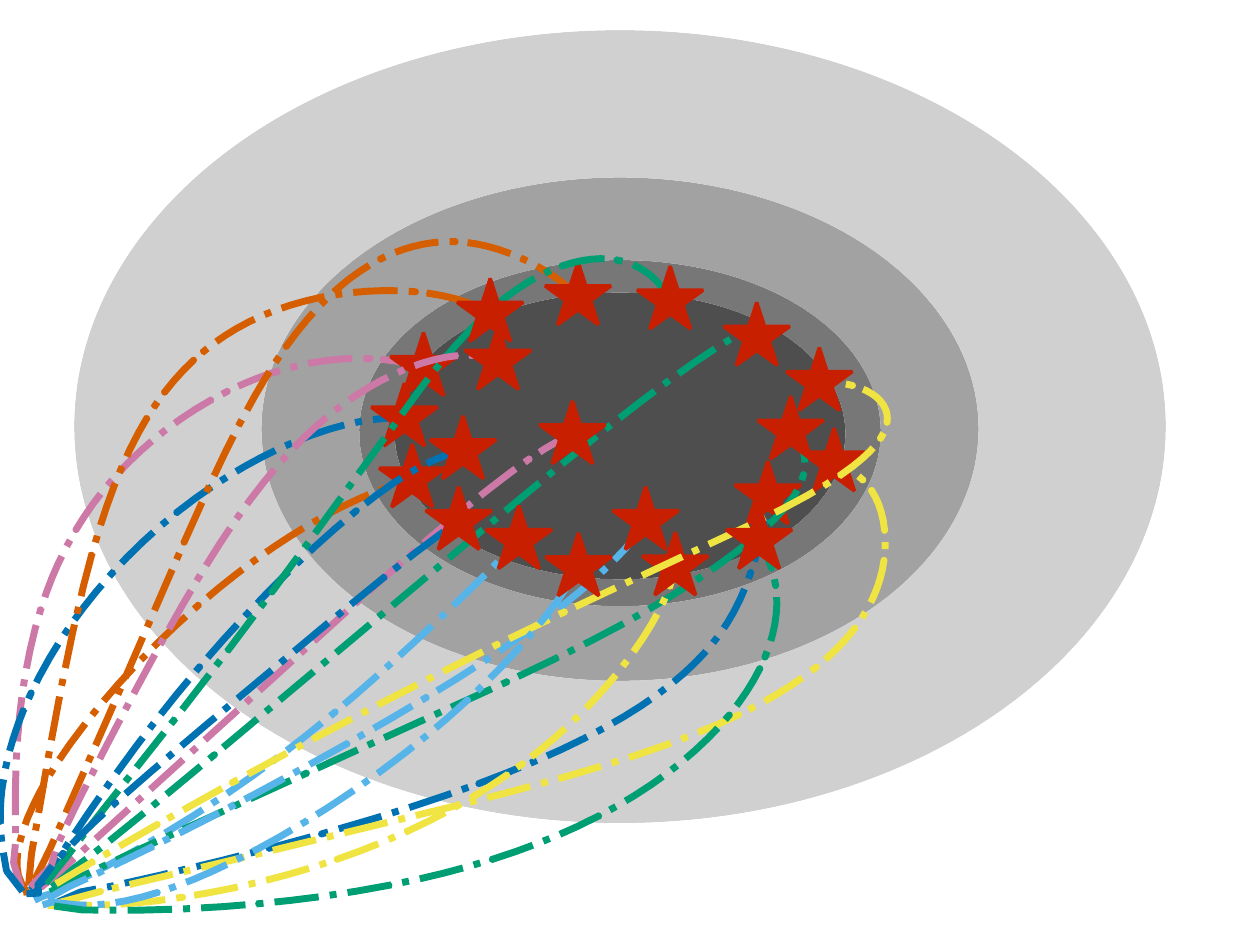} & 
\hspace{-45pt}
\includegraphics[width=.14\textwidth]{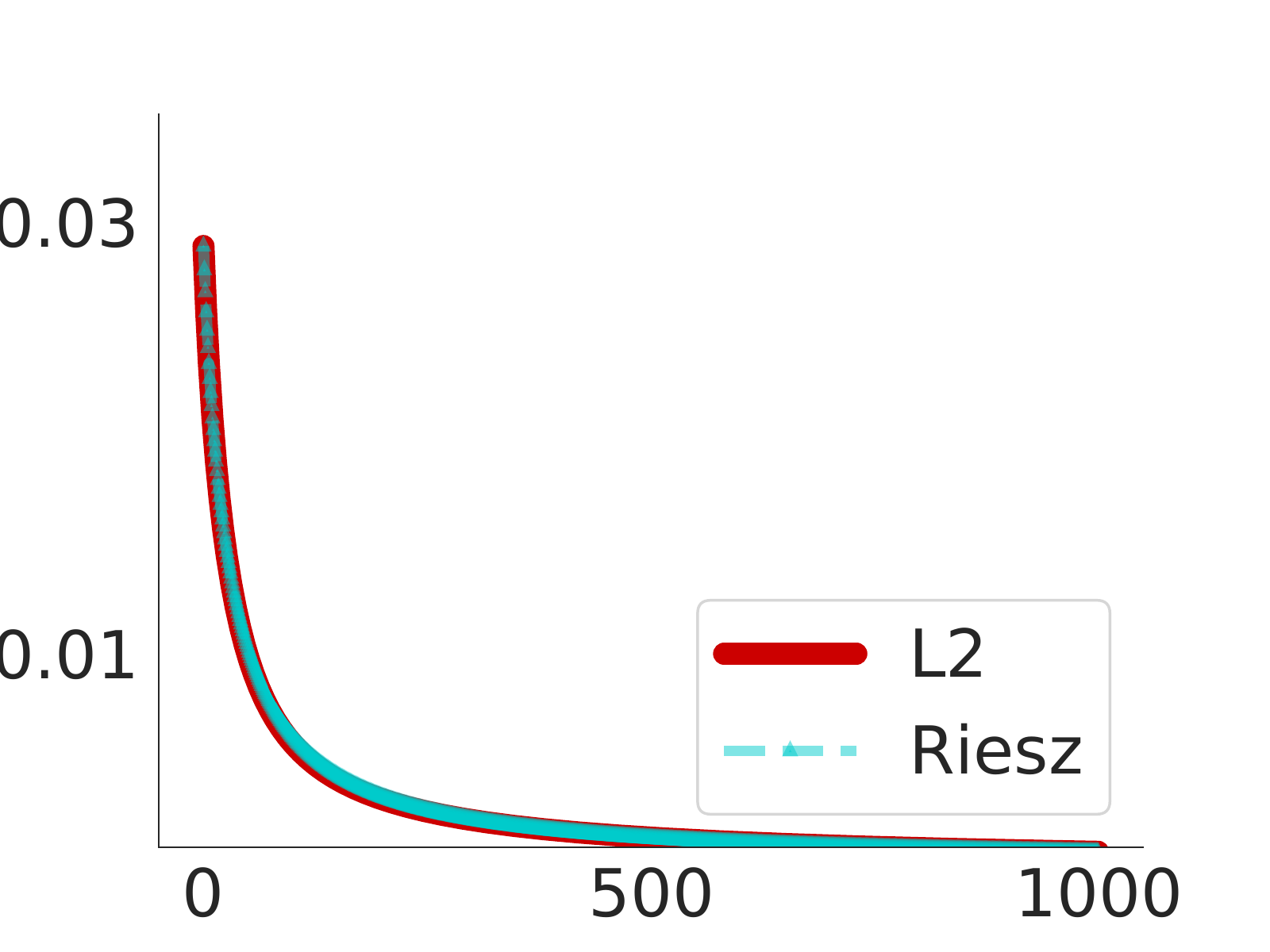}
\vspace{-5pt}
\\
(a) %$\ell_2$ Distance
$s = -2$
~~~ & (b) 
$s=d-2=0$ 
%Riesz  $s$-energy
~~~~~~~~~~~~ & \hspace{-35pt} (c) Gradient Norm~~ \\ 

\end{tabular}}
\vspace{-5pt}
% 1.4467868 -0.49062848
% 1.0383276 -0.14535126
\caption{
Results on a 2D ($d=2$) toy example with $\Fsum$. We test two different choices of $s$ in Riesz energy, including $s=-2$ (variance) and $s=d-2=0$ (logarithm energy). 
We can see that the logarithm energy yields more uniformly distributed points. 
%We notice that Reisz $s$-energy achieves a better loss as well as a larger diversity value.
}
\label{fig:distance}
\vspace{-5pt}
\end{figure}

\textbf{\emph{Q1: How does the choice of $s$ in Riesz energy influence the result?} } One typical measure of diversity is 
the variance, which corresponds to $s=-2$ in Riesz energy.  However, ~\ref{fig:distance}(a) shows that 
it tends to yield many points that are close to very close to each other. This is because variance does not place a strong penalty on 
%close distance 
close points once the overall averaged pairwise distance is large. 
On the other hand, using Riesz $s$-energy with $s\geq 0$ tends to yield more uniformly distributed points (Figure~\ref{fig:distance}(b)). This is because when $s\geq 0$, Riesz $s$-energy places a strong penalty on the points that are very close to each other. 
Figure \ref{fig:distance}(c) shows zero gradient norm and indicates the convergence of both cases.

%

%We replace the $\ell_2$ distance with Riesz $s$-energy function to force diversity. 
%To understand how the energy distance function impacts the training process and result, we visualize the optimization trajectories and the optimization results in Figure \ref{fig:distance}. 
% For this 2-dimension test fuction, we set $s=0$.
%We further notice that Riesz $s$-energy function forces the particles uniformly distribute and makes all the particles visually diverse, while Maximizing $\ell_2$ distance does not make particles diverse from each other.
% while our proposed energy distance can make all the particles visually diverse. 
% while $\ell_2$ distance comes to the results where many particles are close to each other. 
% The behaviour of $\ell_2$ distance is easy to understand. Intuitively, consider there is three particles, $x, y, z$, given that $\|x-y\|_2^2 + \|y-z\|_2^2 + \|x-z\|_2^2 \leq 2 \|x-z\|_2^2$, $x=y$ makes the $\ell_2$ distance larger than other cases. 

\begin{figure}[h]
\centering
\scalebox{0.85}{
\begin{tabular}{c}
\hspace{-25pt}
\includegraphics[width=.54\textwidth]{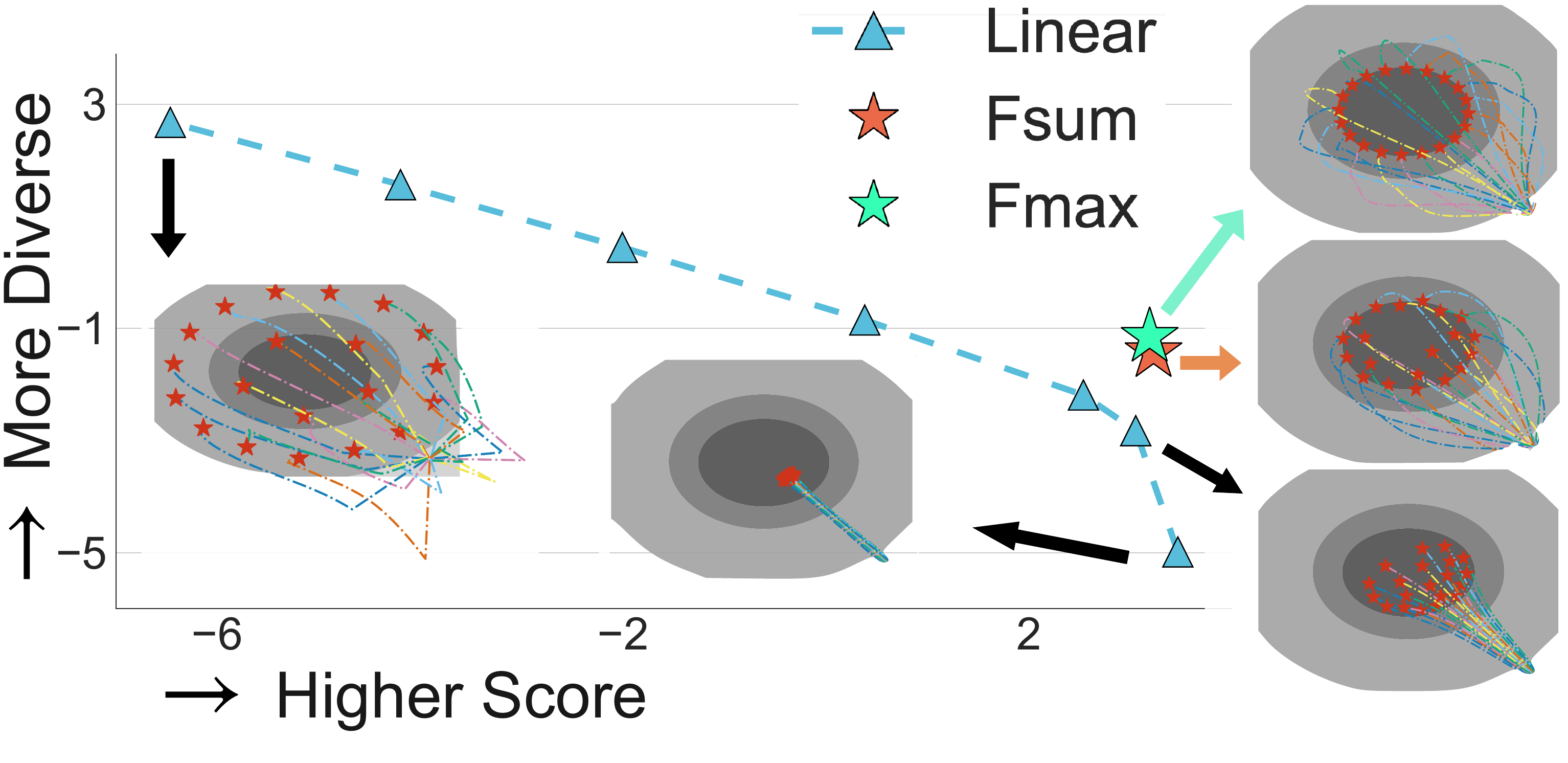} 
\\

\end{tabular}}
\vspace{-5pt}
\caption{
The loss $F$ vs. diversity score $\Phi$ by $\Fsum$ and $\Fmax$ on a 2D example (the orange and green star), and by minimizing $(1-\alpha) \Fsum + \alpha \Phi$ with different values of $\alpha \in [0,1]$ (blue triangles). We can see that $\Fsum$ achieves a better trade-off.
% \qq{the contour figures are too small; make them larger  (the plot can be smaller and scaled to more square like)}
} 
\label{fig:harmless}
\vspace{-5pt}
\end{figure}

\textbf{
\emph{Q2: How does our method compare with the linear combination method?} 
}
\iffalse
The main drawback of the linear combination is that the case-by-case selection of $\alpha$.
The optimal choice of $\alpha$ depends on the relative scale of $F$ and $\Phi$, 
and they are not in the same scale especially for Riesz $s$-energy which goes to infinite when different points collapse together. 
In addition, for $\alpha >0$, the linear combination necessarily scarifies $F$ for diversity (unless $\alpha $ is approached to zero). Our method, however,does not scarify $F$.  
\fi
Varying $\alpha$ in the linear combination (e.g. 0, $10^{-4}$, $10^{-3}$, 0.01, 0.1, 0.5, 1), we can trace a (locally optimal) Pareto front of loss $F$ and diversity $\Phi$.  
In Figure \ref{fig:harmless}, we find that our method can achieve strictly better results than the Pareto front of the linear combination method.
%We notice that our method (red star) obtains better Pareto front compared to the blue curve. 
Compared to the linear combination (the blue triangles), we notice that 
% $\Fsum$ has an impact on the optimization process and results. 
in the early iterations of the trajectory, $\Fsum$ and $\Fmax$ introduce a larger diversity penalty and makes the particles more diverse. % in the beginning. 

\begin{figure}[h]
\centering
\hspace{-2em}
\scalebox{0.96}{
\begin{tabular}{ccccc}
% \hspace{-25pt}
% \includegraphics[width=.2\textwidth]{figs/fmax_log_1000.pdf} & 
% \hspace{-25pt}
% \includegraphics[width=.2\textwidth]{figs/fsum_log_1000.pdf} & 
% \hspace{-25pt}
% \includegraphics[width=.2\textwidth]{figs/fmaxsum_norm.pdf} 
% \vspace{-10pt}\\
\hspace{-10pt}
\raisebox{1.75em}{\scriptsize(a)~~~~~} & 
\hspace{-15pt}
\includegraphics[width=.14\textwidth]{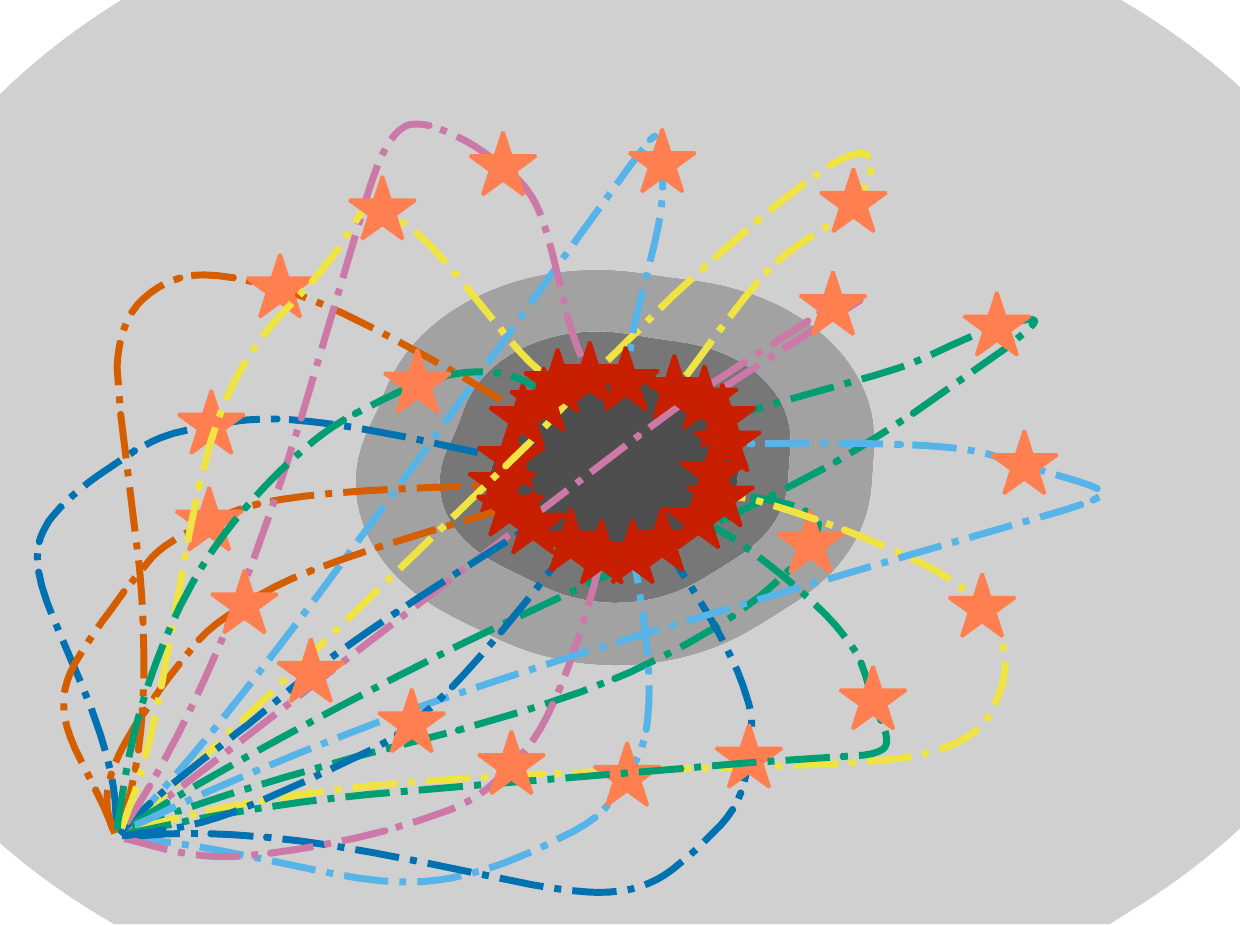} & 
\hspace{-15pt}
\includegraphics[width=.11\textwidth]{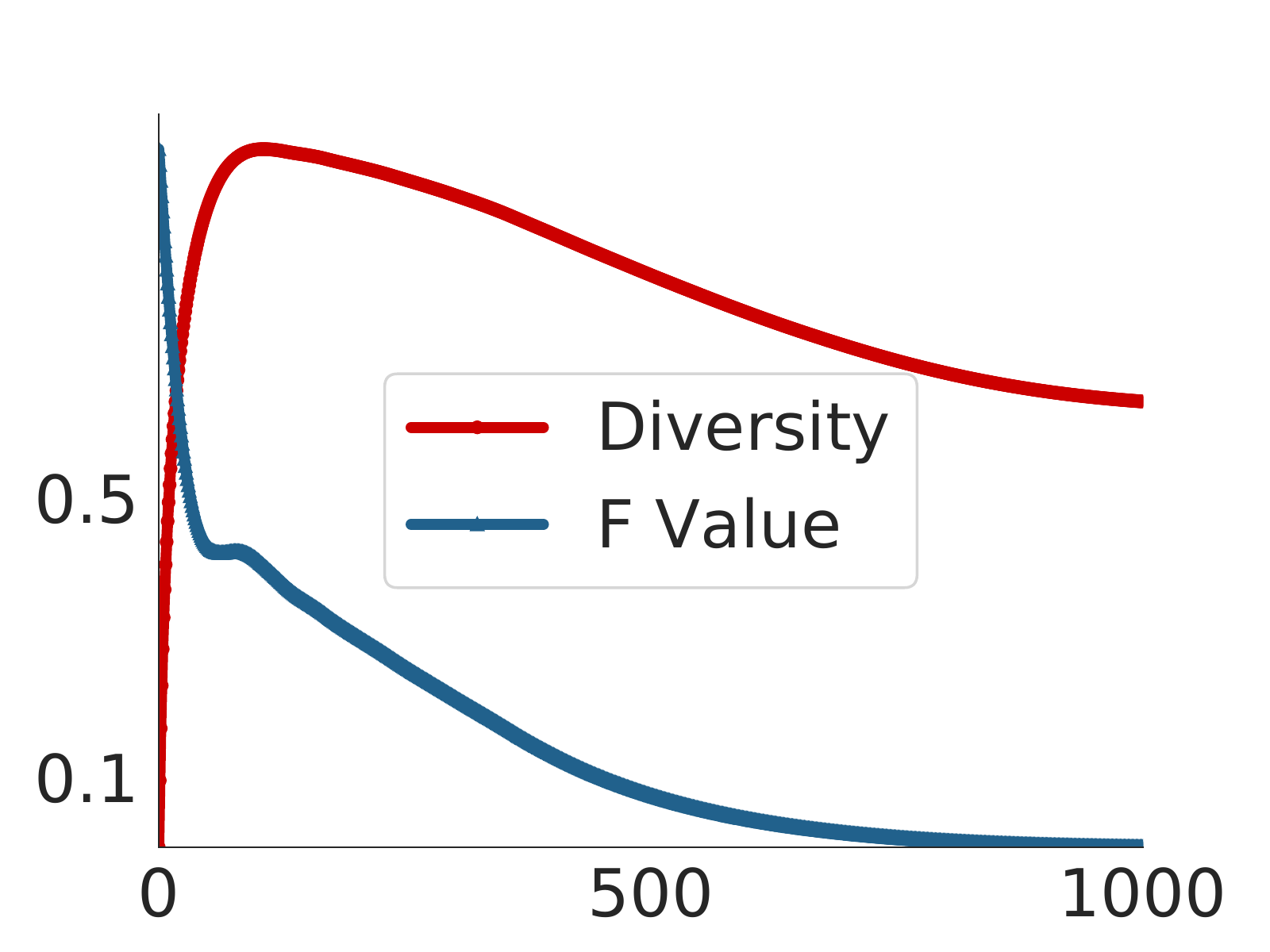}
& 
\hspace{-20pt}
\includegraphics[width=.14\textwidth]{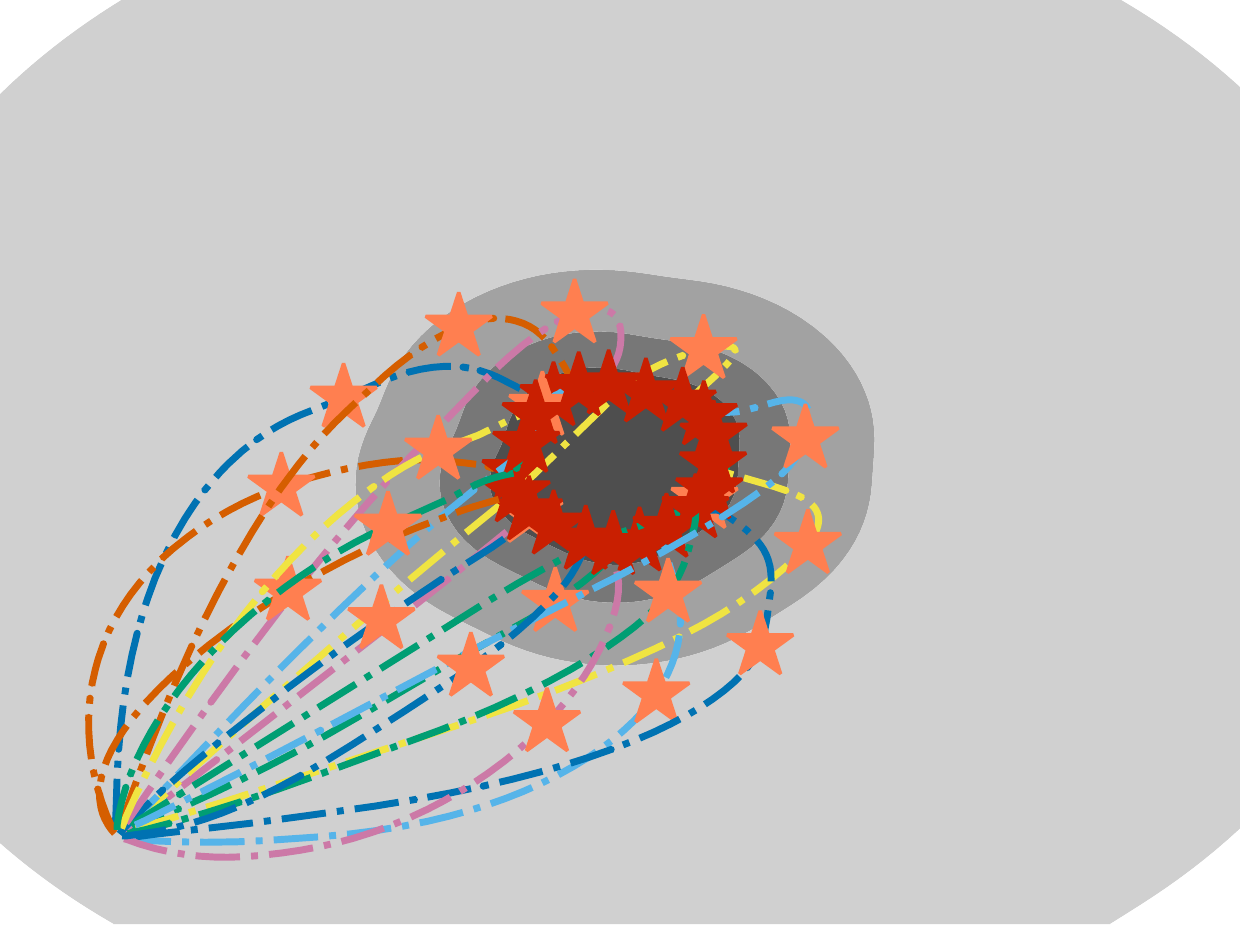}
& 
\hspace{-15pt}
\includegraphics[width=.11\textwidth]{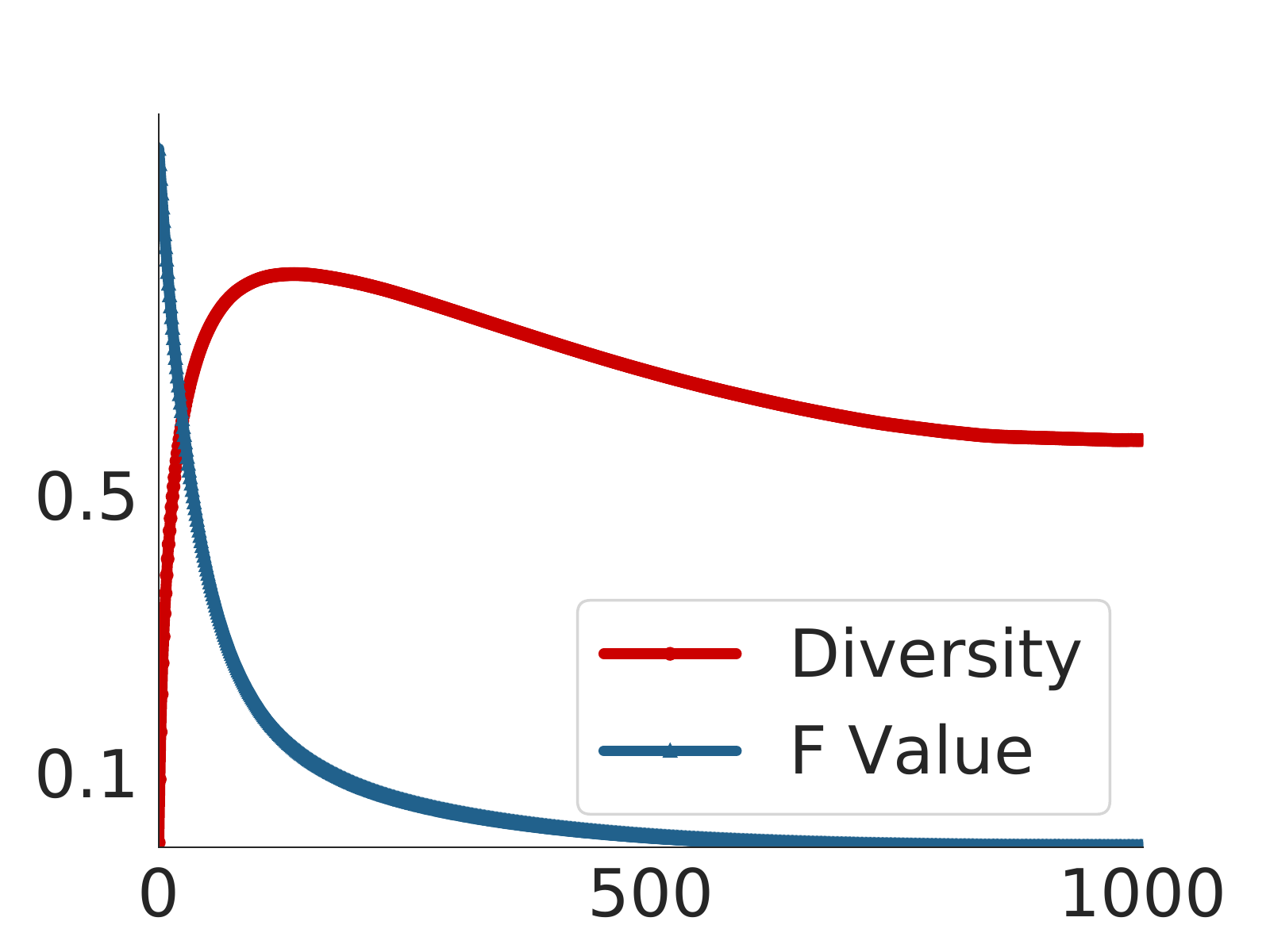}
\vspace{-5pt}
\\
\hspace{-10pt}
\raisebox{1.75em}{\scriptsize(b)~~~~~} & 
\hspace{-15pt}
\includegraphics[width=.14\textwidth]{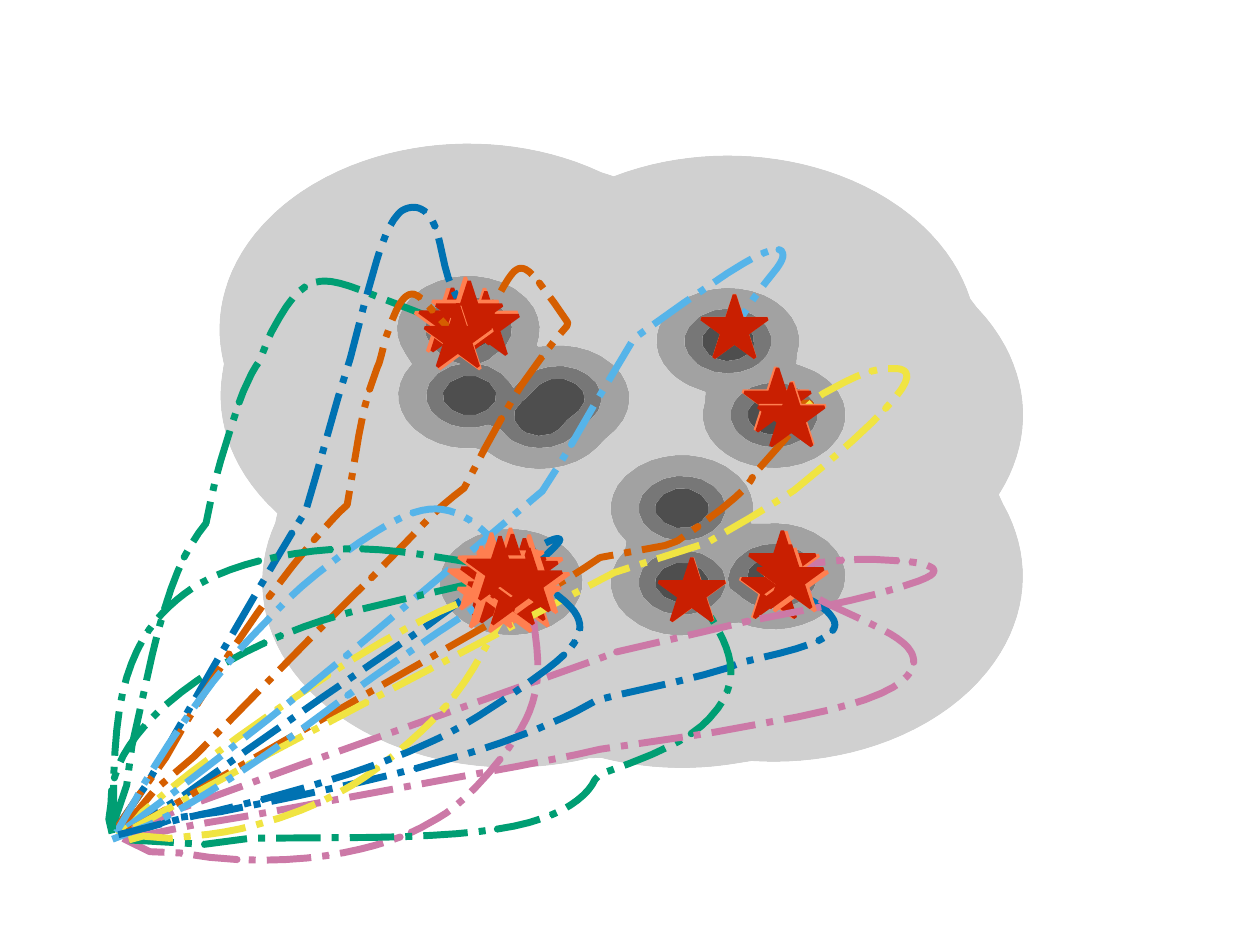} & 
\hspace{-15pt}
\includegraphics[width=.11\textwidth]{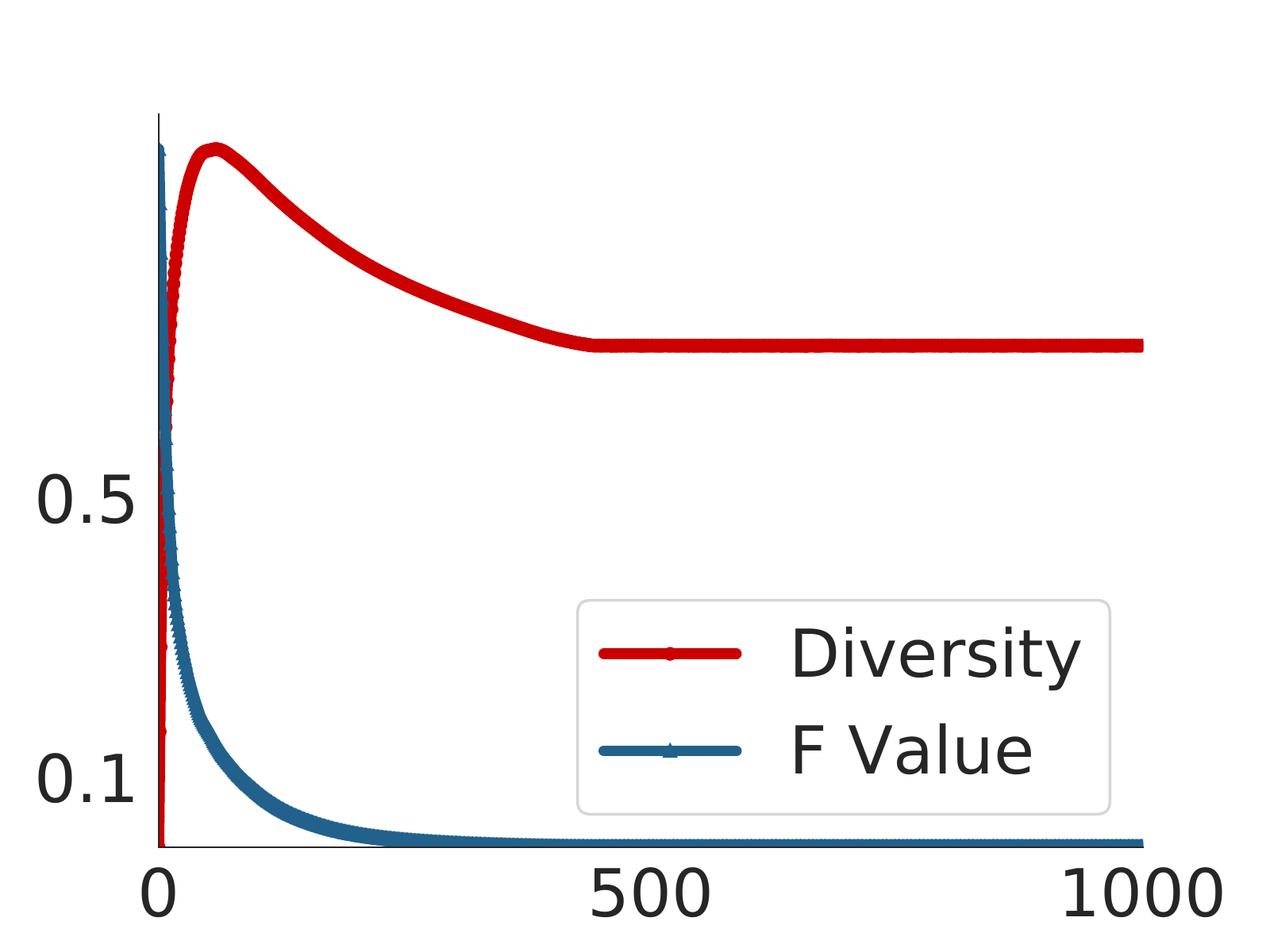}
& 
\hspace{-20pt}
\includegraphics[width=.14\textwidth]{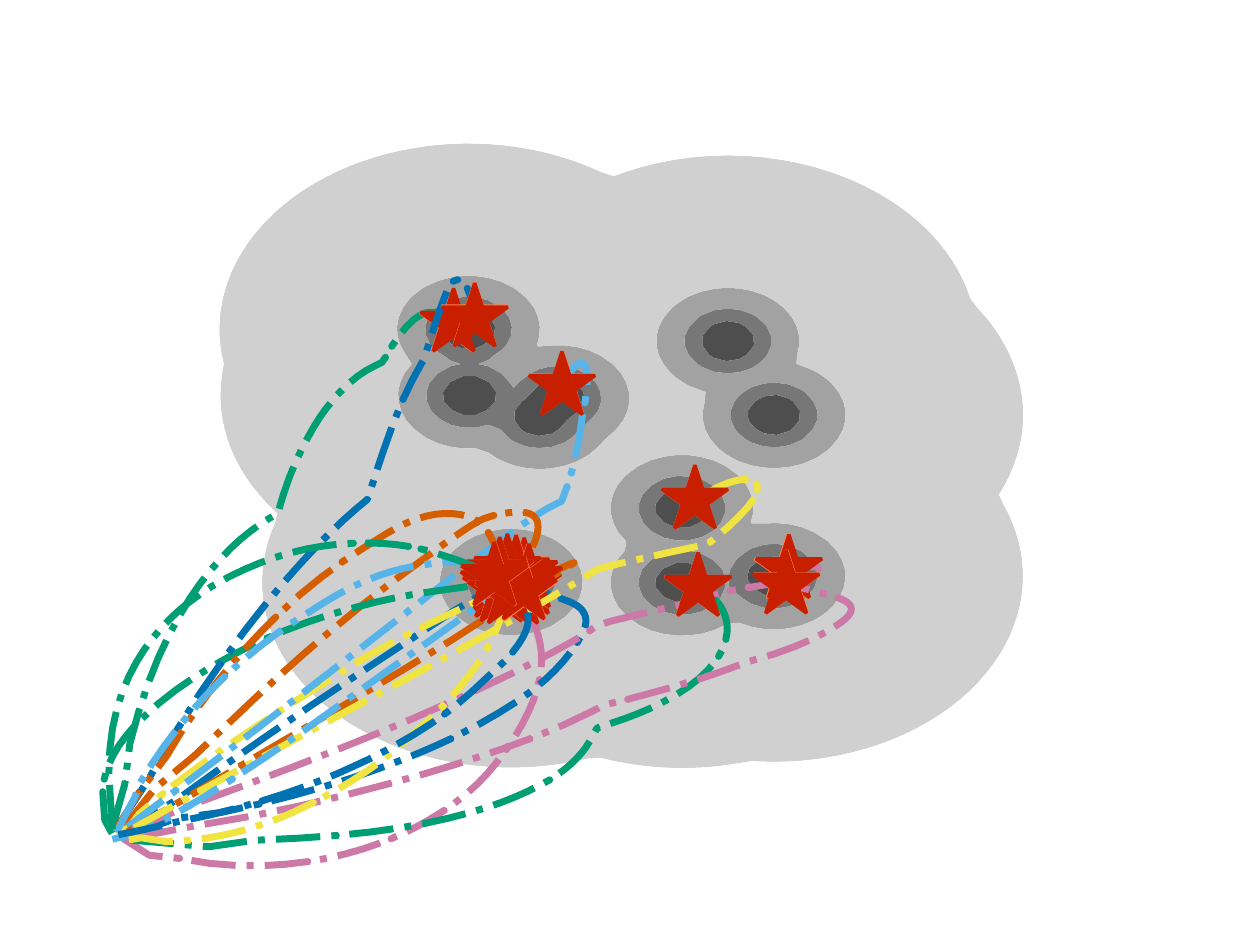}
& 
\hspace{-15pt}
\includegraphics[width=.11\textwidth]{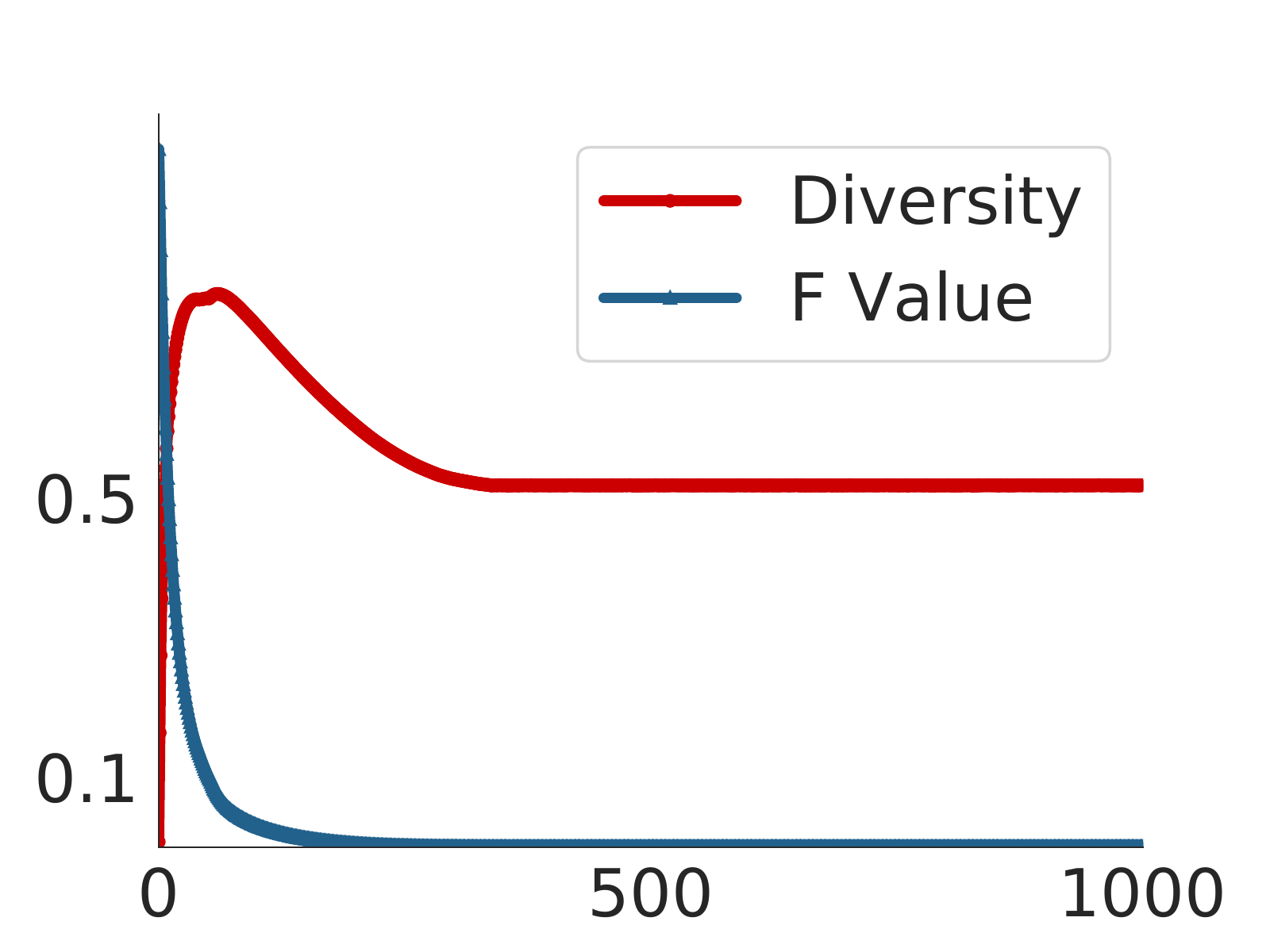}
\vspace{-5pt} \\
\hspace{-10pt}
\raisebox{1.75em}{\scriptsize(c)~~~~} & 
\hspace{-15pt}
\includegraphics[width=.14\textwidth]{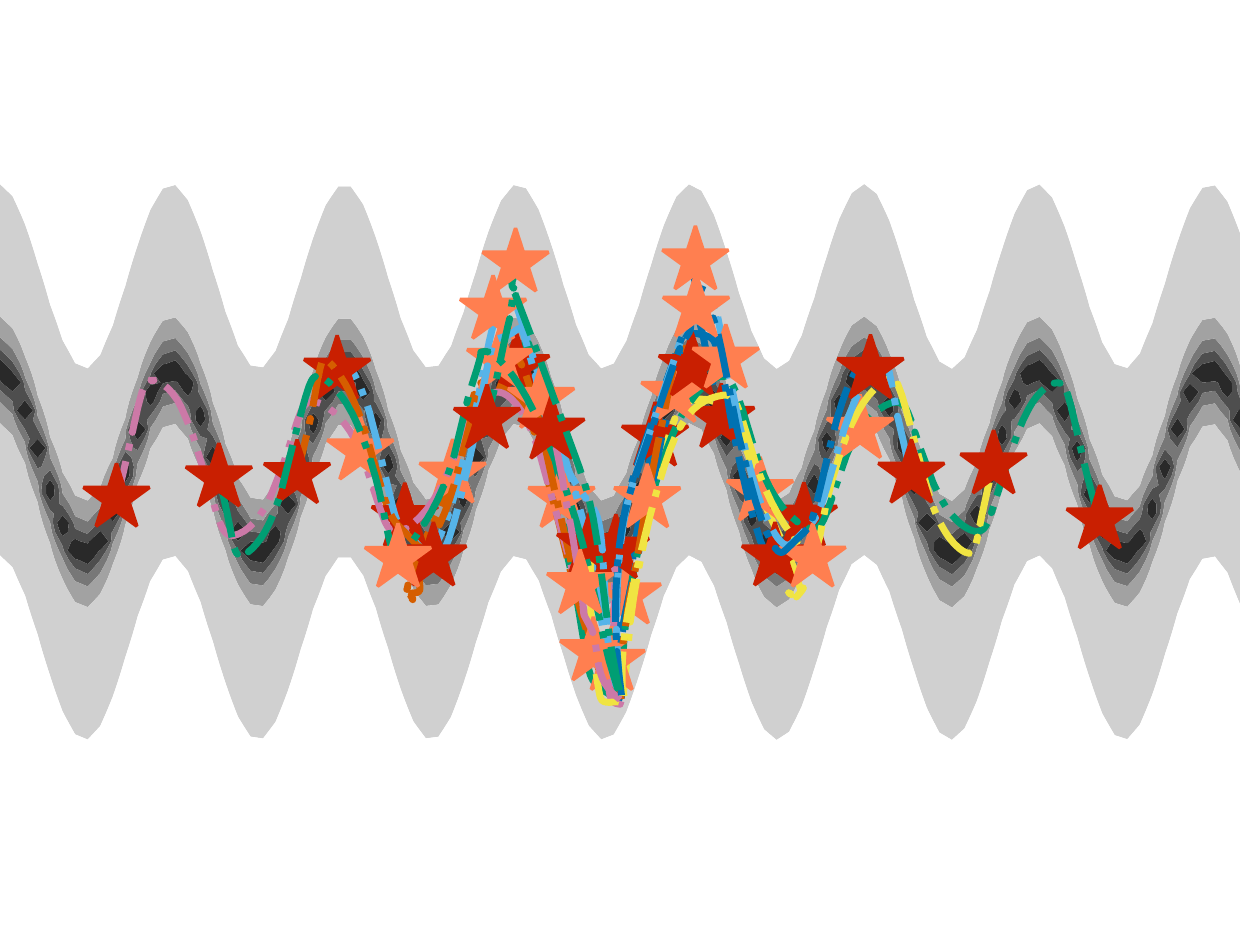} & 
\hspace{-15pt}
\includegraphics[width=.11\textwidth]{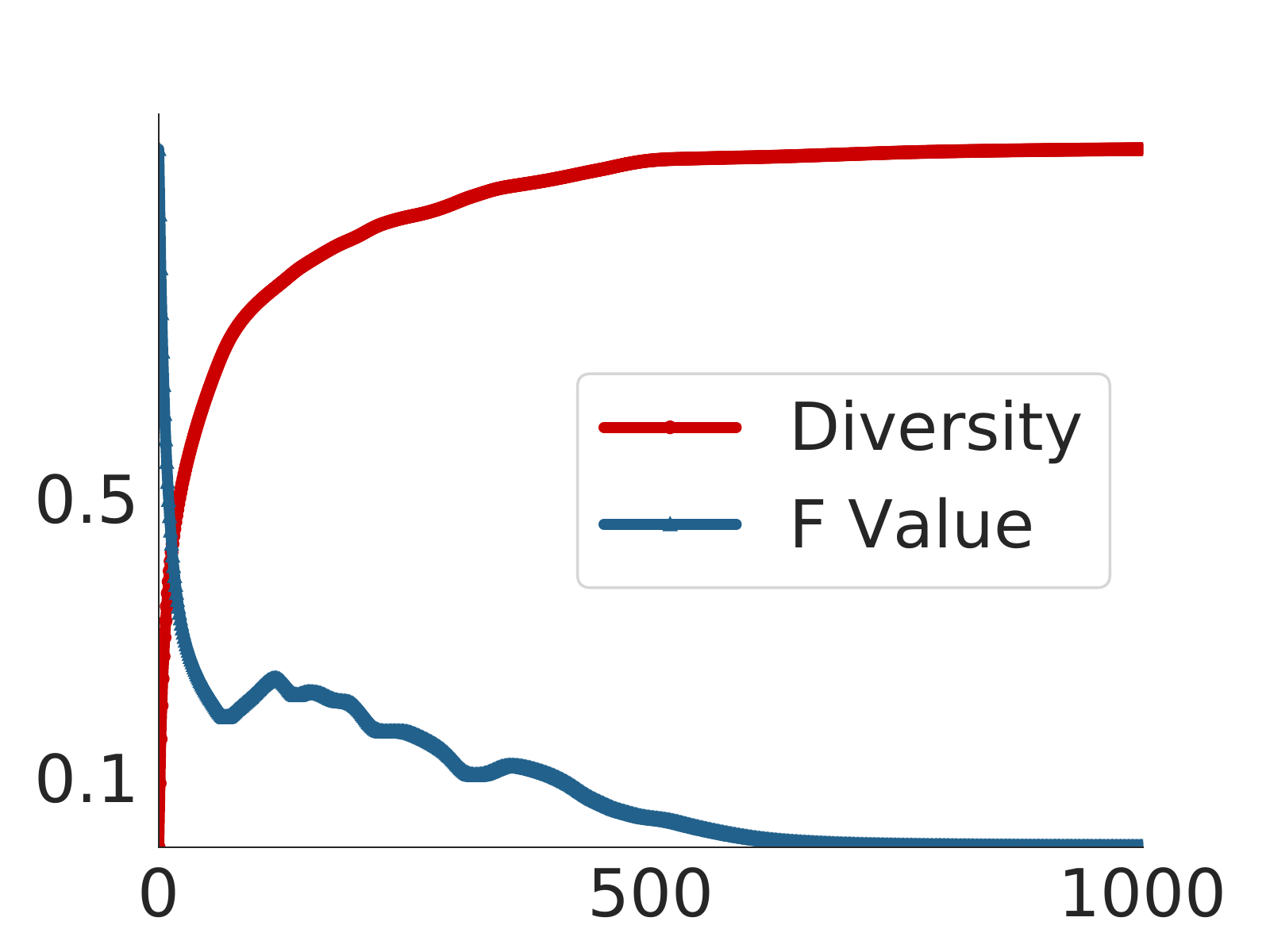}
& 
\hspace{-20pt}
\includegraphics[width=.14\textwidth]{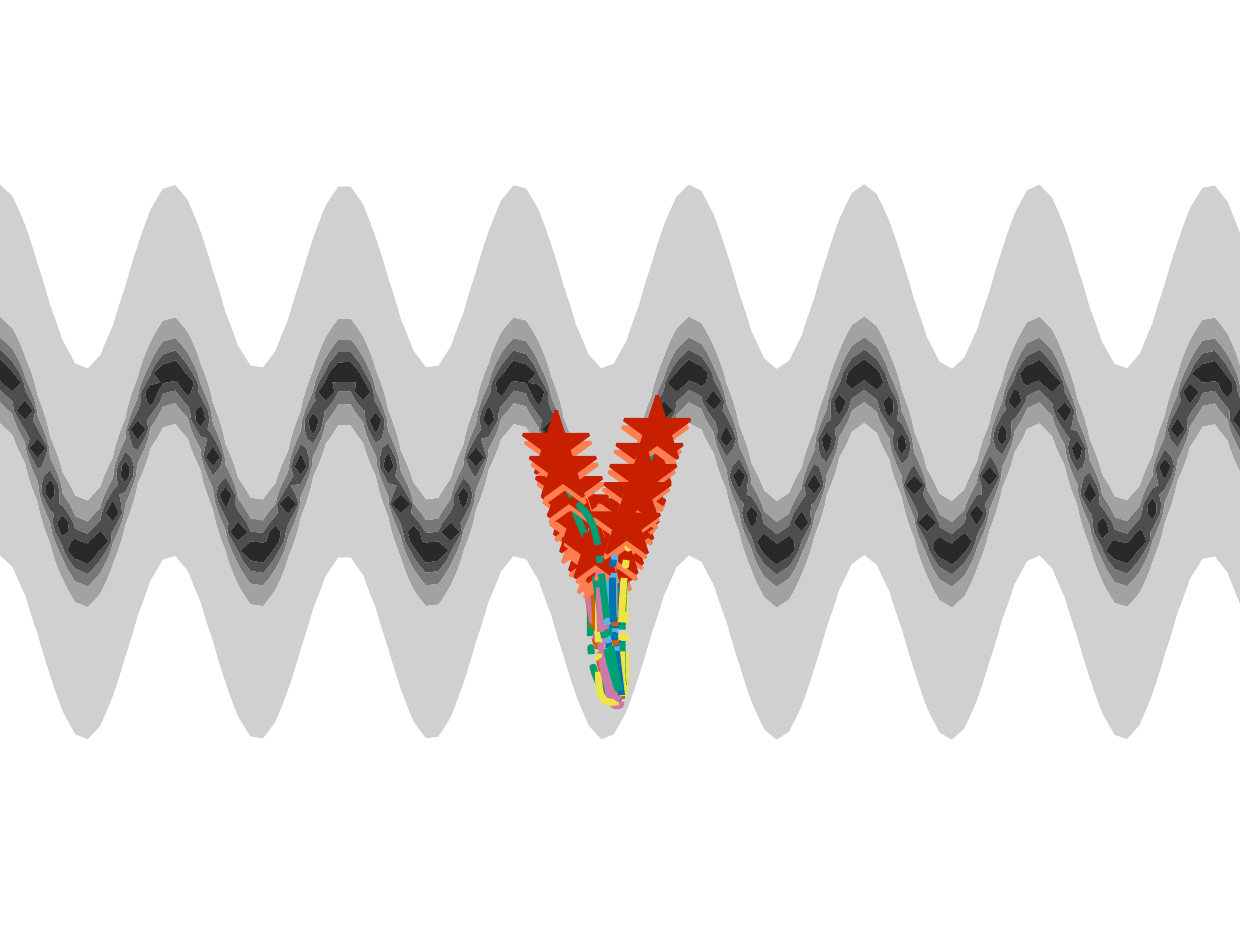}
& 
\hspace{-15pt}
\includegraphics[width=.11\textwidth]{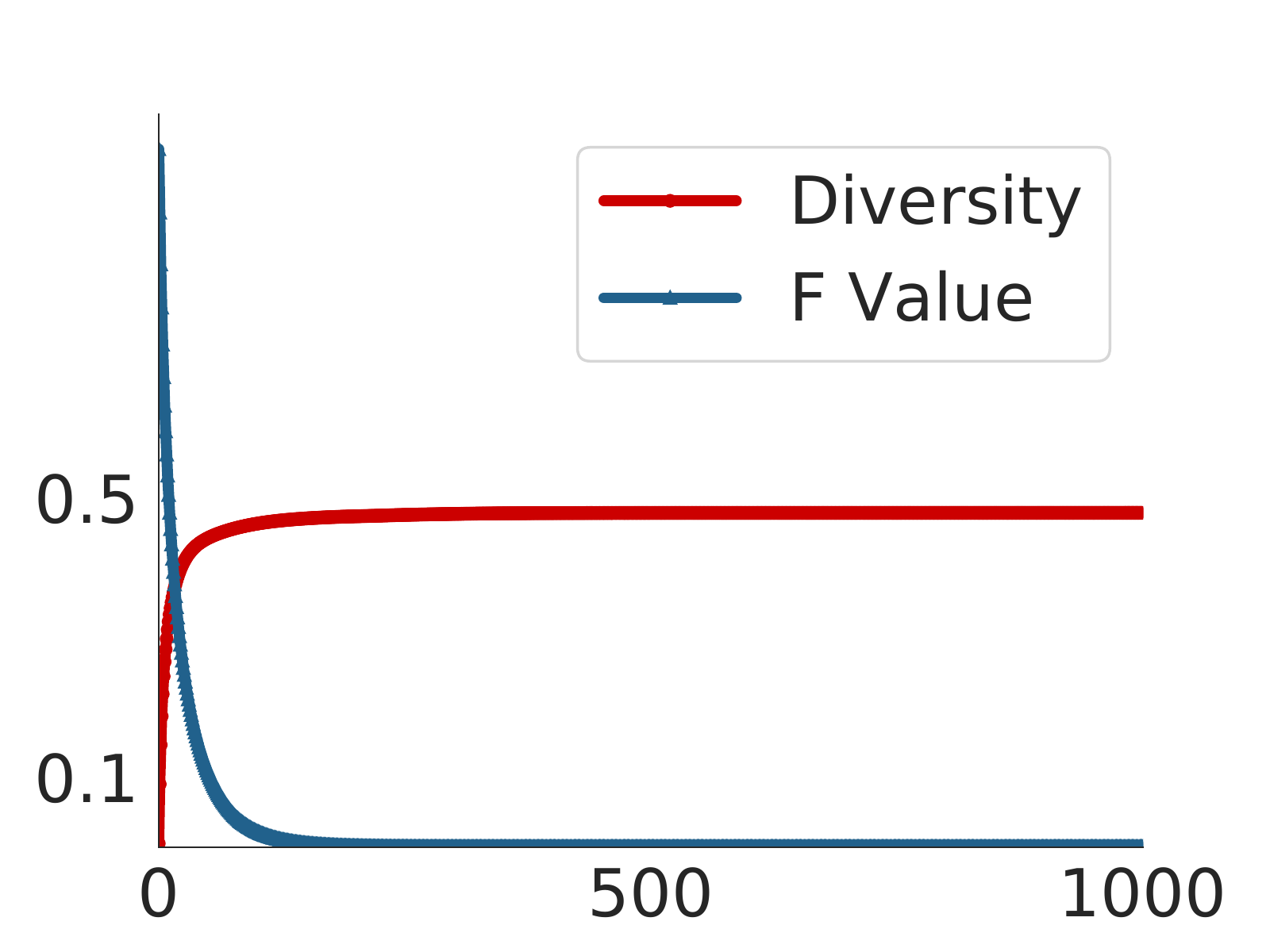}
\vspace{-5pt}

\\
& \multicolumn{2}{c}{$\Fmax$} & \multicolumn{2}{c}{$\Fsum$}  \\
% & $\Fmax$~~~~~~~~~~ & $\Fsum$~~~~~~~~~~& \hspace{-25pt} {\small $\Fmax$: $F$ \emph{v.s.} Diversity}\\ 
\end{tabular}}
\vspace{-10pt}
\caption{
Results on toy examples with $\Fmax$ and $\Fsum$.
The red and orange stars show the 1000th-iteration and  200th-iteration results.
The curve shows the value of the target function and diversity term during optimization.
%\qq{slightly separate the two panels?}
%\qq{the white space on the left can be removed for larger image}
% We observe that $\Fmax$ makes the particles lie on the contour lines during optimization.
}
\label{fig:fmax}
\vspace{-5pt}
\end{figure}

\begin{figure*}[t]
\centering
\scalebox{.92}{
\begin{tabular}{c}
\includegraphics[width=1.07\textwidth]{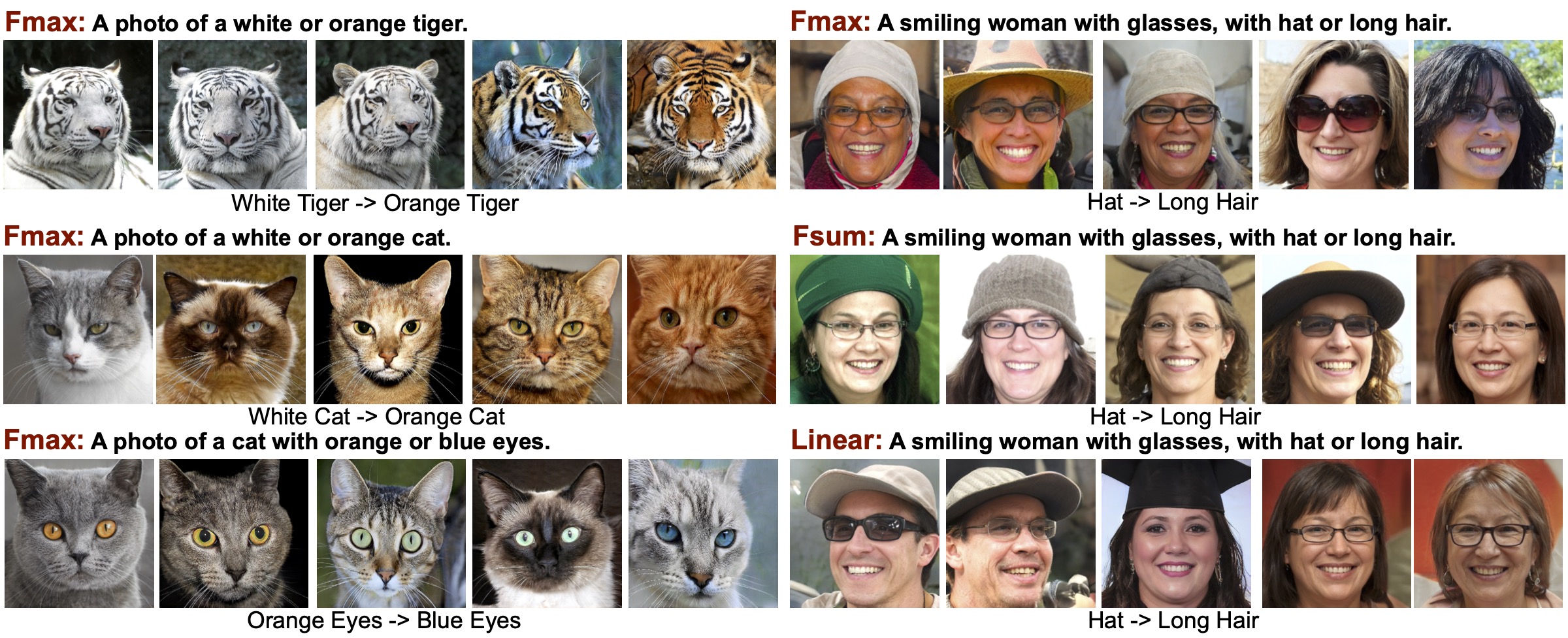}
\\
\end{tabular}}
% \vspace{-15pt}
\caption{
The left column: images generated by $\Fmax$  based on different input text.
Right column: images generated by $\Fmax$, $\Fsum$ and the linear combination \eqref{equ:linear} ($\alpha=0.5$) from the same input text.
% \red{We use the StyleGAN-v2.}\qq{maybe just remove. we introduce in text}
The text above each image denotes the text prompt $\mathcal{T}$, and the text under each image displays the $\mathcal{T}_1 \rightarrow \mathcal{T}_2$ used to define the diversity score in \eqref{eq:psit1}.  Better viewed when zoomed in; see Appendix for more examples. 
}
\label{fig:halfhalf_examples}
\vspace{-15pt}
\end{figure*}

     %3) \emph{When do we need $\Fmax$ instead of $\Fsum$?}
%\end{enumerate}

\paragraph{\emph{Q3: What is difference of using  $\Fmax$ vs. $\Fsum$? }}
\iffalse
In order to capture all the local modes in a multi-mode landscape, 
people usually re-initialize the start point and re-start gradient descent many times \citep{eriksson2019scalable}. 
However, it is usually hard to decide where to initialize since we do not know  the properties of the landscape and the location of each mode. 
Our $\Fmax$ method can be a useful alternative for this kind of problem.
\fi
As suggested in Section~\ref{sec:main}, 
$\Fmax$ is expected to generate more diverse examples, with the trade-off of yielding slower convergence and potentially worse loss value. 
%As demonstrated in Figure \ref{fig:fmax}, $\Fmax$ can capture multiple modes even if it starts from similar initialization.
To verify this, 
we test $\Fmax$ and $\Fsum$ in three different kinds of test functions shown in Figure~\ref{fig:fmax}, whose optimal set is a connected manifold (Figure~\ref{fig:fmax}(a)), multiple isolated modes (Figure~\ref{fig:fmax}(b)), and a curve (Figure~\ref{fig:fmax}(c)).
We observe that: 
1) Compared to $\Fsum$, $\Fmax$ tends to place a larger diversity penalty, especially in the early phase of the optimization. 
%beginning of the optimization. 
2) 
In $\Fmax$, the particles tend to lie on the contour lines during the optimization (Figure \ref{fig:fmax}(a) left). 
%
%3)Figure \ref{fig:fmax}(b) displays that $\Fmax$ captures more local modes even if the mode is far from the initialization, while $\Fsum$ converges to the local modes close to the initialization.
%4) Figure\ref{fig:fmax}(c) says that $\Fmax$ can achieve more diverse results than $\Fsum$ once we need to explore the optimum set.
% During the optimization process, sometimes the main loss of some particles becomes worse than their previous steps. 

% The main weakness of $\Fmax$ is its slower convergence. As displayed in Figure \ref{fig:fmax}, when optimizing for 200 iterations (the pink stars), $\Fsum$ almost converges to the local optima, while $\Fmax$ is still far away from converges. 
% Therefore, when the computation resources is limited in practice, $\Fsum$ is superior to $\Fmax$. When we target capturing more local mode, $\Fmax$ can be a better choice. 

\subsection{Latent Space Exploration for Image Generation}

\paragraph{Text-Controlled Zero-shot Image Generation}
We apply our method to text-controlled image generation.  %has long been an open problem. Researchers focus
%the problem of
% generating high-quality, realistic-looking images given text prompt. %with different kinds of the given information, e.g. depth estimation, segmentation mask, text prompt.
%Recently, with the success of CLIP \citep{radford2021clip}, researchers achieve improvements in controlling the image generation with text prompts. 
% Among these works, 
We base our method on
FuseDream \citep{liu2021fusedream}, 
a training-free text-to-image generator that works by combining the power of
pre-trained BigGAN \citep{brock2018biggan} and the CLIP model \citep{radford2021clip}.
% by searching in the latent space of the pre-trained BigGAN to optimize the CLIP score of the generated image and text prompt. 

\textbf{Basic Setup } A pre-trained GAN model $\mathcal I = g(x)$ is a neural network that takes a latent vector $x\in \RR^d$ and outputs an image $\mathcal I$. 
The CLIP model \citep{radford2021clip}  provides a score $\texttt{CLIP}(\mathcal T, \mathcal I)$ for how an image $\mathcal I$ is related to a text prompt $\mathcal T$.  
We use the augmented clip score $s_{\texttt{AugCLIP}}(\mathcal T, \mathcal I)$ from \citet{liu2021fusedream} which improves the robustness by introducing random augmentation on images. 

\citet{liu2021fusedream} generates an image $\mathcal I = g(x)$ for a given text $\mathcal T$ by solving $\max_{x} s_{\texttt{AugCLIP}}(\mathcal T, g(x))$. 
We introduce diversity on top of \citet{liu2021fusedream}. 
Given text prompt $\mathcal T$, our goal is to find a diversified set of  images $\mathcal I = g(x_i)$, $\forall i\in[m]$ that maximize the $s_{\texttt{AugCLIP}}$ score,  where $\vv x = \{x_1, \ldots, x_m\}$ is obtained by 
%$v_{\mathcal T}=f_\textrm{text}(\mathcal T)$, $v_{\mathcal I} = f_\textrm{image}(\mathcal I)$ embeds  a text $\mathcal T$ and image $\mathcal I$ into continuous vectors $v_{\mathcal T}$ and $v_{\mathcal I}$, such that the 
%a joint text-image embedding, such that, for a pair of  
%Mark the optimizable variables in the GAN latent space $x$, the pre-trained GAN model $g$, and CLIP model $\{f_\textrm{text}, f_\textrm{image}\}$,   %The optimization problem  for a query text prompt $\mathcal{T}$ is defined as 
\begin{align}
\label{eq:image_bilevel}
 \min_{\vv x} \Phi(\vv x),  ~~s.t.~~ \vv x \subseteq \arg\max_{\vv x'} s_{\texttt{AugCLIP}}(\mathcal{T}, g(\vv x')),
\end{align}
where we define the diversity score 
by $\Phi(\vx) = \Phi_s(\psi(x_1),\ldots, \psi(x_m))$, and  
 $\psi$ is a neural network that an input image to a semantic space. In particular, we use 
 %$g(x)$
 %maps each $x_i$ to the feature space, 
%where the $s_{\texttt{AugCLIP}}$ is defined as $s_{\texttt{AugCLIP}}(\mathcal{T}, \mathcal{I}) = \E_{\mathcal I' \sim \pi(\cdot|\mathcal I)} \left [\texttt{CLIP}(\mathcal{T},~ \mathcal{I}') \right].$
%$\texttt{CLIP}$ denotes the score return by the CLIP model,  $\mathcal I'$ is a random perturbation of the input image $\mathcal I$ drawn from a distribution $\pi(\cdot|\mathcal I)$ of candidate data augmentations. We refer the readers to \citep{liu2021fusedream} for more details.
%In our case, 
% in order to visualize the difference between different variants of our methods and baselines, 
%we create the $\Phi$ space with additional query texts following the setting of \emph{Quality Diversity} \citep{cully2017quality}. 
%With $\vv x \defeq \{x_i\}_{x=1}^m$ and
%$\Phi(\vv x) = R_s(\psi(\vv x))$ where $\psi$ maps each $x_i$ to the feature space, 
\begin{align}
    \label{eq:psit1}
\psi(x) = \bigg [ s_{\texttt{AugCLIP}} \left (\mathcal{T}_1, g(x) \right ), s_{\texttt{AugCLIP}}\left (\mathcal{T}_2, g(x) \right ) \bigg ], 
\end{align}
where $\mathcal{T}_1$ and $\mathcal{T}_2$ are two text that specify the semantic directions along which we want to diversify. 
For example, by taking $\mathcal T_1$ = \emph{`White~Tiger'} and $\mathcal T_2$ = \emph{`Orange~Tiger'} (Figure 4), 
we can find images of tigers that interprets from white to orange.
%given two additional text $\mathcal{T}_1$ and $\mathcal{T}_2$, we define $\psi$ as Denote the feature dimension is 2 and therefore we set $s=0$ for $R_s(\cdot)$.
% We target on getting a set of images by optimizing a set of particles $\vv x$, which match $\mathcal{T}$ and vary from maximizing $\mathcal{T}_1$ to maximizing $\mathcal{T}_2$.
%\paragraph{Experiment Settings} We examine the performance of different methods from two perspectives. First, we visualize the optimized images and compare the diversity and quality from a qualitative perspective. Secondly, we directly check the value of the main loss and diversity term from a quantitative perspective.

%In all these cases, 
For the experiments, 
we use the Adam \citep{kingma2014adam} optimizer with constant $5\times10^{-3}$ learning rate.
% and the Adam \citep{kingma2014adam} optimizer.
For BigGAN, the optimizable variable $x$ contains two parts, a feature vector in the GAN latent space and a class vector representing the 1K ImageNet classes.
We set the number of iterations to 500 following \citet{liu2021fusedream}. 
For StyleGAN-v2 \citep{Karras2019stylegan2}, the optimizable variable $x$ is a feature vector in the GAN latent space. 
Because StyleGAN-v2 generates images in higher resolution (e.g.  1024$\times$1024),
we set the number of iterations to 50 to save computation cost. 

\paragraph{Qualitative Analysis} 
Figure~\ref{fig:halfhalf_examples} shows 
examples of images generated from our $\Fmax$, $\Fsum$ and the linear combination ($\alpha=0.5$) when using StyleGAN-v2 trained on FFHQ \citep{karras2019style} and AFHQ \citep{choi2020stargan}.
We can see that the images generated by ours are both \emph{high quality}, semantically related to the prompt $\mathcal T$ (high $s_{\texttt{AugCLIP}}$), 
and \emph{well-diversified} along semantic direction specified by $\mathcal T_1$ and $\mathcal T_2$ (low $\Phi(\vx)$). 
For example, for $\mathcal T$=\emph{`a smiling woman with glasses'}, $\mathcal T_1$=\emph{'hat'}, $\mathcal T_2$=\emph{'long hair'}, our method yields images with diverse  hats and hair lengths. 
As a comparison, the $\alpha=0.5$ linear combination fails to generate `woman' or `glasses' in some cases.

\begin{table*}[th]
    \centering
    \scalebox{0.75}{
    \begin{tabular}{l|c|cc|cc|cc|cc|cc}
    \hline
    \multirow{2}{*}{} & \multirow{2}{*}{$\mathcal{T}$} & \multirow{2}{*}{$\mathcal{T}_1$} & \multirow{2}{*}{$\mathcal{T}_2$} & \multicolumn{2}{c|}{Linear $\alpha=0$} & \multicolumn{2}{c|}{Linear $\alpha=0.5$} & \multicolumn{2}{c|}{$\Fsum$} & \multicolumn{2}{c}{$\Fmax$} \\
    \cline{5-12}
    &&& &  Sc $\uparrow$ & Div $\uparrow$ & Sc $\uparrow$ & Div $\uparrow$ & Sc $\uparrow$ & Div $\uparrow$ & Sc $\uparrow$ & Div $\uparrow$\\ 
    \hline 
    \!\!\!Test 1&\emph{A painting of an either blue or red dog.} & \emph{blue} & \emph{red} & \bf{0.34} & -3.78 & 0.30 & -3.63 & \bf{0.34} & -3.64 & 0.31 & \bf{-3.60} \\
    \hline
    \!\!\!Test 2&\emph{A campus with river or forest.} & \emph{forest and trees} & \emph{river} & \bf{0.27} & -3.81 & 0.25  & -3.72 & \bf{0.28} & -3.72 & 0.26 & \bf{-3.68} \\
    \hline
    \!\!\!Test 3&\emph{Red, Blue and Yellow Squares} & \emph{Mondrian} & \emph{Vincent van Gogh} & \bf{0.31} & -3.80 & 0.26  & -3.71 & \bf{0.30} & -3.71 & 0.26 & \bf{-3.66} \\
    \hline
    \!\!\!Test 4&\emph{Home-cooked meal in Russia.} & \emph{sausage, meat} & \emph{tomato, onion} & \bf{0.29} & -3.80 & 0.27  & -3.64 & \bf{0.29} & -3.65 & 0.27 & \bf{-3.61}  \\
    \hline
    \end{tabular}}
    \vspace{-10pt}
    \caption{The diversity score (Div) and the $\texttt{AugCLIP}$ score (Sc) our different methods on text-to-image generation. 
    %and compare the linear combination with $\Fsum$ and $\Fmax$.
    }
    % \vspace{-5pt}
    \label{tab:image2text_score}
\end{table*}

\begin{figure*}[h]
\centering
\scalebox{.88}{
\begin{tabular}{cccccc}
\raisebox{1.1em}{\rotatebox{90}{$s_{\texttt{AugCLIP}}$}}
\includegraphics[width=.18\textwidth]{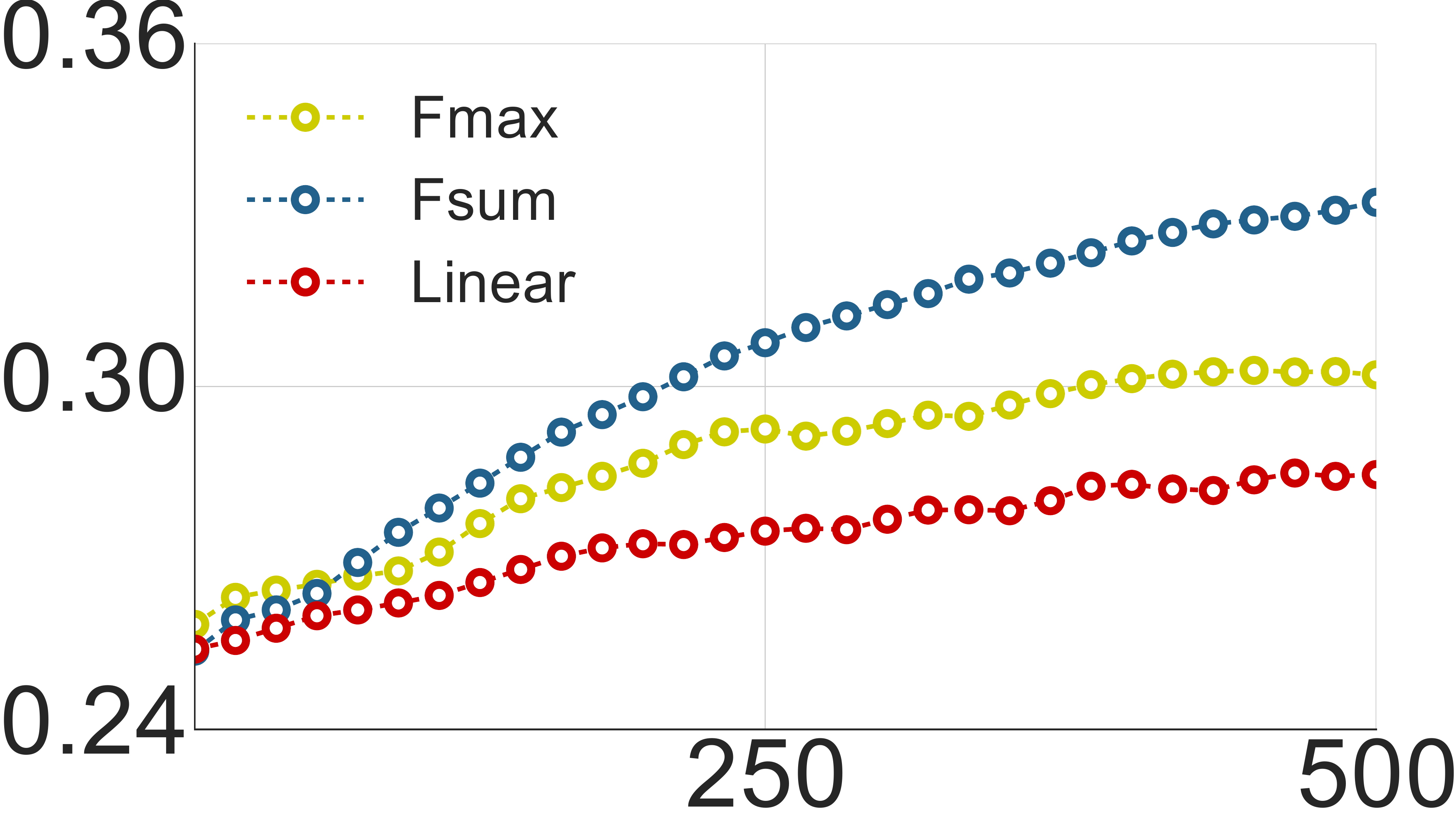} & 
\raisebox{1.1em}{\rotatebox{90}{Diversity }}
\includegraphics[width=.18\textwidth]{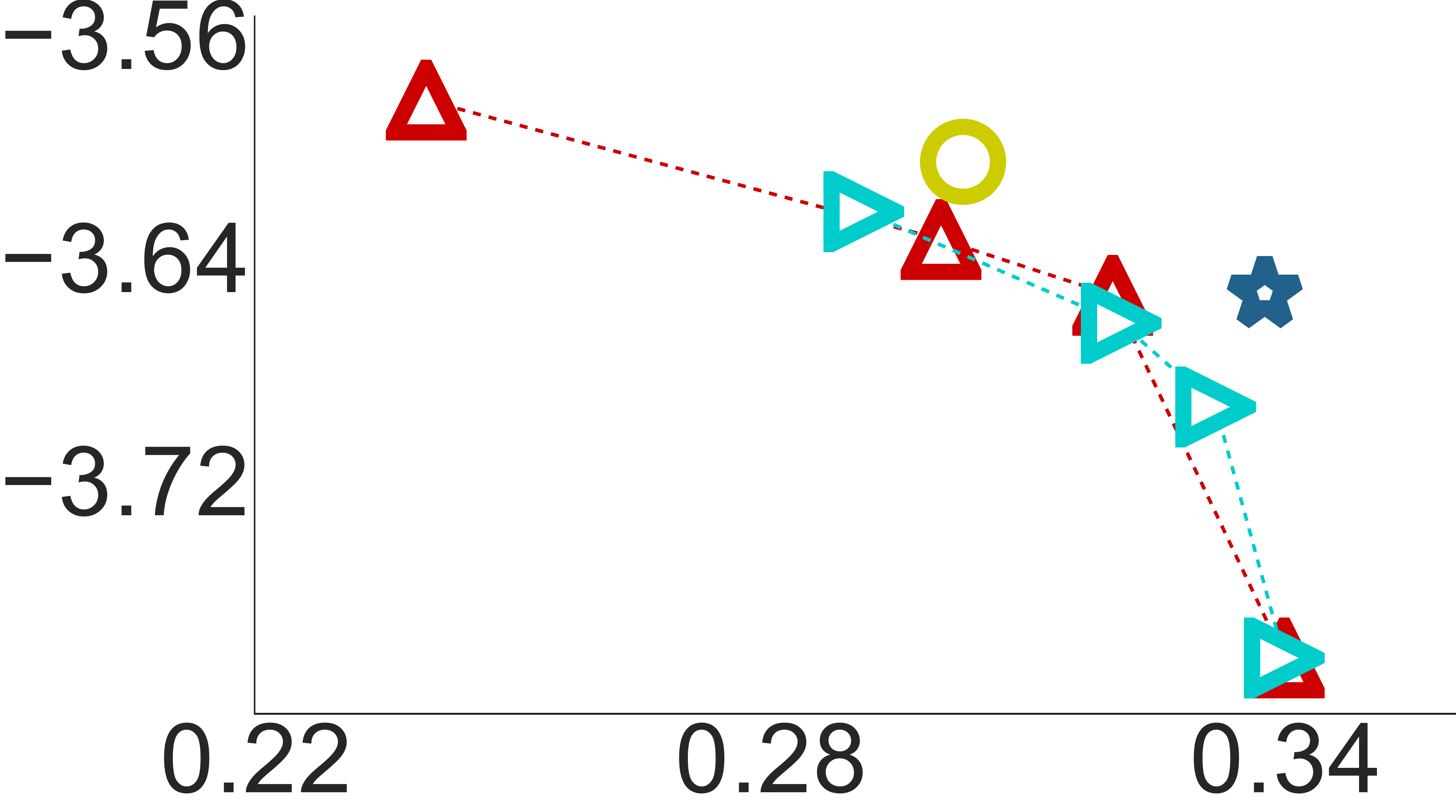}
& 
\includegraphics[width=.18\textwidth]{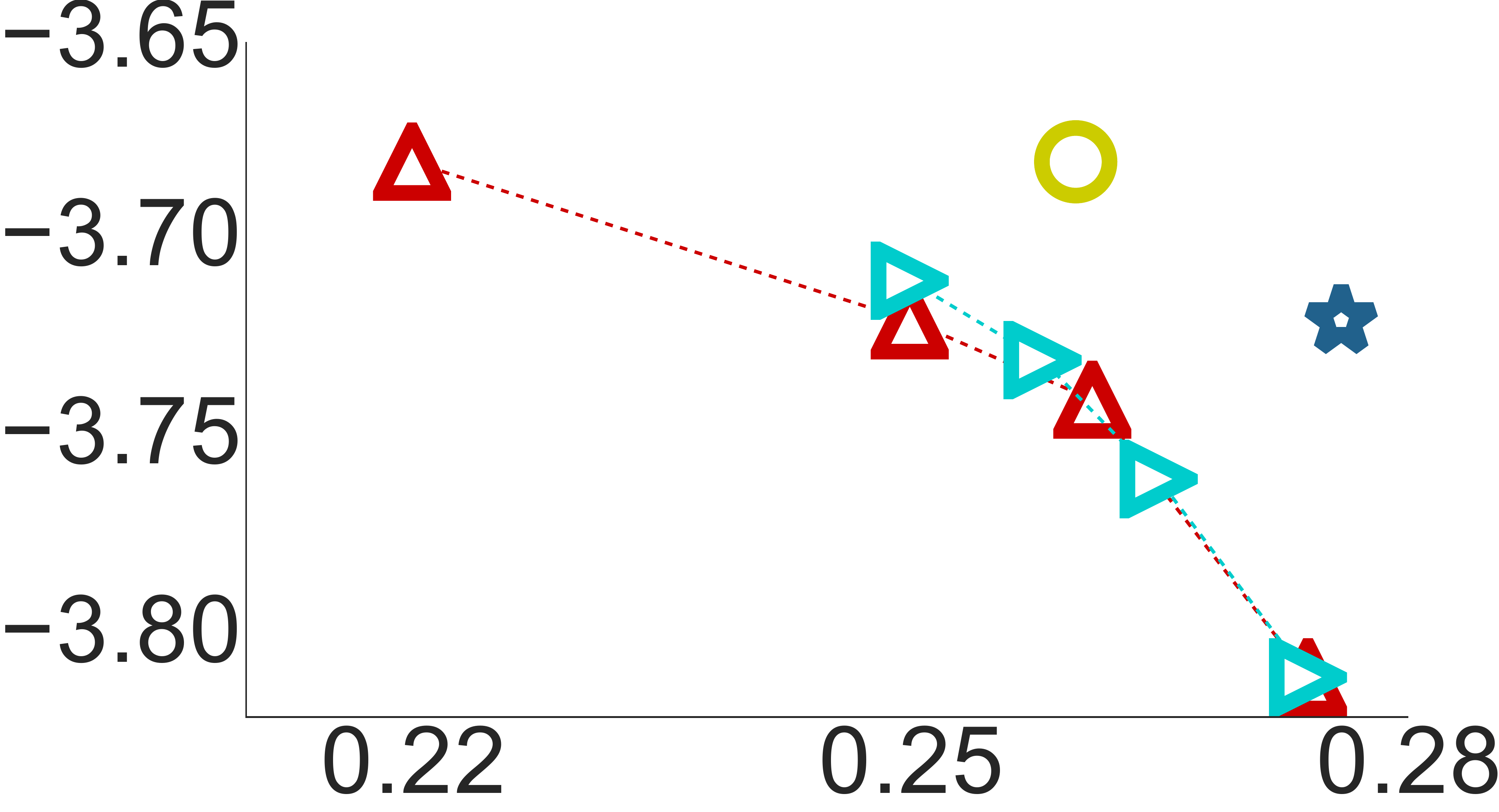} 
& 
\includegraphics[width=.18\textwidth]{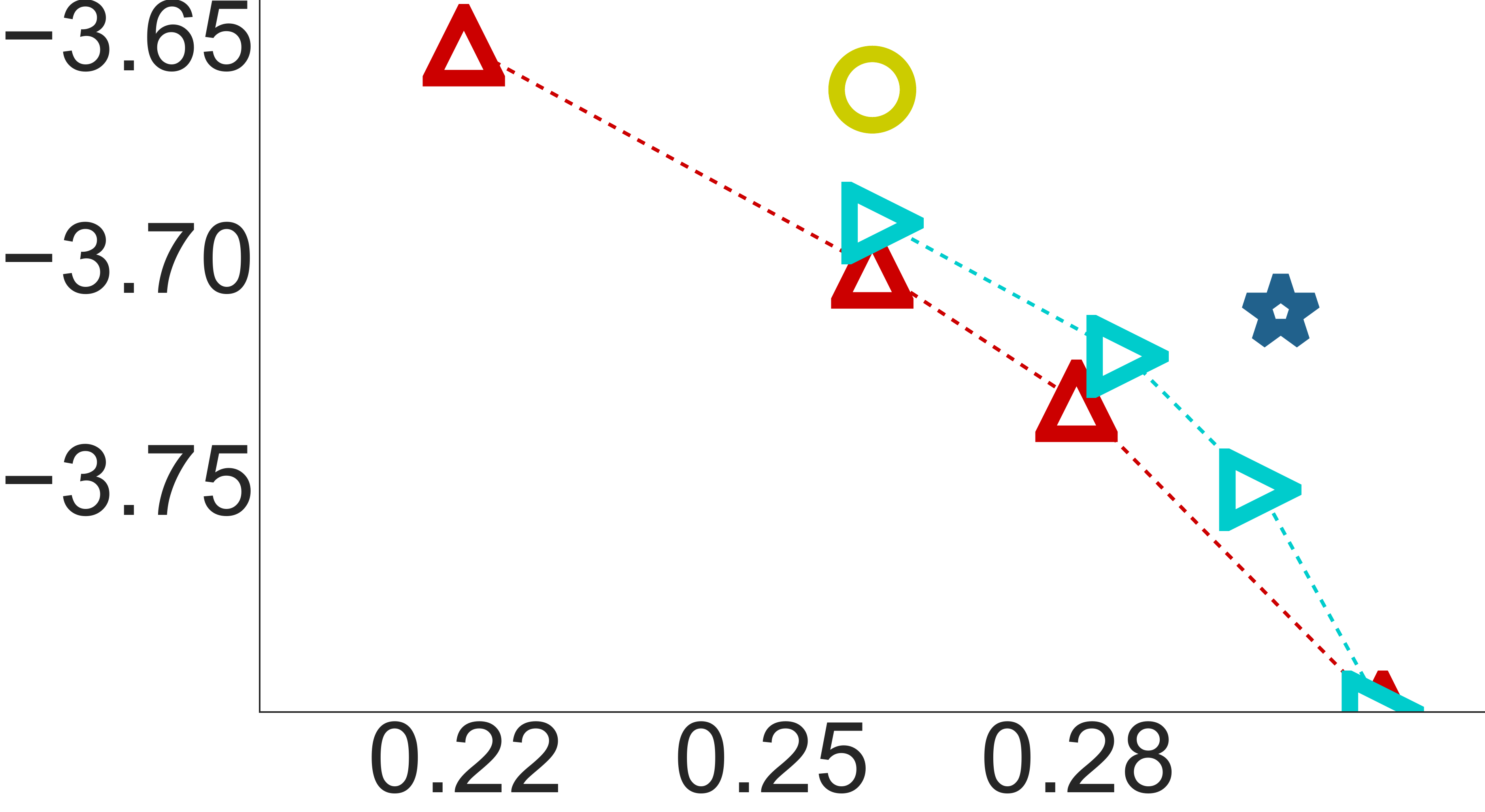} 
&
\includegraphics[width=.18\textwidth]{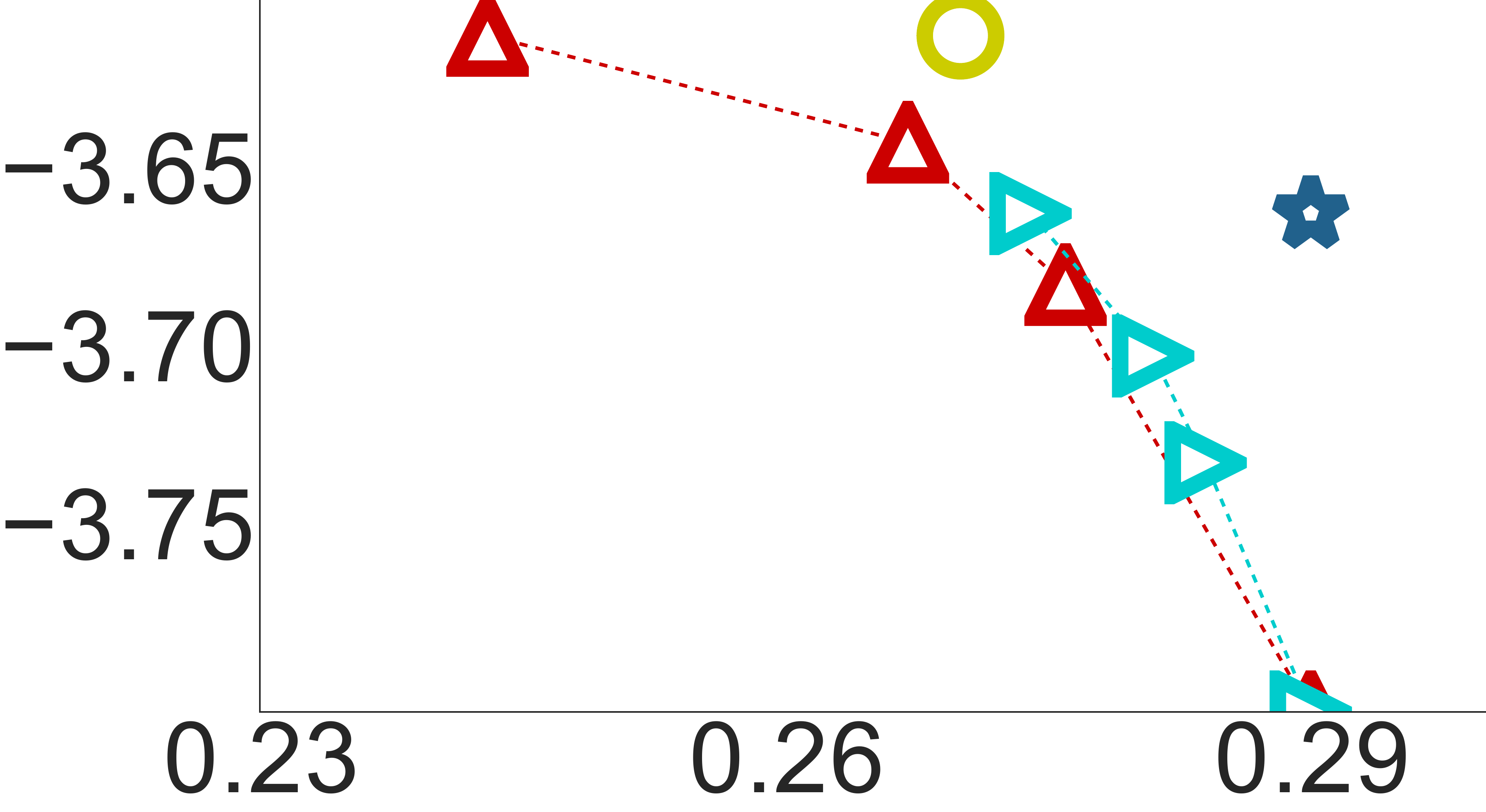} 
& 
\hspace{-0.4em}
\raisebox{1.5em}{\includegraphics[height=0.07 \textwidth]{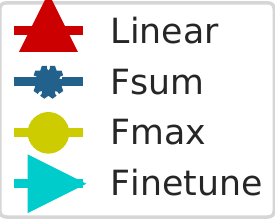}} 
\vspace{-5pt}\\
 Iterations & ~~$s_{\texttt{AugCLIP}}$ & $s_{\texttt{AugCLIP}}$ & $s_{\texttt{AugCLIP}}$ & $s_{\texttt{AugCLIP}}$ & \\ 
 (a) Test 1 & ~~(b) Test 1 & (c) Test 2 & (d) Test 3 & (e) Test 4
 \vspace{-10pt}
\end{tabular}}
% \vspace{-5pt}
\caption{
(a) The $\texttt{AugCLIP}$ score vs. iteration in Test 1 in Table~\ref{tab:image2text_score}. 
(b)-(d) The (quality, diversity) front of $\Fsum$ (blue), $\Fmax$ (yellow), linear combination with $\alpha\in\{ 0, 0.25, 0.5, 0.75\}$ (red), 
and `finetuned' linear combination with the same set of $\alpha$ (Cyan). 
Here `finetune' refers to optimizing the linearly combined loss in \eqref{equ:linear} for 250 iterations and then optimize  $s_\texttt{AugCLIP}(\cdot)$ for another 250 iterations with the diversity term turned off. 
$\Fsum$ and $\Fmax$ outperform both variants of linear combination.
}
\label{fig:text2image_score}
\vspace{-5pt}
\end{figure*}

\begin{figure}[h]
\centering
\scalebox{0.9}{
\begin{tabular}{c}
% \blue{\scriptsize Test 1: \it$\mathcal{T}=$`A painting of an either blue or red dog.' ~~~~~~$\mathcal{T}_1, \mathcal{T}_2=$(`blue', `red')} \\
\includegraphics[width=.5\textwidth]{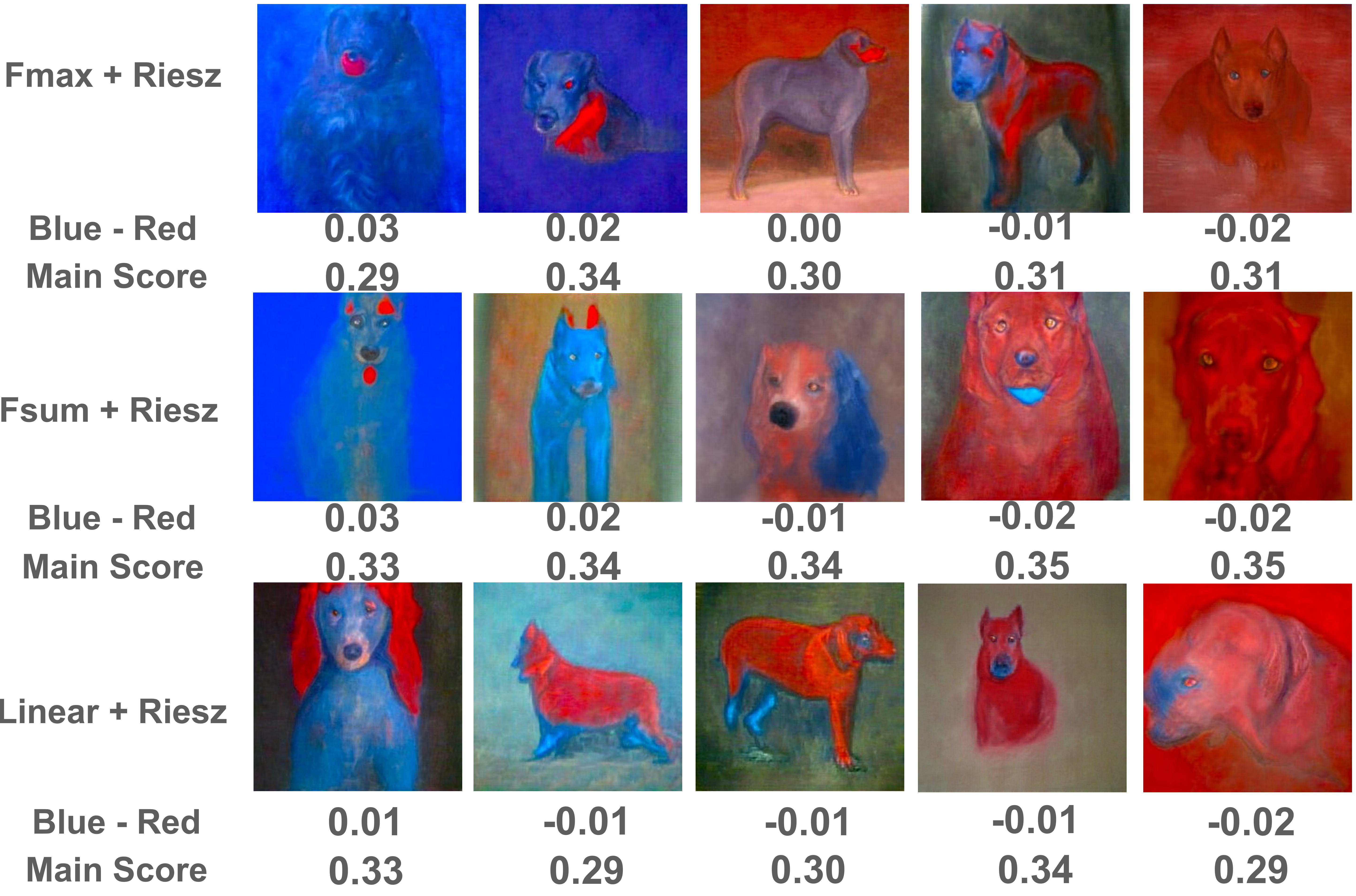}
\end{tabular}}
\vspace{-12pt}
\caption{ 
Images for Test 1 in Table~\ref{tab:image2text_score}. `Linear' refers to the linear combination ($\alpha=0.5$).
There are two numbers under each image:
the numbers on the top row are 
% the difference in clip score, 
$s_{\texttt{AugCLIP}} \left (\mathcal{T}_1, g(x_i) \right ) -  s_{\texttt{AugCLIP}}\left (\mathcal{T}_2, g(x_i) \right )$, which reflects the percentage of blue vs. red colors; the numbers on the second row are $s_{\texttt{AugCLIP}}$.
See Appendix for more examples.
}
\label{fig:test1_examples}
\vspace{-12pt}
\end{figure}

\textbf{Quantitative Analysis } We present the value of quality score 
$s_\texttt{AugCLIP}$ and diversity score $\Phi$ 
given by our methods ($\Fmax$ and $\Fsum$) and the linear combination method \eqref{equ:linear}  
on an additional set of examples. 
In Table~\ref{tab:image2text_score} and Figure \ref{fig:text2image_score}, 
we can see that %We notice
that $\Fmax$ and $\Fsum$ always achieve better results than tuning the coefficient value for the linear combination. $\Fmax$ generates more diverse results while $\Fsum$ optimizes the main loss better.
We also find that if we finetune the linear combination result by turning off the diversity promoting loss at the end of the optimization, it 
% achieves a lightly better $\texttt{AugCLIP}$ score but worse diversity score, and
does not improve the (quality, diversity) Pareto front (see Figure \ref{fig:text2image_score}).
Figure \ref{fig:test1_examples} shows more detailed analysis on the Test 1.  
% We visualize the optimization results for different variants. 

\begin{table}[h]
    \centering
    \scalebox{0.82}{
    \begin{tabular}{c|ccc|ccc}
    \hline
    \multirow{2}{*}{Test} & \multicolumn{3}{c|}{MEGA}  & \multicolumn{3}{c}{$\Fsum$}  \\
    \cline{2-7}
    & Sc $\uparrow$ & Div $\uparrow$ & Hours $\downarrow$ & Sc $\uparrow$ & Div $\uparrow$ & Hours $\downarrow$\\ 
    \hline
    1 & \bf{0.35} & -3.62 & \multirow{4}{*}{0.97} & 0.34 & \bf{-3.64} & \multirow{4}{*}{ \bf{0.16} } \\
    \cline{1-3}\cline{5-6}
    2 & \bf{0.28}  & -3.71 & & \bf{0.28} & -3.72 &\\
    \cline{1-3}\cline{5-6}
    3 & \bf{0.32}  & -3.69 & & 0.30 & \bf{-3.71} & \\
    \cline{1-3}\cline{5-6}
    4 & \bf{0.29} & \bf{-3.66} & & \bf{0.29} & -3.65 & \\
    \hline
    \end{tabular}}
    % \vspace{-5pt}
    \caption{Comparison with MEGA. 
    % MEGA and $\Fsum$ can achieve comparable score-diversity trade-off.
    Hours is measured on a \emph{NVIDIA GeForce RTX3090} GPU. 
    }
    \label{tab:image2text_compare_es}
    \vspace{-15pt}
\end{table}

\begin{table}[h]
    \centering
    \scalebox{0.72}{
    \begin{tabular}{c|cc|cc|cc|cc}
    \hline
    \multirow{3}{*}{Test} & \multicolumn{4}{c|}{Iteration: 250} & \multicolumn{4}{c}{Iteration: 500}\\
    \cline{2-9}
     & \multicolumn{2}{c|}{\small \citet{gong2021automatic}}  & \multicolumn{2}{c|}{$\Fsum$}  & \multicolumn{2}{c|}{\small \citet{gong2021automatic}} & \multicolumn{2}{c}{$\Fsum$}  \\
    \cline{2-5}\cline{6-9}
    & Sc $\uparrow$ & Div $\uparrow$ & Sc $\uparrow$ & Div $\uparrow$ & Sc $\uparrow$ & Div $\uparrow$ & Sc $\uparrow$ & Div $\uparrow$ \\ 
    \hline
     1& 0.26 & -3.62 & \bf{0.31} & -3.66 & 0.32 & -3.65 & \bf{0.34} & -3.64 \\
     2&  0.23 & -3.68 & \bf{0.26} & -3.73  & 0.26 & -3.71 & \bf{0.28} & -3.72  \\
     3& 0.24 & -3.67 & \bf{0.28} & -3.71 &  0.27 & -3.71 & \bf{0.30} & -3.71  \\
     4&  0.23 & -3.62 & \bf{0.26} & -3.64 & 0.27 & -3.63 & \bf{0.29} & -3.65 \\
    \hline
    \end{tabular}}
    % \vspace{-5pt}
    \caption{Comparison with \citet{gong2021automatic}.  
    % `S' and `Div' stand for the main score $s_{\texttt{AugCLIP}}$ and the diversity value, respectively. 
    }
    \label{tab:image2text_compare_lexico}
    \vspace{-10pt}
\end{table}

\textbf{Compare with MEGA} We compare with 
MAP-Elites via a Gradient Arborescence (MEGA) \citep{fontaine2021differentiable}, a recent improvement of MAP-Elites \citep{mouret2015illuminating} in Table~\ref{tab:image2text_compare_es}.  
%Compared to MEGA
%our method can achieve similar performance but with less time cost. 
% Instead of 10,000-iteration evolutionary updates, we can achieve similar performance with much fewer iterations. As displayed inTable\ref{tab:image2text_compare_es}, 
Compared with MEGA, we find that our method requires far less optimization time (1 hours \emph{v.s.} 10 minutes) and achieves comparable results.
See Appendix for more detailed comparisons. % on generated examples.

\textbf{Compare with \citet{gong2021automatic} }
%We compare with 
\citet{gong2021automatic} provides 
an off-the-shelf method for solving general lexicographic optimization of form \eqref{equ:phiF}. We apply it
to the case of $F = \Fsum$ and compare it with our $\Fsum$ method in Table \ref{tab:image2text_compare_lexico}. Our $\Fsum$ method yields faster convergence on the main loss, 
and \citet{gong2021automatic} also %compared to Lexico. 
 outperforms  the  linear combination baseline with $\alpha=0.5$ shown in Table \ref{tab:image2text_score}. On the other hand, we find that \citet{gong2021automatic} fails to work when $F=\Fmax$ due to the non-smoothness of the max function.

\begin{figure}[h]
    \centering
    \includegraphics[width=0.48\textwidth]{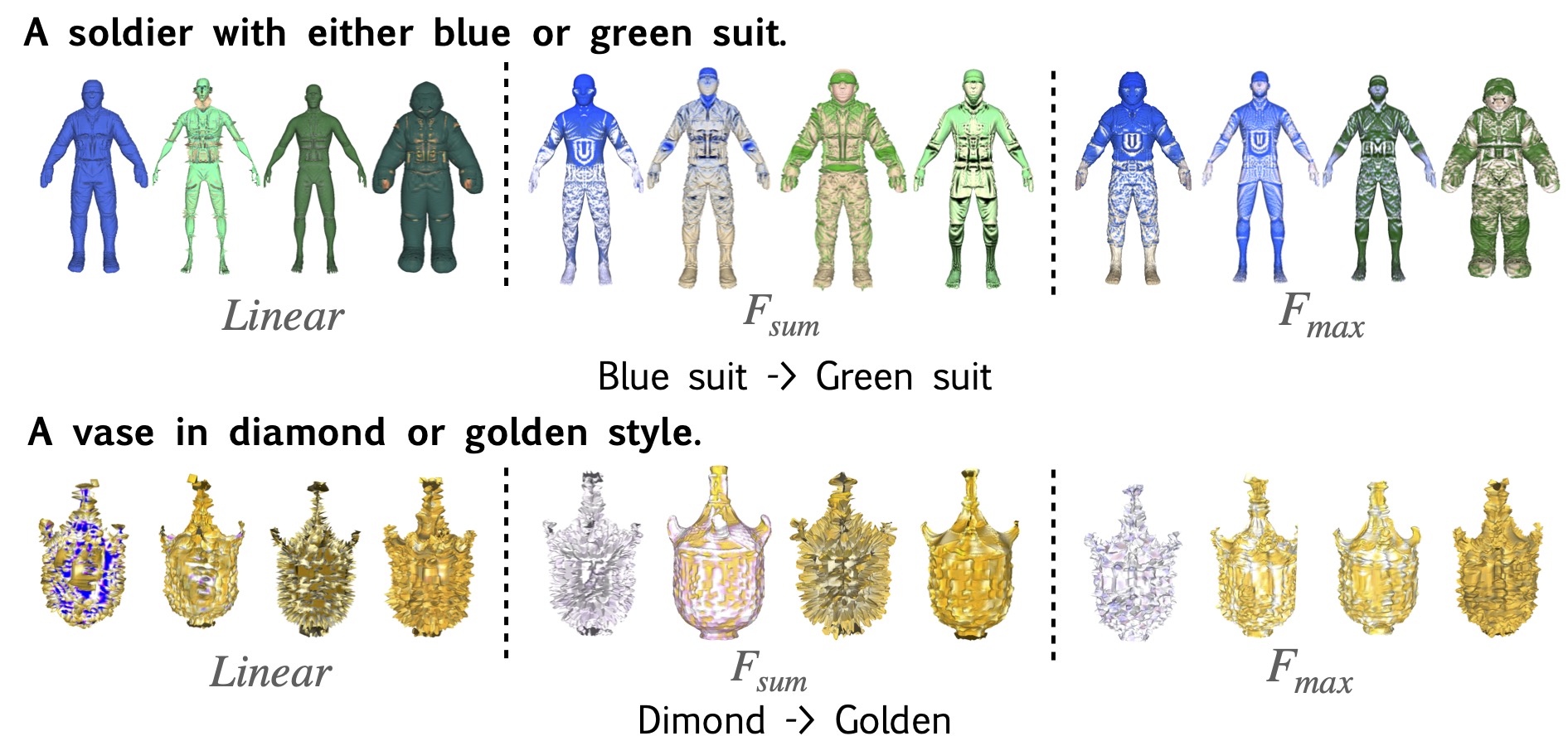}
    % \vspace{-20pt}
    \caption{
    Results on Text2Mesh generation form $\Fsum$, $\Fmax$ and linear combination ($\alpha=0.5$).
    See supplementary materials for a detailed description video.}
    \label{fig:mesh}
    \vspace{-15pt}
\end{figure}

\subsection{Controllable Diverse Generation on Meshes} 
We apply our method to generate diversified meshes of 3D objectives from text.
We base our method on 
Text2Mesh \citep{michel2021text2mesh}, 
and promote the diversity with a CLIP-based semantic diversity score similar to  \eqref{eq:psit1} 
that is based on a pair of test $\mathcal T_1, \mathcal T_2$.  
See Appendix for more detailed setup.

Figure~\ref{fig:mesh} shows the result of the $\Fsum$, $\Fmax$ and the linear combination method $\alpha = 0.5$. 
We set $\alpha = 0.5$ to have a similar diversity score with the $\Fsum$, $\Fmax$ methods.
We run 1,500 iterations for $\Fsum$ and $\Fmax$. 
For the linear combination method, we apply the diversity term for 750 steps and finetune for another 750 steps with the diversity promotion turned off.
We can see that 
$\Fsum$ and $\Fmax$ generate the 3D models with different cloth styles in the green and blue colors which satisfy the text prompt. 
On contrast, the linear combination baseline fails to keep the reasonable geometry and provide a reasonable color; see e.g., the overly-thin and overly-thick mesh displacement on soldier case, and the non-smoothing meshes and the purple color in vase case.
% See supplementary material for the video showing more view angles of the 3D objects.  

\subsection{Diversified Molecular Conformation Generation}
%We study 
A fundamental problem in computational  chemistry is molecular conformation generation, predicting stable 3D structures from 2D molecular graphs. 
The goal is to take a 2D molecular graph representation $G$ of a molecule
and predict its 3D conformation (i.e., the 3D coordinates of the atoms in the molecule). Diversity is essential in this problem since there are multiple possible conformations of a single molecule and we hope to predict all of them. 

Specifically, we are interested in generating a set of possible conformations $\vv x= \{x_1, \ldots x_m\}$ of a given molecule, where $x_i \in \RR^{3\times d}$ is the 3D coordinates of $d$ atoms in the molecule.
Let $E(x)$ be an energy function of which the true configurations are local minima, we generate $\vv x$ by solving 
\begin{align}
\label{eq:confg}
 \min_{\vv{x}} \Phi_s(\vv{x}),  ~~s.t.~~ \vv{x} \subseteq \arg\min   %s_{\theta}
 E.    
\end{align}
For our experiments, we adopt the energy function from ConfGF \citep{shi2021learning}, which is implicitly defined with a learnable gradient field trained on the GEOM-QM9 dataset~\citep{axelrod2020geom}. 

We evaluate the method by comparing the conformations predicted from  \eqref{eq:confg} with the set of ground truth conformations (denoted by $\vv x^*$) of the molecule of interest from GEOM-QM9. Because the 3D coordinates are unique upto rotation and translation. We measure the difference between two conformations $x, x'$ using the root mean square deviation: 
$\text{RMSD}(x,x') = \min_{T} \norm{T(x) - x'}_2$, where $T$ is minimized on the set of all possible rotations and translations. 
For the set of predicted   $\vv x $ and ground truth $\vv x^*$ conformations,  %Following the setting in \citep{shi2021learning}, 
%we measure the quality and diversity with the
we calculate the 
matching score (MAT) for evaluating quality and the coverage score (COV) to measure diversity following~\citet{shi2021learning},  
% this two scores are defined as 
\begingroup\makeatletter\def\f@size{9}\check@mathfonts
\def\maketag@@@#1{\hbox{\m@th\large\normalfont#1}}
\begin{align*}
    & \text{COV}(\vx, \vx^*)= {\sharp \{ x^* \in \vx^* | \text{RMSD}(x^*, x) < \delta, x \in \vx \}  } /\sharp \vx^*,  \\
    & \text{MAT}(\vx, \vx^*) = \sum_{x\in \vx^*} \min_{x \in \vx}      \text{RMSD}(x^*, x)/\sharp \vx^*, 
\end{align*}
\endgroup
where $\sharp$ denotes the number of elements of a set. 
Both COV and MAT measure the precision of the prediction (how may predictions are found in ground truth list); 
to measure recall (how every ground truth conformation is found by at least a prediction), we also calculate 
$\text{RMAT}(\vx, \vx^*):=\text{MAT}(\vx^*, \vx)$, the recall matching score (RMAT).

\textbf{Baselines } We use the same model trained from ConfGF and use our $\Fsum$ and $\Fmax$ method as a way to enhance diversification during inference.  % which has two variants:
%We adopt two strategies during the inference stage.
We test two inference strategies for both  our method and the baselines:
One is the original ConfGF strategy, which randomly initializes $2\times \sharp \vx  ^*$ number of conformations 
% as the reference set to get a fully coverage 
and filters half of them to get $1\times  \sharp \vx ^*$  number of predicted conformations. 
The original inference strategy of ConfGF can be viewed as a naive multi-initialization strategy.
%The other is only
We also test another strategy that directly predicts $\sharp \vx  ^*$ confirmations as $\vv x^*$. 
%initializing the same number of conformations candidates at the beginning.
We denote these two baselines as $2\times$Ref and $1\times$Ref, respectively. 

\begin{table}[h]
    \centering
    \scalebox{0.82}{
    \begin{tabular}{l|l|cc|cc|cc}
    \hline
    \hline
    \# Init & Method  & \multicolumn{2}{c|}{COV (\%) $\uparrow$}  & \multicolumn{2}{c|}{MAT ($\mathring{\text{A}}$) $\downarrow$}  & \multicolumn{2}{c}{RMAT ($\mathring{\text{A}}$) $\downarrow$}  \\
    \hline
     & ConfGF & 77.7 & 78.0 & 0.338 & 0.346 & 0.530 & 0.514  \\
    $1\times$Ref & $\Fsum$ & \textbf{79.2} & \textbf{80.9} & \textbf{0.332} & \textbf{0.339} & \textbf{0.504} & \textbf{0.490}  \\
    & $\Fmax$ & 79.4 & 80.5 & 0.336 & 0.340 & 0.512 & 0.501  \\
    \hline
     & ConfGF & 90.0 & 94.6 & \textbf{0.267} & 0.269 & 0.502 & 0.499  \\
    $2\times$Ref & $\Fsum$ & \textbf{90.3} & \textbf{94.9} & 0.270 & \textbf{0.268} & \textbf{0.483} & \textbf{0.475}  \\
    & $\Fmax$ & 89.8 & 94.3 & 0.273 & 0.271 & 0.495 & 0.497  \\

    \hline
    \hline
    \end{tabular}}
    % \vspace{-5pt}
    \caption{Results on diversified molecule conformation generation using $\Fsum$, $\Fmax$ and the linear combination method. 
    }
    \label{tab:confgf}
    \vspace{-10pt}
\end{table}

\textbf{Results } See the results in Table \ref{tab:confgf}. 
%we notice that
We find that 
both $\Fsum$ and $\Fmax$ yield better results than the baseline in all the metrics (e.g. COV, MAT, and RMAT), 
and $\Fsum$ yields the best COV diversity score and Pareto front (precision, recall) among all methods.

\subsection{Training Ensemble Neural Networks}
Another natural application of our method is learning diversified neural network ensembles. 
%Let $f_{\theta}$ be a neural network whose parameter is
Let $\theta$ be the parameter of a neural network model. We are interested in learning a set of neural networks $\vv \theta = (\theta_1, \ldots, \theta_m)$ by solving 
\begin{align}
\label{eq:nn_loss}
\vspace{-5pt}
     \min_{\vv \theta} \Phi(\vv \theta),  ~~s.t.~~ \vv \theta \subseteq \arg\min  \ell_{\text{train}},
\vspace{-5pt}
\end{align}
where $\ell_{\text{train}}(\theta)$ is a standard training loss of the neural network, and $\Phi(\vv \theta) = \E_{x\sim \mathcal D}[\Phi_s(f_{\theta_1}(x)\ldots f_{\theta_m}(x))]$ is the diversity defined w.r.t. the hidden nodes of the last layer $f_\theta$ of the neural networks on training data $\mathcal D$. % \qq{how we do define $\mathcal D$? is it the same as the training data?}

\begin{table}[h]
    \centering
    \scalebox{0.9}{
    \begin{tabular}{l|ccc|cc}
    \hline
    \hline
    & \multicolumn{3}{c|}{Linear Combination} & \multirow{2}{*}{$\Fsum$} & \multirow{2}{*}{$\Fmax$} \\
    \cline{2-4}
    & $0.0$ & $0.1$ & $0.9$ & & \\
    \hline
    Single Acc $\uparrow$ & 91.4 & 90.9 & 89.8 & 91.2 & 90.7 \\
    Ensemble Acc $\uparrow$ & \bf{92.0} & 91.4 & 90.5 & \bf{92.0} & 91.3 \\
    \hline
    Diversity $\uparrow$ & -4.11 & -4.04 & \bf{-4.01} & -4.07 & -4.03 \\
    \hline
    ECE $\downarrow$ & 4.03 & \bf{3.39} & 3.53 & \bf{3.38} & 3.51 \\
    \hline
    \hline
    \end{tabular}}
    % \vspace{-5pt}
    \caption{
    Results on learning diversified ensemble neural networks with $\Fsum$, $\Fmax$ and the linear combination with $\alpha\in\{0,0.1, 0.9\}.$ 
    }
    \vspace{-5pt}
    \label{tab:ensemble}
\end{table}

\textbf{Results } Table \ref{tab:ensemble} shows the results when we train three $(m=3)$ ResNet-56 models on CIFAR-10 dataset.  
We observe that the linear combination often hurts the single network accuracy and hence yields poorer accuracy compared to the case without diversity regularization ($\alpha=0$). 
In comparison, $\Fsum$ improves both the diversity and ECE score without hurting the model accuracy and achieves the best accuracy-diversity trade-off.
% \qq{check/modify; not sure}

%% file: tex/conclusion.tex
\section{Conclusion}
In this work, we propose a framework of gradient-based optimization methods to find diverse points in the optimum set of a loss function with a harmless diversity promotion mechanism. 
We find that our methods yield 
 both diverse and high-quality solutions on a broad spectrum  of applications. % tasks,
%our methods efficiently achieve diverse but accurate solutions.
Another important application that we have not explored is robotics, 
finding diverse policies of critically important for planning and reinforcement learning. We will explore it in future works. 
%In the future, we plan to apply our tools to robotics and other related fields.
% \paragraph{Social Impacts and Limitations}

%% file: tex/appendix.tex
\section{Details about Generation on Meshes}
Let $\mathcal M = g(\vx)$ be a mesh generator, and $I = P(\mathcal M)$ be a differentiable render that generates (a set of) images from the three objective specified by $\mathcal M$. 
Given a text prompt $\mathcal T$, and a diversity score $\Phi$, we want to find a diverse set of meshes $\mathcal M_i = g(x_i)$ by solving 
%our method solves
$$
\min_{\vx}\Phi(\vx) ~~~ s.t.~~~ 
\vx \in \arg\min_{\vv x'} s_{\texttt{CLIP}}(\mathcal T, P(g(\vv x'))),
$$
where $\Phi(\vx) = \Psi_s(\psi(\vx))$ with $\psi$ defined similar to \eqref{eq:psit1} based on a pair of text $\mathcal T_1$, $\mathcal T_2$: 
$$
\psi(x) =  
\bigg [
s_{\texttt{CLIP}}(\mathcal T_1, P(g(x))), s_{\texttt{CLIP}}(\mathcal T_2, P(g(x))) 
\bigg ].
$$
%on top of it. 
We follow the model architecture and generation pipeline in Text2Mesh \citep{michel2021text2mesh}. Text2Mesh proposes a neural style field network, which directly outputs the value displacement on the mesh normal and the color on each vertex. A differentiable renderer is then rendering multiple 2D views for the styled mesh as the image set. 
 The pipeline wants to get a higher CLIP-based similarity score between the rendered image set and query text. 
Similar to image generation, we create the diversity measure with additional text prompts and hope to make the Text2mesh model generate a variety of different mesh styles with RGB colors and textures.
%
%
%\paragraph{Generated Mesh}
%Following \citep{michel2021text2mesh}, 
We apply 
four same meshes with the zero RGB color and geometry displacement and then optimize the color and displacement variables with 1,500 steps. We attach videos for the results in Figure~\ref{fig:mesh} in the supplementary material for a better visualization with more angles.

\section{More Examples}

\begin{figure}[h]
\centering
\scalebox{0.92}{
\begin{tabular}{ccc}
% \hspace{-25pt}
\includegraphics[width=.3\textwidth]{figs/fmax_log_multi2.pdf} & 
% \hspace{-10pt}
\includegraphics[width=.3\textwidth]{figs/fsum_log_multi2.pdf}
& 
% \hspace{-10pt}
\includegraphics[width=.3\textwidth]{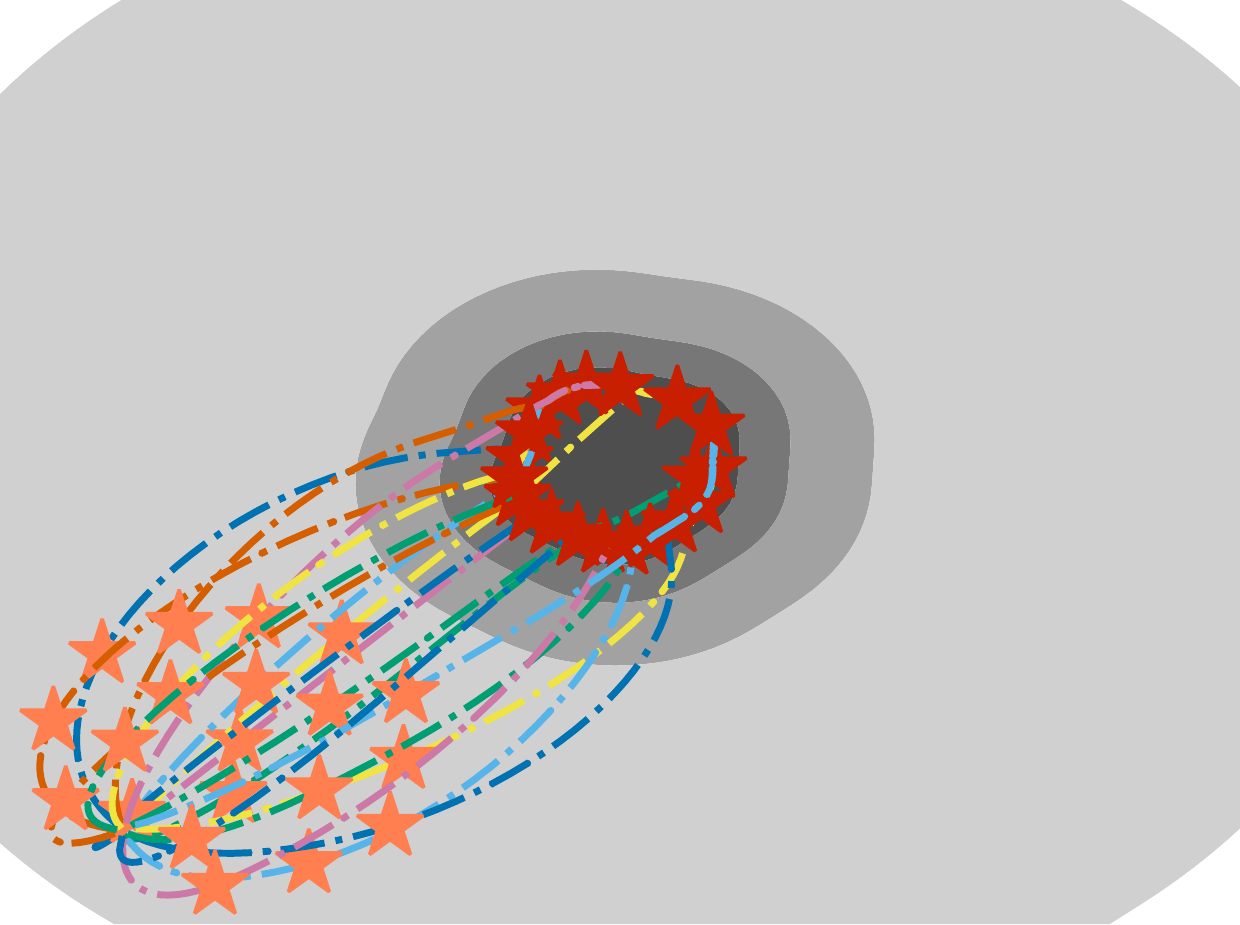}
% \vspace{-20pt}
\\
% \hspace{-25pt}
\includegraphics[width=.3\textwidth]{figs/fmax_log_curve.pdf} & 
% \hspace{-10pt}
\includegraphics[width=.3\textwidth]{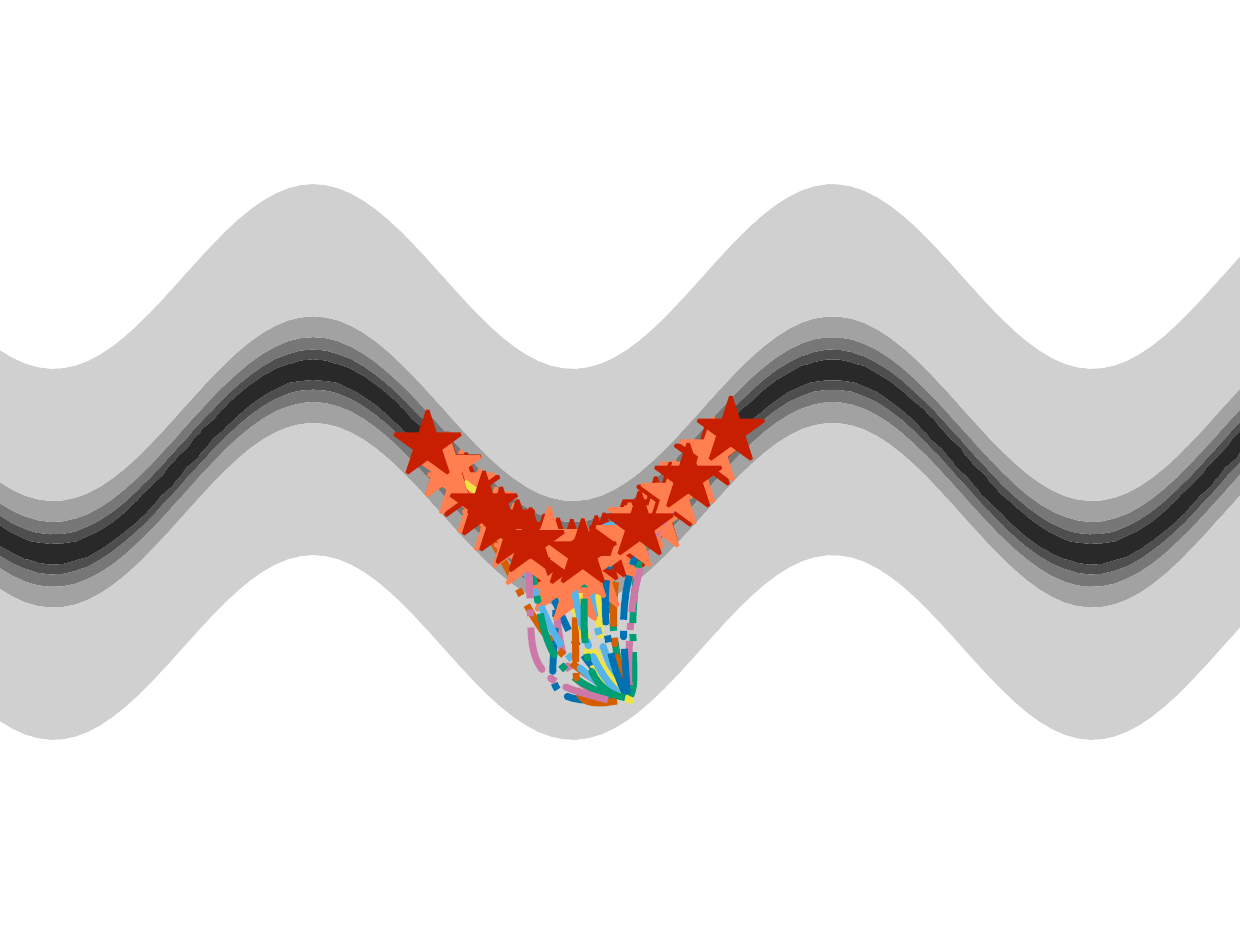}
& 
% \hspace{-10pt}
\includegraphics[width=.3\textwidth]{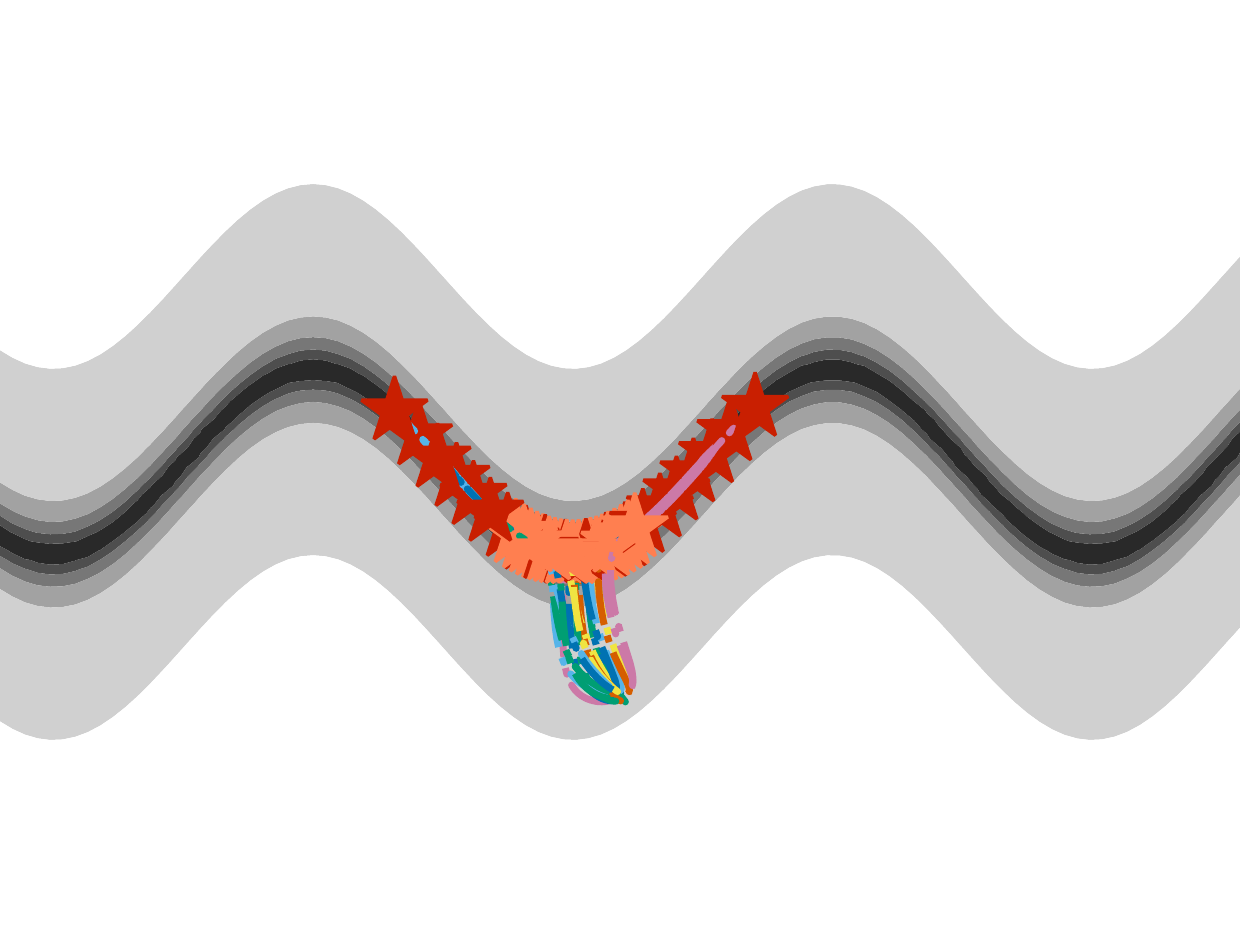}
\vspace{-20pt}\\
\includegraphics[width=.3\textwidth]{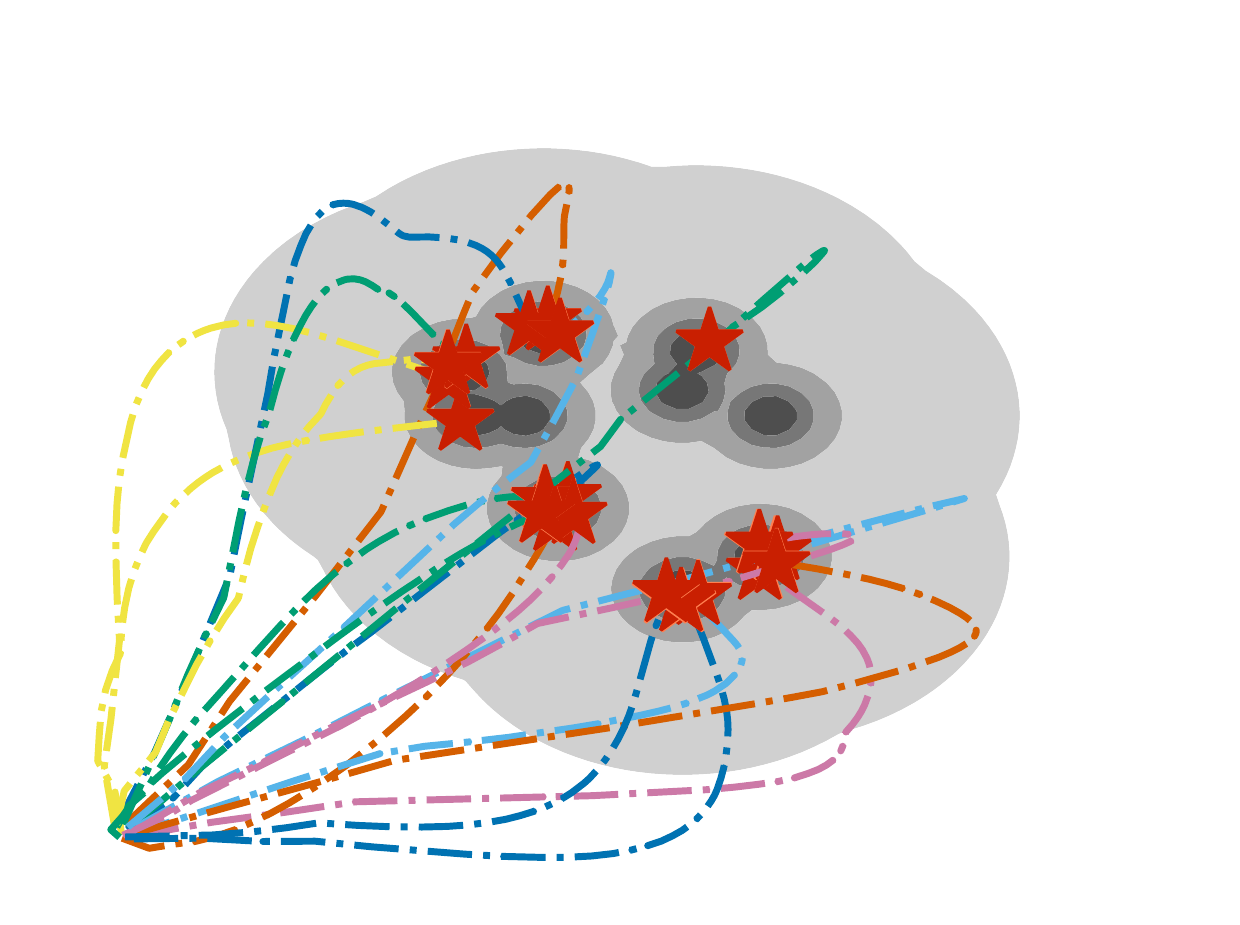} & 
% \hspace{-10pt}
\includegraphics[width=.3\textwidth]{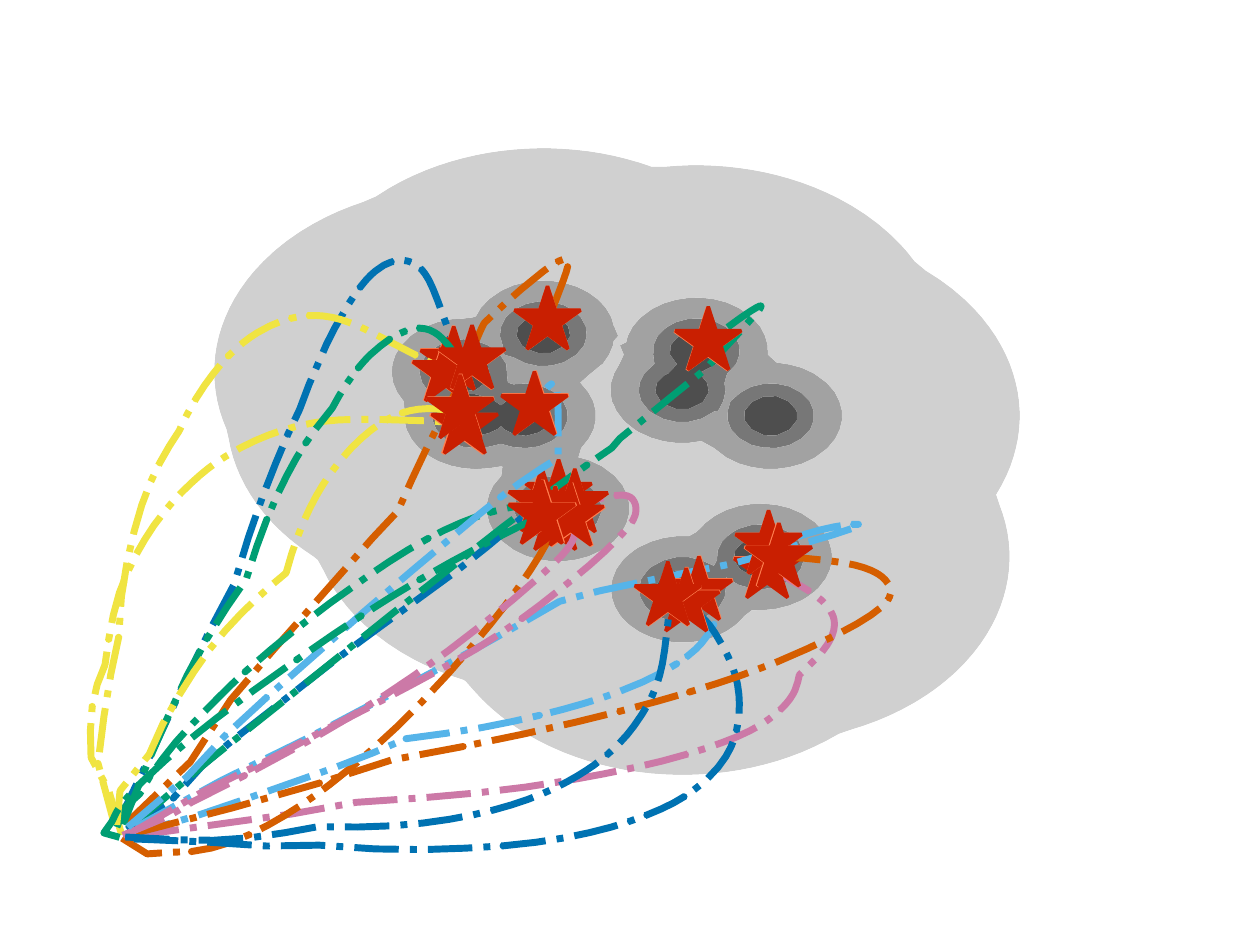}
& 
% \hspace{-10pt}
\includegraphics[width=.3\textwidth]{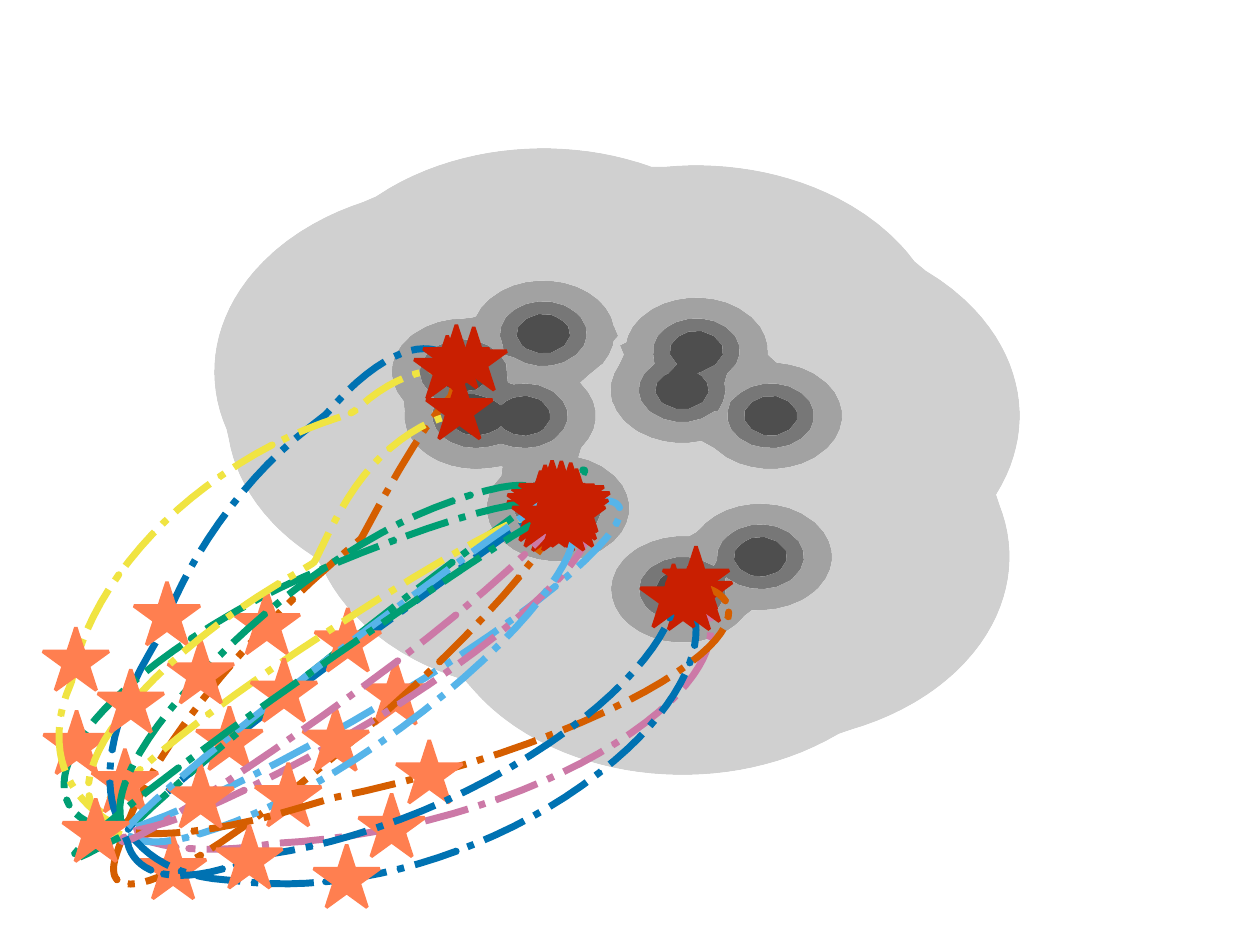} \\
$\Fmax$ & $\Fsum$ & \citet{gong2021automatic}  \\
% & $\Fmax$~~~~~~~~~~ & $\Fsum$~~~~~~~~~~& \hspace{-25pt} {\small $\Fmax$: $F$ \emph{v.s.} Diversity}\\ 
\end{tabular}}
\vspace{-10pt}
\caption{
Results on toy examples with $\Fmax$, $\Fsum$ and \citet{gong2021automatic} . We notice that $\Fmax$ and $\Fsum$ empirically converge faster. For multi-modal examples, our version captures more modals than \citet{gong2021automatic}.
}
\label{fig:compare_with_lexico}
\vspace{-5pt}
\end{figure}

\begin{figure}[h]
\centering
\scalebox{.8}{
\begin{tabular}{c}
\includegraphics[width=1.\textwidth]{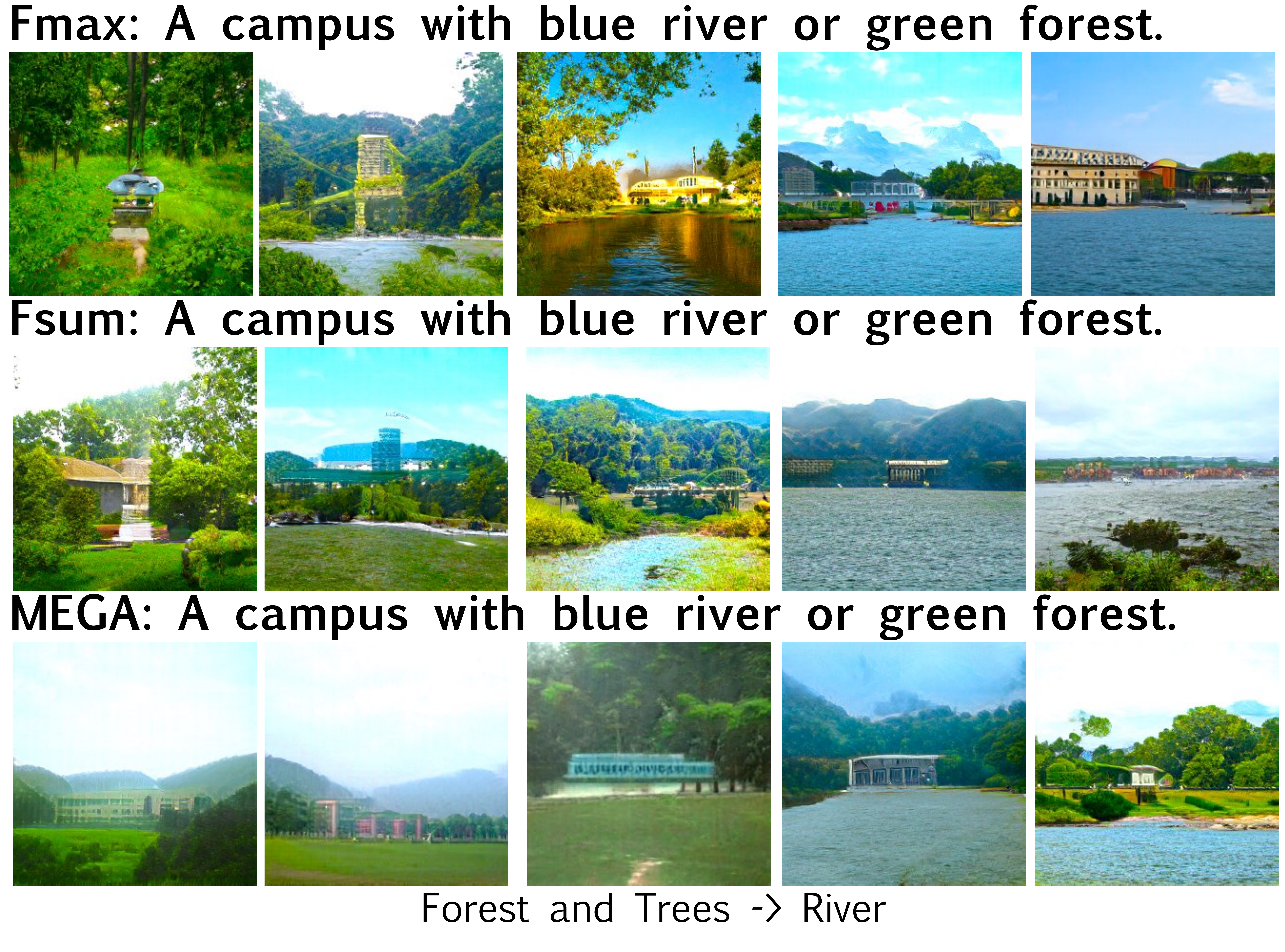}
\\
\end{tabular}}
% \vspace{-5pt}
\caption{
Results on the test function 2 in Table~\ref{tab:image2text_score}. 
% Better viewed when zoomed in.
We notice $\Fmax$, $\Fsum$ and MEGA achieve comparable quality and diversity, while $\Fmax$ and $\Fsum$ uses less time as shown in Table  \ref{tab:image2text_compare_es}.
}
\label{fig:appendix_biggan_examples2}
\vspace{-5pt}
\end{figure}

\begin{figure}[h]
\centering
\scalebox{.8}{
\begin{tabular}{c}
\includegraphics[width=1.\textwidth]{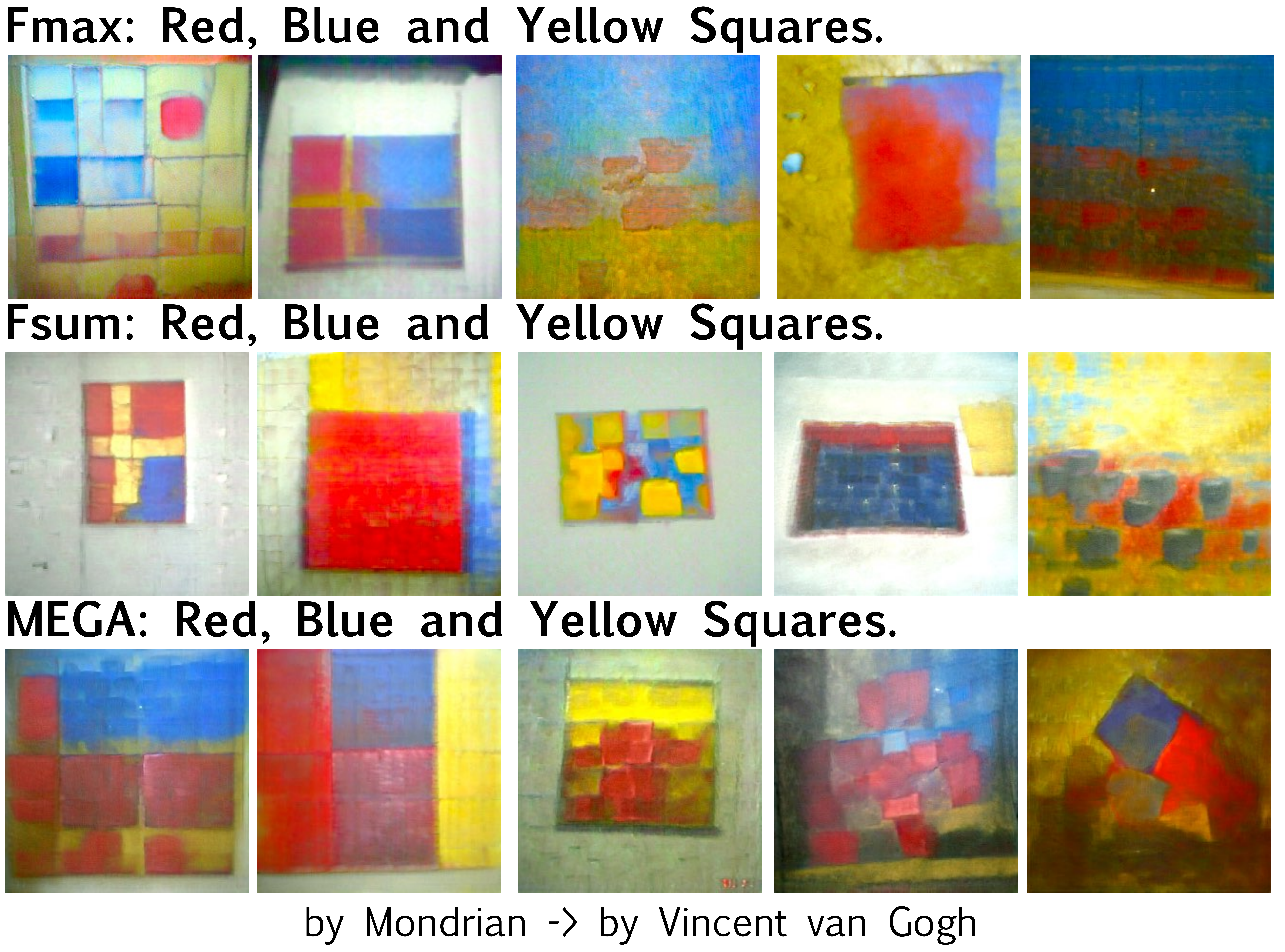}
\\
\end{tabular}}
% \vspace{-5pt}
\caption{
Results on the test function 3 in Table~\ref{tab:image2text_score}. 
% Better viewed when zoomed in.
We notice $\Fmax$, $\Fsum$ and MEGA achieve comparable quality and diversity, while $\Fmax$ and $\Fsum$ uses less time as shown in Table  \ref{tab:image2text_compare_es}.
}
\label{fig:appendix_biggan_examples3}
\vspace{-5pt}
\end{figure}

\begin{figure}[h]
\centering
\scalebox{.8}{
\begin{tabular}{c}
\includegraphics[width=1.\textwidth]{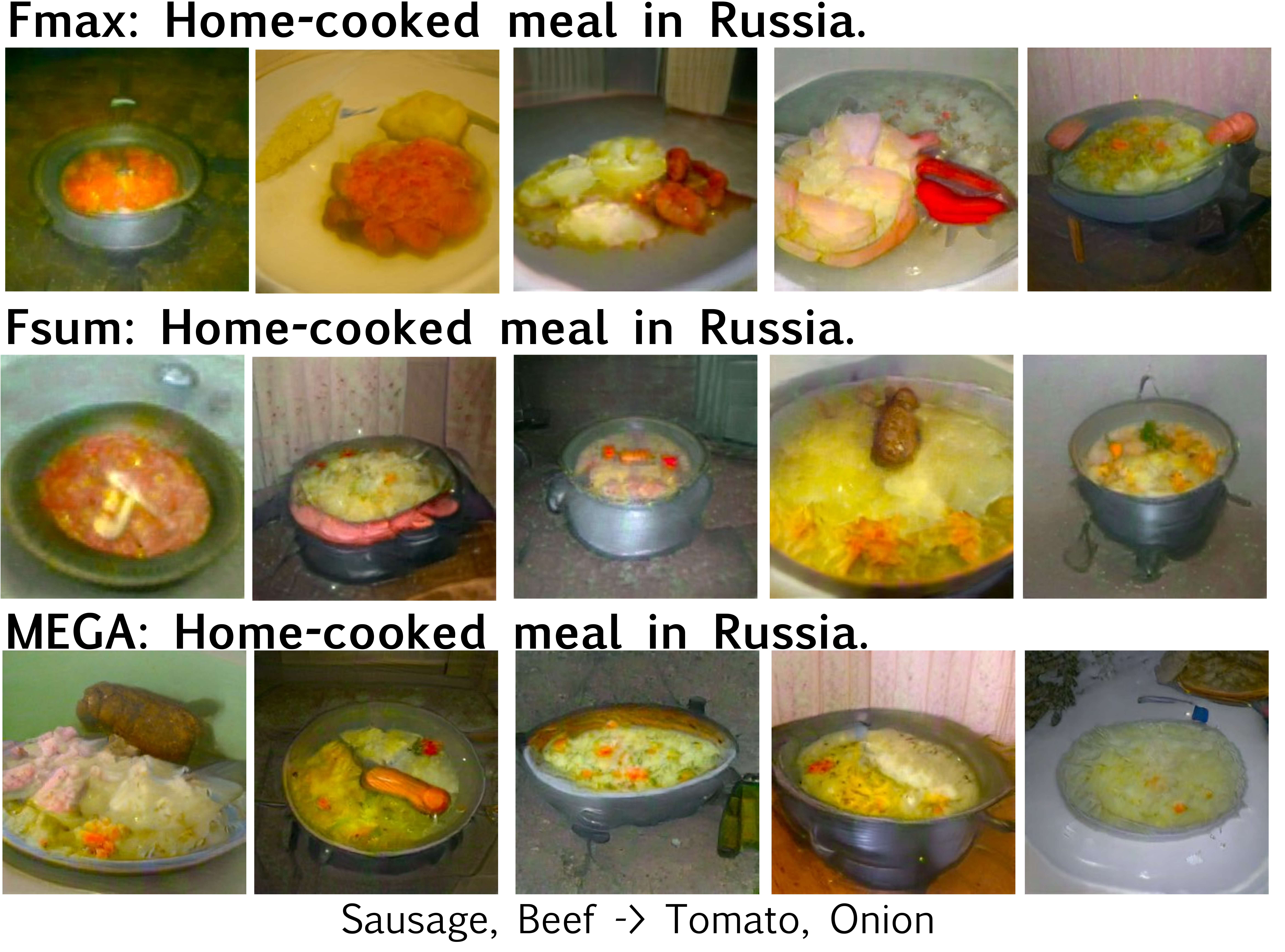}
\\
\end{tabular}}
% \vspace{-5pt}
\caption{
Results on the test function 4 in Table~\ref{tab:image2text_score}. 
% Better viewed when zoomed in.
We notice $\Fmax$, $\Fsum$ and MEGA achieve comparable quality and diversity, while $\Fmax$ and $\Fsum$ uses less time as shown in Table  \ref{tab:image2text_compare_es}.
}
\label{fig:appendix_biggan_examples4}
\vspace{-5pt}
\end{figure}

\begin{figure}[t]
\centering
\scalebox{.8}{
\begin{tabular}{c}
\includegraphics[width=1.\textwidth]{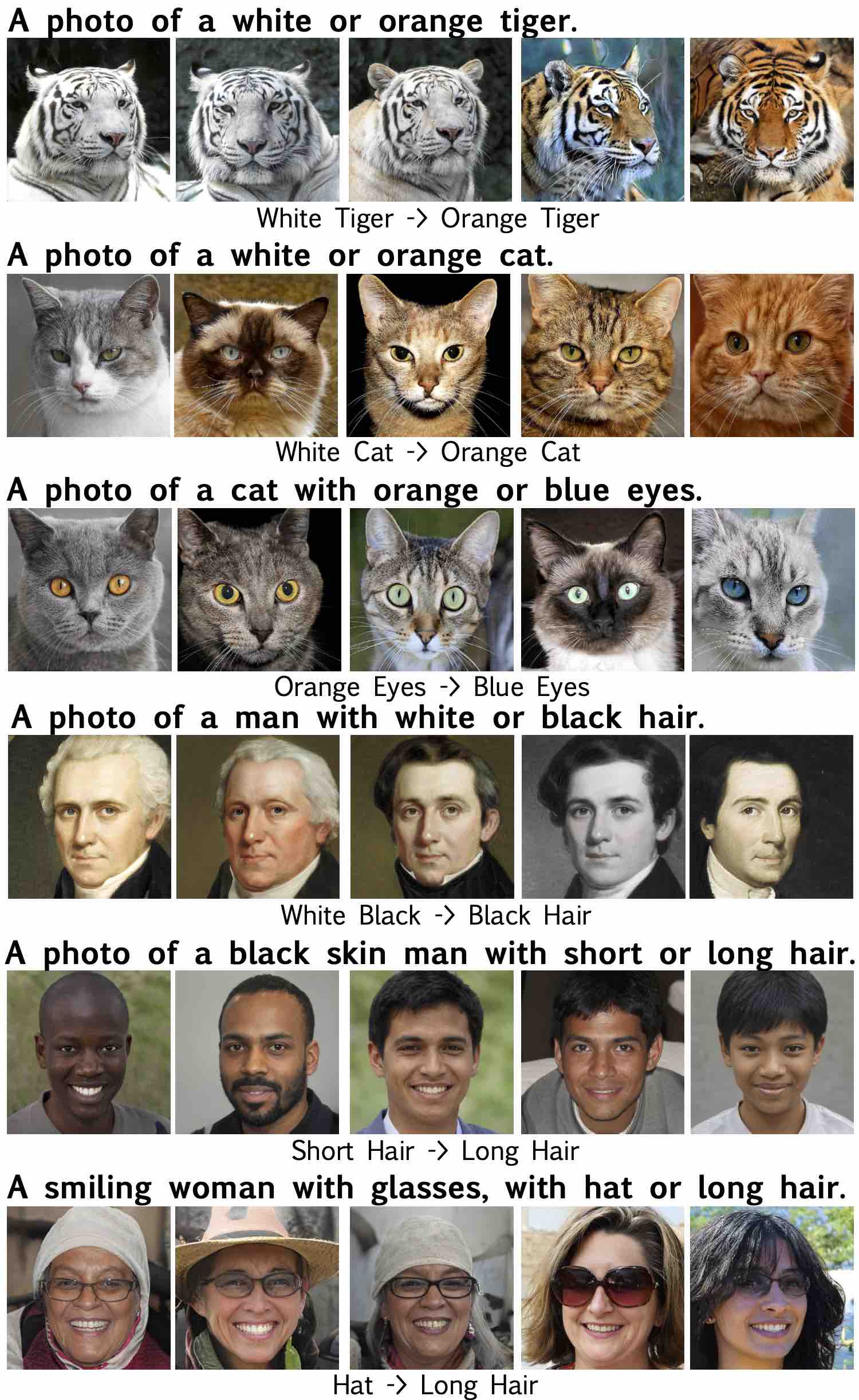}
\\
\end{tabular}}
% \vspace{-5pt}
\caption{
We apply $\Fmax$ to pre-trained StyleGAN-v2 checkpoints. The text above each image denotes $\mathcal{T}$, while the text under each image displays $\mathcal{T}_1 \rightarrow \mathcal{T}_2$. 
% Better viewed when zoomed in and see supplementary materials for examples with $\Fsum$. 
}
\label{fig:appendix_stylegan_examples}
\vspace{-5pt}
\end{figure}

\begin{figure*}[t]
\centering
\scalebox{.8}{
\begin{tabular}{c}
\includegraphics[width=1.\textwidth]{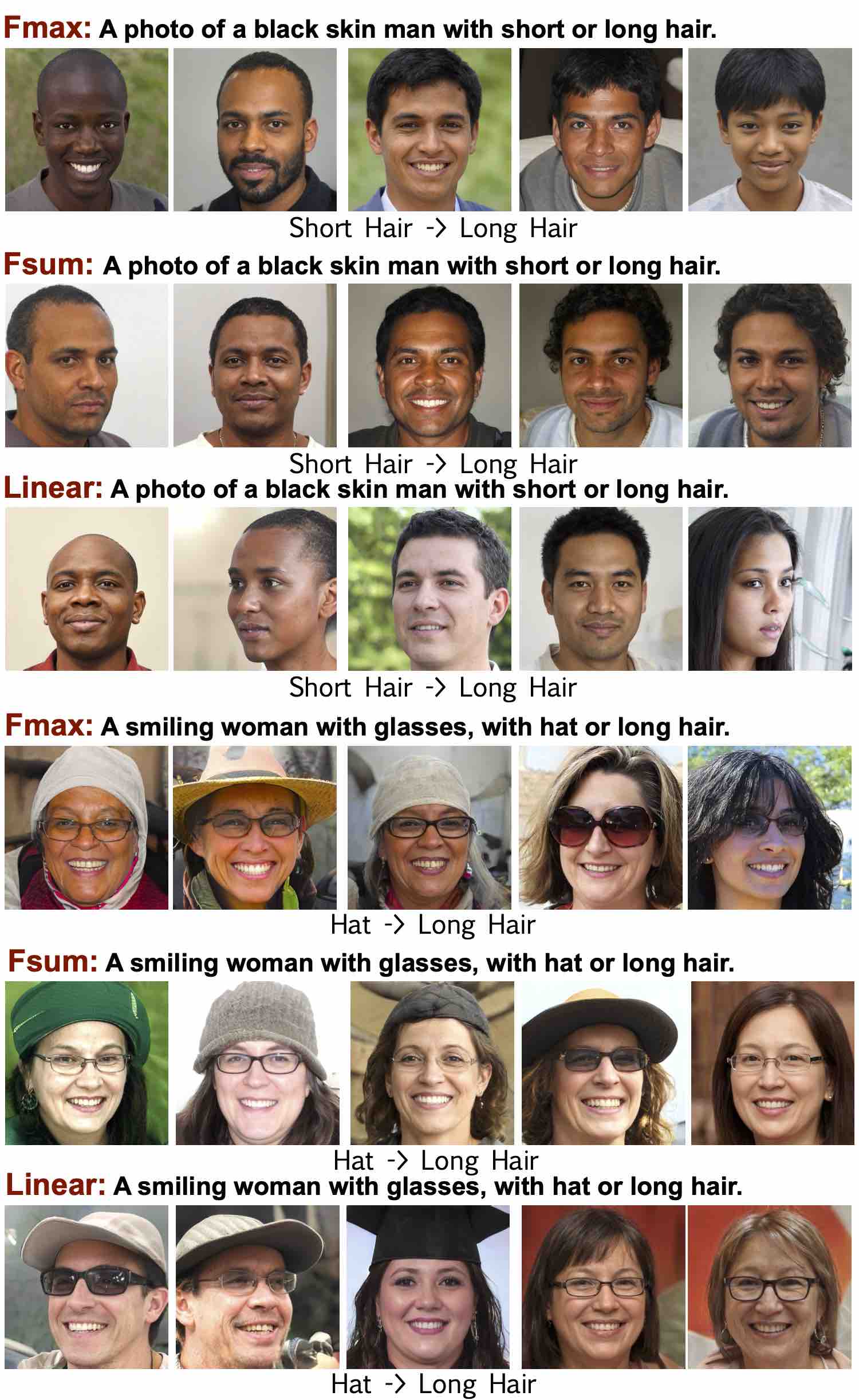}
\\
\end{tabular}}
% \vspace{-15pt}
\caption{
We apply $\Fmax$,  $\Fsum$ and the linear combination \eqref{equ:linear} ($\alpha=0.5$) to StyleGAN-v2 pre-trained on FFHQ \citep{karras2019style}. The text above each image denotes $\mathcal{T}$, while the text under each image displays $\mathcal{T}_1 \rightarrow \mathcal{T}_2$. 
% Better viewed when zoomed in.
}
\label{fig:examples}
\vspace{-15pt}
\end{figure*}

We display more examples of different methods in this section.

In Figure \ref{fig:compare_with_lexico}, we compare our method with Lexico \citep{gong2021automatic}. 
We notice that by 1) replacing the inner product constraint with quadratic constraint, 2) changing $F$ function, our approaches yield faster convergence than Lexico, and capture more local modal for the multiple modal objective.

In Figure \ref{fig:appendix_biggan_examples2}, \ref{fig:appendix_biggan_examples3} and \ref{fig:appendix_biggan_examples4}, we list the optimization results for $\Fmax$, $\Fsum$ and MEGA, the evolutionary algorithm. Visually, these methods achieve similar performance, while our approaches spend less time as shown in Section \ref{sec:experiments}.

In Figure \ref{fig:appendix_stylegan_examples}, we demonstrate the results of $\Fmax$ and $\Fsum$, and both achieve good performance. We use the checkpoints provided by StyleGAN-v2 \footnote{\url{https://github.com/NVlabs/stylegan2-ada-pytorch}}.
For cats and tigers, we use the checkpoint trained on AFHQ Cat and AFHQ Wild using adaptive discriminator augmentation..
For people face, we use the checkpoint trained on FFHQ at 1024$\times$1024 resolution.
For portraitures, we use the checkpoint trained on MetFaces at 1024$\times$1024 resolution, which does transfers learning from FFHQ using adaptive discriminator augmentation.
% We notice that the linear combination baseline ($\alpha=0.5$) fails to generate images which match $\mathcal{T}$. For example, given `smiling woman with glasses', it fails to generate women for some cases and does not generate glasses for other cases. given `black skin main', it generated women instead of men for some cases and does not generate men with block skin for other cases. 